\pdfoutput=1
\documentclass[11pt]{article}

\usepackage{acl}

\usepackage{times}
\usepackage{latexsym}

\usepackage[T1]{fontenc}
\usepackage[utf8]{inputenc}
\usepackage{multirow} 

\usepackage{microtype}

\usepackage{inconsolata}
\usepackage{longtable}
\usepackage{graphicx}
\usepackage{subcaption}
\usepackage{soul}
\usepackage{rotating}

\usepackage{adjustbox}
\usepackage{array} % Required for new column type
\usepackage{tikz}
\usepackage{pgfplots}
\newcolumntype{C}[1]{>{\centering\arraybackslash}m{#1}}

\title{A Systematic Review of Machine Learning in Sports Betting: Techniques, Challenges, and Future Directions}

\author{\fontsize{11}{11}\selectfont René Manassé Galekwa,$^{1,2}$ Jean Marie Tshimula,$^{2,3}$ Etienne Gael Tajeuna,$^{4}$ Kyamakya Kyandoghere$^{1,5}$ \\
  \fontsize{10}{10}\selectfont $^{1}$Institute of Smart Systems Technologies, University of Klagenfurt, 9020 Klagenfurt am Wörthersee, Austria \\
  \fontsize{10}{10}\selectfont $^{2}$Mathematics, Statistics and Computer Science Department, University of Kinshasa, Kinshasa, DRC \\
  \fontsize{10}{10}\selectfont $^{3}$Department of Computer Science, Université de Sherbrooke, QC J1K 2R1, Canada \\
  \fontsize{10}{10}\selectfont $^{4}$Department of Computer Science and Engineering, Université du Québec en Outaouais, QC J8X 3X7, Canada \\
  \fontsize{10}{10}\selectfont
  $^{5}$Polytechnic Faculty, University of Kinshasa, Kinshasa, DRC \\
  \fontsize{10}{10}\selectfont Correspondence email: {\tt kyandoghere.kyamakya@aau.at} \\ }

\begin{document}
\maketitle
\begin{abstract}
%The sports betting industry has become a significant financial sector, driven by technological advances and online platforms, leading to massive data generation. Over the past decade, machine learning (ML) has transformed odds setting, risk management, and strategy optimization, making it essential for both bookmakers and bettors. This review examines the latest ML approaches in sports betting, assessing their effectiveness in predicting outcomes and maximizing profitability. We explore dual motivations: enhancing profits for bookmakers through dynamic odds and empowering bettors with data-driven strategies. As the industry integrates real-time data and predictive analytics, the demand for innovative ML techniques grows, highlighting the need for transparency and ethical considerations in this evolving field.

The sports betting industry has experienced rapid growth, driven largely by technological advancements and the proliferation of online platforms. Machine learning (ML) has played a pivotal role in the transformation of this sector by enabling more accurate predictions, dynamic odds-setting, and enhanced risk management for both bookmakers and bettors. This systematic review explores various ML techniques, including support vector machines, random forests, and neural networks, as applied in different sports such as soccer, basketball, tennis, and cricket. These models utilize historical data, in-game statistics, and real-time information to optimize betting strategies and identify value bets, ultimately improving profitability.

% For bookmakers, ML supports dynamic odds adjustment and effective risk management, while bettors use data-driven insights to exploit market inefficiencies. The review also highlights ML's role in fraud detection, where anomaly detection models help identify suspicious betting patterns. Despite these advances, challenges such as data quality, real-time decision-making, and the unpredictability of sports outcomes persist. Ethical concerns around transparency and fairness are also critical.

For bookmakers, ML facilitates dynamic odds adjustment and effective risk management, while bettors leverage data-driven insights to exploit market inefficiencies. This review also underscores the role of ML in fraud detection, where anomaly detection models are used to identify suspicious betting patterns. Despite these advancements, challenges such as data quality, real-time decision-making, and the inherent unpredictability of sports outcomes remain. Ethical concerns related to transparency and fairness are also of significant importance.

% Future research should focus on developing adaptive models that integrate multimodal data and manage risk similarly to financial portfolios. This review provides a comprehensive look at the current applications of ML in sports betting, identifying both the potential and limitations of these technologies.

Future research should focus on developing adaptive models that integrate multimodal data and manage risk in a manner akin to financial portfolios. This review provides a comprehensive examination of the current applications of ML in sports betting, and highlights both the potential and the limitations of these technologies.

\end{abstract}

\section{Introduction}

Sports betting, traditionally seen as a recreational activity, has become a significant financial sector driven by technological advancements and the proliferation of online betting platforms \cite{williams2011social}. The industry allows bettors to place wagers on the outcomes of sporting events with odds that reflect the likelihood of various scenarios. As this market has grown, it has evolved from traditional betting shops to sophisticated online platforms that offer a wide range of betting options, including real-time and in-play bets. The accessibility and convenience of online betting have significantly contributed to the sector's rapid expansion, attracting a global audience and generating billions in revenue annually \cite{gainsbury2018behavioral}

The growth of sports betting has been paralleled by an explosion of data generation, making it one of the most data-intensive industries \cite{forrest2008sentiment}. This sector mirrors traditional financial markets, where odds and betting strategies are akin to stock market predictions. Bookmakers collect a large amount of data from various sources, including player statistics, team performance, live game data, and even social media sentiment \cite{haghighat2013review}. This data-driven environment provides fertile ground for the application of machine learning (ML) techniques, which have become essential to managing the complexities of odds setting, risk assessment, and optimization of betting strategy. Machine learning models, particularly those that incorporate real-time data, are crucial to maintaining competitive odds that attract bettors while ensuring profitability for bookmakers.

\begin{table*}[h!]
\centering
\caption{Research questions and their purposes}
\label{tab:research_questions}
\begin{tabular}{|>{\raggedright\arraybackslash}p{0.45\textwidth}|>{\raggedright\arraybackslash}p{0.45\textwidth}|}
\hline
\textbf{Research questions} & \textbf{Purpose} \\ \hline
\textbf{RQ1:} How can machine learning algorithms be leveraged to predict match outcomes and maximize profitability in sports betting? & To explore the potential of machine learning models to enhance predictive accuracy and maximize profitability in sports betting. \\ \hline
\textbf{RQ2:} What challenges and limitations are associated with the application of machine learning in sports betting? & To identify the existing barriers and constraints that impact the performance and adoption of machine learning models within the sports betting industry. \\ \hline
\textbf{RQ3:} How can machine learning be utilized to develop adaptive betting portfolios that optimize returns while minimizing risk, in a manner analogous to financial portfolio management? & To address a significant gap in the literature by exploring how machine learning can be employed to apply principles from financial portfolio management to the domain of sports betting. \\  \hline

\end{tabular}
\end{table*}

Machine learning has significantly impacted the sports betting landscape by improving both the accuracy of predictions and the efficiency of betting strategies. For bookmakers, ML models enable dynamic odds setting and sophisticated risk management, adjusting for new information as events unfold \cite{thabtah2019nba}. For bettors, ML provides the tools to develop data-driven strategies that improve the chances of success by identifying value bets and exploiting market inefficiencies \cite{horvat2020use, haruna2021predicting}. As a result, the sports betting industry increasingly resembles a financial sector, with both bettors and bookmakers leveraging advanced predictive analytics to maximize returns. This growing reliance on ML underscores the need for ongoing research on new techniques and emphasizes the importance of ethical considerations, such as transparency and fairness, in the implementation of ML in sports betting.

% Sports betting has long been a popular pastime, with enthusiasts relying on various analytical methods to gain an edge and increase \hl{their chances} of success \cite{matheson2021overview,gainsbury2018behavioral,williams2011social,forrest2003sport}. \hl{However}, the advent of machine learning has introduced a new paradigm in the sports betting landscape, offering the potential to revolutionize the way predictions are made and strategies are developed \cite{loeffelholz2009predicting,haghighat2013review,bunker2019machine,jimenez2023sports,pudaruth2015using,ajgaonkar2021prediction}.

Machine learning, a subset of artificial intelligence, involves the use of algorithms and statistical models to identify patterns and make predictions from data. In the context of sports betting, machine learning techniques can be applied to vast amounts of historical data, including team statistics, player performance metrics, injuries, weather conditions, and even odds movements of bookmakers \cite{hubavcek2019exploiting}. By analyzing these diverse data sources, machine learning models can uncover intricate relationships and trends that may not be apparent to human analysts. This leads to the research question ({\it RQ1}): {\it how can machine learning algorithms be leveraged to predict match outcomes and maximize profitability in sports betting?}

The application of machine learning in sports betting has garnered significant attention from researchers and industry professionals alike. Numerous studies have explored the use of various machine learning algorithms, such as support vector machines, random forests, neural networks, Bayesian and ensemble methods, to predict the outcomes of sporting events with greater accuracy \cite{walsh2024machine}. These predictive models have the potential to outperform traditional analytical methods and provide valuable insights to bettors, enabling them to make more informed decisions and potentially increasing their profitability.

In addition, machine learning techniques have been employed to identify mispriced odds offered by bookmakers, presenting opportunities for savvy bettors to capitalize on these inefficiencies \cite{ramirez2023betting,clegg2023not}. By developing models that can accurately predict match outcomes and compare them with the odds offered by bookmakers, bettors can identify instances where the odds are mispriced, allowing them to place bets with a positive expected value. Recently, anomaly detection models have been developed to identify suspicious betting patterns that may indicate match-fixing \cite{kim2024ai,ramirez2023betting,mravec2021match}. These models analyze a range of variables including sports results, team rankings, player data, and betting odds to detect abnormal behaviors that deviate from the expected. Classifying matches as normal, caution, danger, or abnormal based on ensemble model predictions, the system aims to ensure fairness and integrity in sports competitions \cite{deutscher2017match,ibrahim2016integrity}.

Despite the promising potential of machine learning in sports betting, there are several challenges and limitations. Data availability and quality can be a significant hurdle, as some sports may have limited historical data or incomplete records. Furthermore, the dynamic nature of sports, with factors such as injuries and team dynamics, can introduce uncertainties that may not be fully captured by predictive models \cite{taber2024holistic}. This raises another critical research question ({\it RQ2}): {\it what are the challenges and limitations associated with the application of machine learning in sports betting, and how can novel multimodal approaches be developed to address these issues and improve predictive performance?}

Building on this, a more holistic approach to risk management can be found by drawing parallels to financial portfolio management, where investments are balanced to maximize returns while minimizing risk. Similarly, adaptive betting portfolios could be developed using ML techniques to optimize returns for bettors. This introduces the third research question ({\it RQ3}): {\it How can machine learning be applied to create adaptive betting portfolios that optimize returns while minimizing risk, similar to financial portfolio management?} (Table \ref{tab:research_questions}).

This systematic review aims to synthesize the current state of research on the application of machine learning techniques in sports betting. By examining the existing literature, we seek to provide a comprehensive overview of the methodologies used, the challenges encountered, and the potential benefits and limitations of using machine learning in this domain. Furthermore, we explore future directions and opportunities for further research, as the field of machine learning continues to evolve and offers new avenues for innovation in sports betting.

In this review, we limit our scope to the following sports: soccer, basketball, tennis, cricket, American football, baseball, horse racing, rugby, golf, and hockey. The remainder of the paper is organized as follows. Section \S\ref{methodology} presents the methodology employed in this study; Section \S\ref{related_work} reviews the related work in the field of machine learning in sports betting; Section \S\ref{ml_sport_betting} delves into the various machine learning techniques applied to sports betting; Section \S\ref{discussion} provides a detailed discussion of the findings; Section \S\ref{dataset_fts_evaluation} evaluates the datasets and features used in the studies; Section \S\ref{ml_platforms_tips} explores the machine learning platforms for betting tips; Section \S\ref{challenge_limits} outlines the challenges and limitations encountered; and finally, Section \S\ref{future_directions} presents future directions for research in this area.

\section{Methodology}\label{methodology}
\begin{figure*}[h!]
    \centering
    \small
\begin{tikzpicture}[>=latex, font={\sf \small}]

\tikzstyle{bluerect} = [rectangle, rounded corners, minimum width=1.5cm, minimum height=0.5cm, text centered, draw=black, fill=cyan!60!gray!45!white, rotate=90, font=\sffamily]

\tikzstyle{roundedrect} = [rectangle, rounded corners, minimum width=12cm, minimum height=0.8cm, text centered, draw=black, font=\sffamily]

\tikzstyle{textrect} = [rectangle, minimum width=5.25cm, text width=5.24cm, minimum height=0.8cm, draw=black, font={\sffamily \footnotesize}]

% Adjusting the positions of nodes to reduce arrow length
\node (top1) at (0, 10cm) [draw, roundedrect, fill=yellow!80!red!70]
  {Identification of studies via database and registers};

\node (r1blue) at (-5.85cm, 7.0cm) [draw, bluerect, minimum width=3.5cm]{Identification};

\node (r1left) at (-2.75cm, 7.0cm) [draw, textrect, minimum height=2.5cm]
  {Records identified from:
     \begin{itemize}
     \item IEEE Xplore (25)
     \item Springer (20)
     \item Science Direct (19)
      \item Other databases/registers (286)
     \end{itemize}   
  };

\node (r1right) at (3.25, 7.0cm) [draw, textrect, minimum height=2.5cm]
  {Records removed before screening: 
    \begin{itemize}
    \item Duplicates (5)
    \item Ineligibles (80)
    \item Other reasons (6)
    \end{itemize} 
  };

\node (r2blue) at (-5.85cm, 2.5cm) [draw, bluerect, minimum width=4.5cm]
  {Screening};

\node (r2left) at (-2.75cm, 4.0cm) [draw, textrect, minimum height=1.2cm]
  {Records screened (350)};

\node (r2right) at (3.25, 4.0cm) [draw, textrect, minimum height=1cm]
  {Records excluded (91)};

\node (r4left) at (-2.75cm, 1cm) [draw, textrect, minimum height=1.2cm]
  {Reports assessed for eligibility (259)};

\node (r4right) at (3.25, 1cm) [draw, textrect, minimum height=3.0cm]
  {Reports Excluded: 
    \begin{itemize}
    \item Abstract only (9)
    \item Not considered machine learning (9)
    \item Not english (5)
    \item Review (3)
    \item Not during the study period(14)
    \end{itemize}     
  };

\node (r5blue) at (-5.85cm, -2.5cm) [draw, bluerect, minimum width=2cm]
  {Included};

\node (r5left) at (-2.75cm, -2.5cm) [draw, textrect, minimum height=1.5cm]
  {Studies included in review \\ 
   $n=$219 };

% Adjusted arrow lengths by moving nodes closer
\draw[thick, ->] (r1left.east) -- (r1right.west);
\draw[thick, ->] (r1left.south) -- (r2left.north);
\draw[thick, ->] (r2left.east) -- (r2right.west);
\draw[thick, ->] (r2left.south) -- (r4left.north);
\draw[thick, ->] (r4left.east) -- (r4right.west);
\draw[thick, ->] (r4left.south) -- (r5left.north);
\end{tikzpicture}
    \caption{Preferred reporting items for systematic reviews and meta-analyses flowchart of article screening process.}
    \label{fig:enter-label-prisma}
\end{figure*}
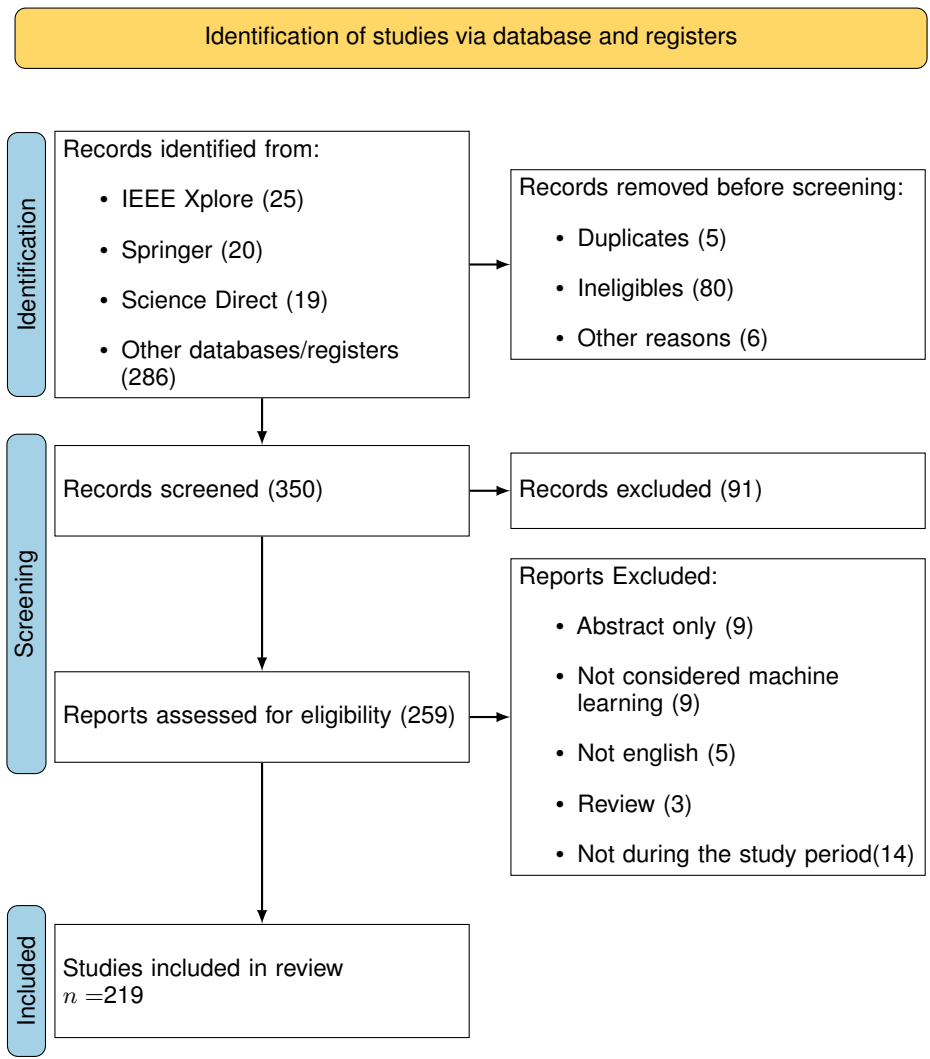

The primary objective of this systematic review is to explore the current challenges and advances in applying machine learning techniques to sports betting. The insights derived from this review will serve as a basis for future research in this rapidly evolving field. The research questions and objectives addressed in this study are described in Table \ref{tab:research_questions}.
This systematic review follows the PRISMA guidelines (Preferred Reporting Items for Systematic Reviews and Meta-Analyses) to ensure a rigorous and transparent review process (Figure \ref{fig:enter-label-prisma}). PRISMA was used to structure the review process, including the formulation of research questions, the identification of relevant studies, the evaluation of the quality of the study, the extraction of data, and the synthesis of the findings.

\subsection{Data selection}
A comprehensive search was conducted in multiple electronic databases, including IEEE Xplore, Springer, Science Direct, MDPI, arXiv, and Google Scholar. The search terms were carefully selected to capture all relevant studies and included combinations of keywords such as "machine learning", "sports betting", "predictive analytics", "odds estimation", "value betting", and specific sports such as "soccer", "basketball", and "tennis".

\subsection{Inclusion and exclusion criteria}
The inclusion criteria for this review include studies published between January 2010 and July 2024 that apply machine learning techniques to predict outcomes in sports betting. These studies consist of peer-reviewed articles, conference papers, and preprints that evaluate the effectiveness of predictive models.
A total of 259 articles were initially identified through the search. After applying the inclusion criteria, 219 articles remained for further analysis (Table \ref{nb_articles}). Exclusion criteria include non-English publications, studies focusing on machine learning methods, and articles that do not contain empirical data, such as purely theoretical or opinion papers.

\subsection{Data extraction}
Data were extracted from each included study, focusing on several key aspects: study details (including author, year, title and journal), the machine learning algorithms used, the characteristics of the dataset (such as size, type and characteristics), performance metrics (including accuracy, precision, recall and Ranked Probability Score), and key findings and limitations.

\section{Related work}\label{related_work}

The application of machine learning techniques to sports betting has gained significant attention in recent years, and researchers have explored various approaches to leverage data and predictive models to identify profitable betting opportunities.

\textbf{Outcome prediction.} A substantial portion of the literature focuses on the development of predictive models to accurately forecast the results of sporting events. \citet{bunker2019machine} proposed a machine learning framework for the prediction of sports results, evaluating the performance of several algorithms, including support vector machines (SVMs), decision trees and neural networks, on various sports datasets. Their findings suggest that advanced feature engineering, incorporating factors such as team form, head-to-head records, and home advantage, can improve predictive performance compared to using basic box-score statistics.

\citet{miljkovic2010use} explored the use of data mining techniques, such as decision trees and neural networks, to predict the outcomes of basketball matches. Their results demonstrated the potential of these methods to outperform traditional statistical models, particularly when incorporating diverse features such as team rankings, player statistics, and betting odds.

\begin{table}[h!]
\centering
\caption{Number of articles per journal/conference}
\label{nb_articles}
\fontsize{8.5}{8.5}\selectfont
\renewcommand{\arraystretch}{1.2} % Adjusts the row height
\setlength{\tabcolsep}{8pt} % Adjusts column separation
\begin{tabular}{|>{\raggedright\arraybackslash}p{4cm}|>{\centering\arraybackslash}p{2cm}|}
\hline
\textbf{Journal/Conference} & \textbf{Number of Articles} \\ \hline
Others & 65 \\ \hline
IEEE & 30 \\ \hline
Springer & 25 \\ \hline
Science Direct & 20 \\ \hline
MDPI & 18 \\ \hline
arXiv & 16 \\ \hline
Journal of Quantitative Analysis in Sports & 12 \\ \hline
Taylor and Francis Online & 10 \\ \hline
Journal of Sports Analytics & 8 \\ \hline
International Journal of Computer Science in Sport & 6 \\ \hline
IMA Journal of Management Mathematics & 5 \\ \hline
ACM & 4 \\ \hline
\end{tabular}
\end{table}

\citet{horvat2020use} conducted an initial review of machine learning techniques in the literature on sports betting, examining more than 100 studies on predicting outcomes. They identified neural networks and SVMs as the most common models and highlighted the importance of feature extraction and selection to enhance prediction accuracy. The review also pointed out the lack of standardized datasets and the need to include contextual factors such as player injuries and psychological states.

\citet{kollar2021betting} discussed the use of Artificial Neural Networks (ANN), Markov chains, and SVMs to handle complex patterns and enormous data sets to forecast sports results. They emphasized the dynamic nature of sports events, which poses challenges such as inconsistent data and the need for frequent model retraining. Feature selection and extraction were found to be crucial in improving model performance and accuracy.

\textbf{Odds estimation and value betting.} Another line of research investigates the use of machine learning models to estimate the true probability of outcomes, which can then be compared to bookmakers' odds to identify value bets--examples where the model's predicted probability differs significantly from the implied odds. \citet{franck2013inter} examined inter-market arbitrage opportunities in betting markets, highlighting the potential to exploit inefficiencies and biases in bookmakers' odds-setting processes.

\textbf{Betting strategies and risk management.} \citet{walsh2024machine} conducted a comprehensive study on the importance of model calibration in sports betting. They showed that optimizing predictive models for calibration, rather than accuracy, leads to significantly higher betting profits, with a calibration-optimized model generating 69.86\% higher average returns compared to an accuracy-optimized model. Building on this idea of model optimization in the sports betting market, \citet{arscott2022risk} showed that illegal bookmakers engage in risk management activities 6.5 times more frequently than legal bookmakers. Offshore bookmakers employ pricing policies to balance their books, reducing cash flow variance but also decreasing profits, with illegal bookmakers adjusting commissions on 39\% of their prices, compared to 6\% for legal bookmakers. This distinction highlights the differing priorities between illegal and legal bookmakers, where illegal firms prioritize risk management due to their limited access to external financing.

Continuing the examination of sports betting strategies, \citet{matej2021optimal} conducted an experimental review of the most popular betting approaches using modern portfolio theory and the Kelly criterion. They demonstrated that formal investment strategies, when applied with risk control modifications, significantly enhance profitability. Their adaptive fractional Kelly method was especially effective in different sports, highlighting the practical importance of mitigating the unrealistic assumptions inherent in pure mathematical strategies. Testing in horse racing, basketball, and soccer confirmed the necessity of these risk control methods to achieve optimal results.

\textbf{Technological advancements in sports analytics.} \citet{keys22workload} conducted a systematic review investigating innovative techniques to monitor training loads for the prediction of injury and performance. They highlighted the use of Global Positioning System (GPS), accelerometers, and Rated Perceived Exertion (RPE) to track and predict athlete performance and injury risk. Machine learning was noted for its potential to identify important predictive features, though standardized methods and more research are needed to optimize its application in sports.

\citet{naik2022comprehensive} conducted an extensive investigation into computer vision in sports, concluding with significant improvements in video analysis. This review included recognition of sports players and balls, tracking, trajectory prediction, and event classification. It emphasized how AI and machine learning improve accuracy and efficiency while tackling challenges such as occlusions, low-resolution video, and real-time analysis. For sports video analysis, the review suggested the use of complex algorithms, standard datasets, and GPU-based workstations and embedded platforms.

\begin{figure*}
    \centering
    \includegraphics[width=0.99\textwidth]{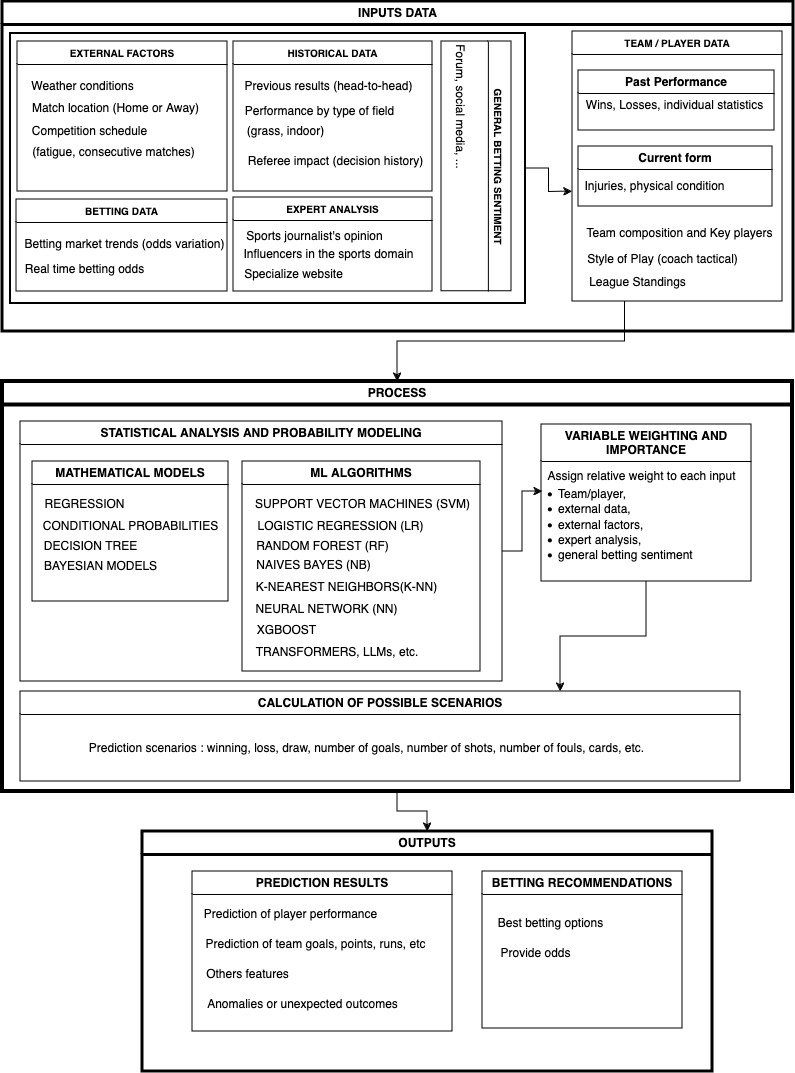}
    \caption{General overview of how to construct machine learning models for predicting sports outcomes, calculating odds, recommending betting options, and more.}
    \label{fig:enter-label-soccer-performance-overview}
\end{figure*}

\section{Machine learning in sport betting}\label{ml_sport_betting}

%{Machine learning techniques have been extensively applied to various sports betting scenarios, showcasing their potential to enhance prediction accuracy and profitability. Research has demonstrated the effectiveness of models like artificial neural networks, support vector machines, and ensemble methods such as XGBoost in sports like soccer, basketball, tennis, cricket, American football, baseball, horse racing, rugby, golf, and hockey. These models leverage vast datasets, including historical match data, player statistics, and betting odds, to uncover patterns and trends that inform betting strategies.}
Machine learning techniques have been extensively applied in various sports betting scenarios, demonstrating their potential to improve prediction accuracy and profitability. Research has demonstrated the effectiveness of models, including artificial neural networks, support vector machines, and ensemble methods in sports (soccer, basketball, tennis, cricket, American football, baseball, horse racing, rugby, golf, and hockey). These models leverage vast datasets, including historical match data, player statistics, and betting odds, to uncover patterns and trends that inform betting strategies. For instance, in soccer, methods such as the Rank Probability Score (RPS) and Principal Component Analysis (PCA) are utilized to identify betting inefficiencies and predict the outcomes of matches (\S\ref{lbl_soccer}). In basketball, the integration of data mining techniques and feature engineering has led to a high prediction accuracy through methods such as logistic regression and gradient boosting (\S\ref{lbl_basketball_sec}). Similarly, in tennis, models that incorporate player-specific statistics and match conditions have shown high returns on investment, underscoring the economic viability of machine learning in sports betting (\S\ref{lbl_tennis_sec}). Cricket predictions have used decision trees and multilayer perceptrons, demonstrating the importance of robust feature selection and data preprocessing techniques to improve model performance (\S\ref{lbl_cricket}). In American football, hidden Markov models and ensemble methods such as XGBoost have achieved high prediction accuracy (\S\ref{lbl_american_football}). Baseball research has used Markov processes and machine learning algorithms to predict pitch types and game outcomes, emphasizing the importance of detailed pitch data (\S\ref{lbl_baseball_sec}). Horse racing models have employed XGBoost and agent-based models to develop profitable betting strategies (\S\ref{lbl_horse_racing_sec}). Rugby predictions have used continuous-time models and random forests to predict match outcomes and player performances (\S\ref{lbl_rugby_sec}). Golf studies have focused on advanced analytics and proprietary data to forecast player performance, highlighting the challenges of predicting outcomes in individual sports (\S\ref{lbl_golf_sec}). Lastly, hockey predictions have utilized various machine learning models to predict match outcomes and player actions, showcasing the potential of machine learning to provide valuable insights for betting strategies in complex dynamic team sports (\S\ref{lbl_hockey_sec}). These advancements highlight the transformative potential of machine learning in optimizing betting strategies in different sports, making it a valuable tool for bettors and analysts alike.

Figure \ref{fig:enter-label-soccer-performance-overview} illustrates a comprehensive system for predicting sports betting outcomes using machine learning. The construction of such machine learning models includes multiple data sources, including external factors (for example, weather and match location), historical performance data, real-time betting odds, expert analysis, and general public sentiment from social media. 

The outputs generated by this system include probability-based predictions for match outcomes (such as win, loss, or draw), detailed performance metrics (number of goals, number of shots, cards, etc.), and betting recommendations (such as Bet Boosts). By combining historical and real-time data with expert opinions and betting market trends, the model can improve the accuracy of predictions and provide bettors and stakeholders with informed betting options and odds.

\begin{figure*}
    \centering
    \includegraphics[width=0.85\textwidth]{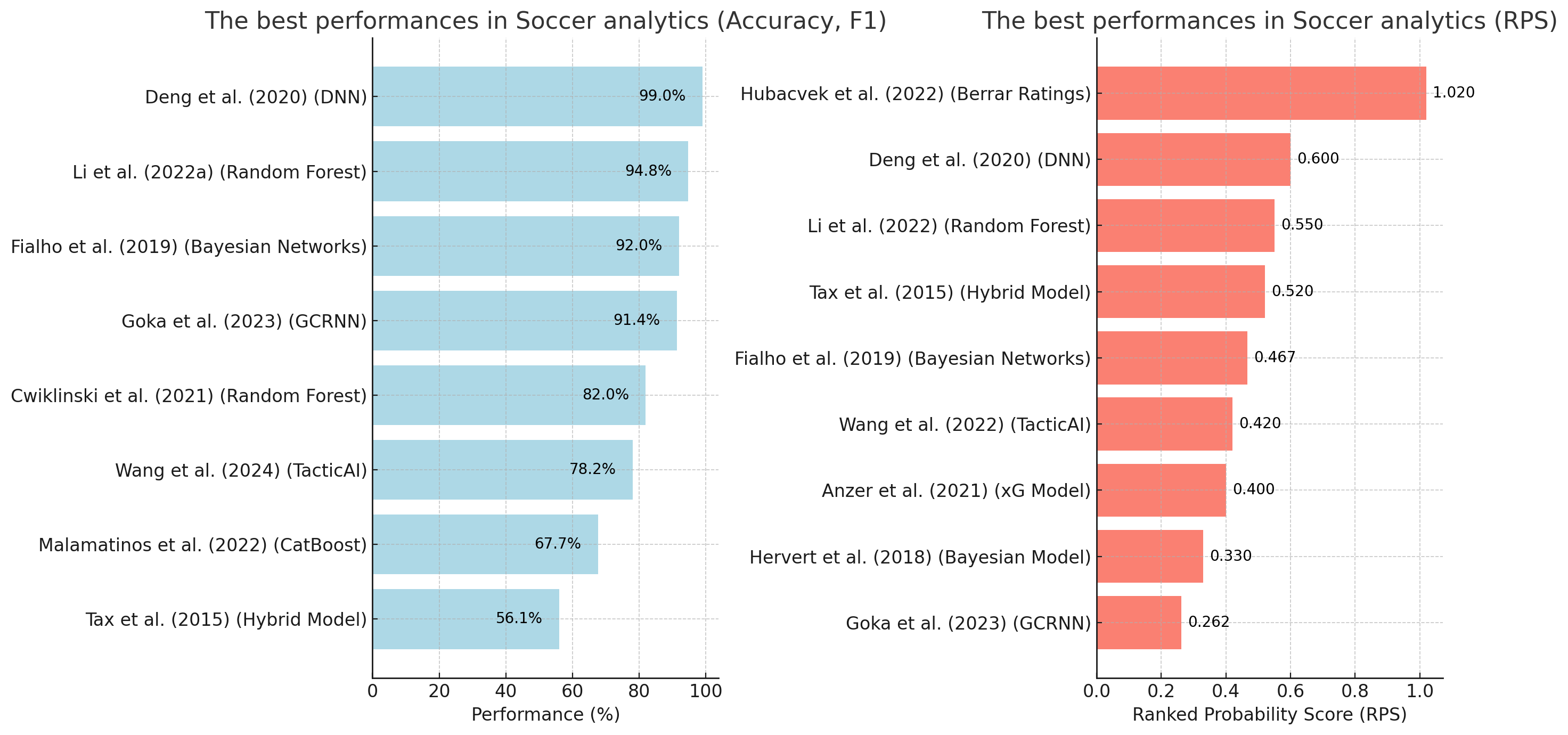}
    \caption{The best performances in Soccer analytics based on accuracy, F1 and RPS}
    \label{fig:enter-label-soccer-performance-sc}
\end{figure*}

\subsection{Soccer \includegraphics[height=0.3cm]{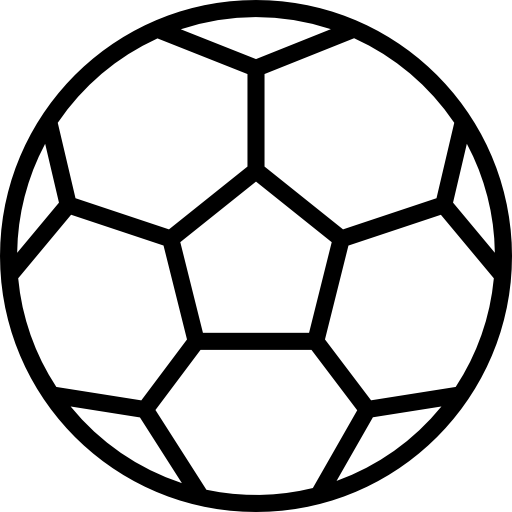}}\label{lbl_soccer}

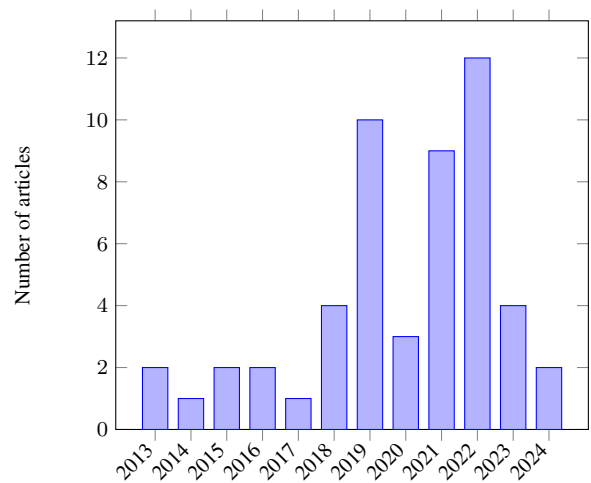
\begin{figure}[h!]

\fontsize{8}{8}\selectfont

\begin{tikzpicture}
\begin{scope}[shift={(-2cm,0)}] % Déplace l'histogramme de 2 cm vers la gauche
\begin{axis}[
    ybar,
    symbolic x coords={2013, 2014, 2015, 2016, 2017, 2018, 2019, 2020, 2021, 2022, 2023, 2024},
    xtick=data,
    x tick label style={rotate=45, anchor=east},
    ylabel={Number of articles},
    xlabel={},
    ymin=0,
    bar width=9.5pt,
    width=7.8cm,
    height=7cm
]
\addplot coordinates {(2013,2) (2014,1) (2015,2) (2016,2) (2017,1) (2018,4) (2019,10) (2020,3) (2021,9) (2022,12) (2023,4) (2024,2)};
\end{axis}
\end{scope}
\end{tikzpicture}
\caption{Histogram showing the number of articles per year in soccer betting.}
    \label{fig:articles_per_year_soccer}
\end{figure}

Soccer analytics has evolved significantly over the years, with numerous studies exploring various methodologies to predict match outcomes, player performance, and tactical strategies. This section reviews notable research contributions in the field, emphasizing methodologies, datasets, and results (Table \ref{lbl_soccer_tbl} and Figures \ref{fig:enter-label-soccer-performance-sc} and \ref{fig:articles_per_year_soccer}).

The study of inefficiencies in the online European football gambling market by \citet{constantinou2013profiting} spaned seven seasons (2005/06 to 2011/12) and used data from 14 European leagues to explore arbitrage opportunities and biases in bookmakers' odds. The methodology involved calculating profit margins and assessing bookmakers' accuracy using the Rank Probability Score (RPS). Significant arbitrage opportunities were revealed due to the varied profit margins between bookmakers, with no improvement in the accuracy of the odds over time. The dataset used was from \url{www.football-data.co.uk}.

In the context of match outcome prediction, \citet{tax2015predicting} predicted results in the Dutch Eredivisie using public data, covering thirteen seasons (2000-2013). Utilizing dimensionality reduction techniques such as PCA and classification algorithms such as Naïve Bayes, Multilayer Perceptron, and LogitBoost, the study achieved a highest prediction accuracy of 56.054\% with a hybrid model combining public data and betting odds. The evaluation metrics included prediction accuracy and McNemar's test, with datasets sourced from \url{www.football-data.co.uk}, \url{www.elfvoetbal.nl}, \url{www.transfermarkt.co.uk}, and \url{www.fcupdate.nl}.

\begin{table*}[h!]
\centering
\caption{Summary of soccer studies by families of approaches}\label{lbl_soccer_tbl}
\fontsize{6.5}{6.5}\selectfont
\renewcommand{\arraystretch}{1.6}
\begin{tabular}{|C{2cm}|C{2cm}|C{2cm}|C{1.5cm}|C{2cm}|C{3.5cm}|}
\hline
\textbf{Approaches} & \textbf{Work} & \textbf{Performance} & \textbf{Metrics} & \textbf{Features} & \textbf{Datasets} \\
\hline

\textbf{Dimensionality reduction and classification} 
& \citet{tax2015predicting} & Highest accuracy 56.054\% & Prediction accuracy, McNemar's test & Dimensionality reduction, classifier combinations & \url{www.football-data.co.uk}, \url{www.elfvoetbal.nl}, \url{www.transfermarkt.co.uk}, \url{www.fcupdate.nl} \\
\hline

\textbf{Bayesian approaches} 
& \citet{hervert2018prediction}, \citet{hervert2018bayesian}, \citet{fialho2019predicting} & RPS of 0.2620, RPS average 33\%, 92\% accuracy & Prediction accuracy, RPS & Historical patterns, team ranking, player attributes & Over 200,000 match results from 52 leagues, Open International Soccer Database \\
\hline

\textbf{Geometric Deep Learning and Spatio-Temporal Tracking} 
& \citet{wang2024tacticai}, \citet{goka2023prediction} & High prediction accuracy, AP of 0.967, F1 score of 0.914 & Top-3 accuracy, F1 score, human expert assessments, AP & Spatio-temporal trajectory frames, event stream data, player profiles, Players' spatial-temporal relations & Liverpool FC, 7,176 corner kicks from the 2020-2021 Premier League seasons, 400 video clips from the 2019 and 2020 Japan J1 League seasons \\
\hline

\textbf{Extreme gradient boosting} 
& \citet{anzer2021goal}, \citet{geurkink2021machine}, \citet{malamatinos2022predicting}, \citet{toda2022evaluation} & RPS of 0.197, Accuracy 89.6\%, Best-performing model CatBoost accuracy 67.73\%, Mean F1 score greater than 0.483 & Precision, recall, AUC, F1 score, RPS & Synchronized positional and event data, Shots on target, distance in speed zones, number of accelerations, ELO-ratings, total transfer value of benched players, Player actions & ChyronHego’s TRACAB system, Belgian Pro League, SportVU system, \url{www.football-data.co.uk}, team budget data from \url{www.transfermarkt.de}, Meiji Yasuda Seimei J1 League \\
\hline

\textbf{Regression approaches} 
& \citet{li2022machine}, \citet{andrews2021analysis}, \citet{hubavcek2022forty} & R² value of 0.606 (linear), R² value of 0.948 (random forest), F1 score 0.6119, precision 0.6563 (training), F1 score 0.5652, precision 0.6000 (test), Berrar ratings RPS 0.2101, cross-entropy 1.0246 & AIC, BIC, RMSE, R², F1 score, precision, cross-entropy, accuracy & Age, goals, shots on target, achievement index, Goals scored, team rankings, match-related statistics, Various statistical models and rating systems & FBREF, CAPOLOGY, 15 seasons of EPL, CRISP-DM framework, Open International Soccer Database v2, 218,916 match results \\
\hline

\textbf{Random forest} 
& \citet{stubinger2019machine}, \citet{eryarsoy2019predicting}, \citet{rico2023machine} & Return of 1.58\% per match, Accuracy 74.6\%, Injury prediction accuracy over 66\% & Accuracy, RMSE, MAD, sensitivity, specificity & 40 features per player, general characteristics, ball skills, passing, shooting, defense, Situational variables, team performance indicators, Injury risks, performance metrics & \url{www.fifaindex.com}, Bet365, Turkish Super League, 32 original studies applying machine learning to soccer data \\
\hline

\textbf{Deep neural networks} 
& \citet{deng2020analysis}, \citet{lee2022dnn} & DNN accuracy 99\%, Logistic Regression 95\%, Decision Tree 91\%, Random Forest 84\%, Formations accuracy up to 94.97\%, game styles accuracy up to 83.82\%, game outcomes accuracy up to 57.82\% & Prediction accuracy, Accuracy & Possession, total shots, shot efficiency, goal efficiency, Player performance data, player positions & Kaggle, 11 European countries, 2008-2016, 11 seasons of EPL (Tottenham Hotspur), \url{www.whoscored.com} \\
\hline

\textbf{Ensemble learning} 
& \citet{hubavcek2022forty}, \citet{joseph2022time} & Berrar ratings RPS 0.2101, cross-entropy 1.0246, highest accuracy of the LSTM model 43.5\%, MAE 0.62 & RPS, cross-entropy, accuracy, MAE & Various statistical models and rating systems, Match results, team rankings & Open International Soccer Database v2, 218,916 match results, EPL matches, 1993/1994 to 2020/2021 seasons \\
\hline

\textbf{Hybrid approaches} 
& \citet{elmiligi2022predicting} & Best-performing model accuracy 46.6\%, RPS 0.2176 & Prediction accuracy, RPS & League and team statistics, home/away games, historical match data & 205,182 match results from 2000/2001 to 2016/2017 seasons \\
\hline

\textbf{Miscellaneous} 
& \citet{shin2014novel}, \citet{rahman2020deep} & Virtual predictor linear SVM accuracy up to 80\%, real data models accuracy 75\%, FIFA World Cup 2018 accuracy 63.3\% & Prediction accuracy, AUC & Physical and technical skills, Team rankings, performances, historical match results & FIFA 2015 video game, Spanish La Liga fixtures, ``International Football results from 1872 to 2018,'' ``FIFA World Cup 2018 Dataset,'' ``FIFA Soccer Rankings'' \\
\hline

\end{tabular}
\end{table*}

Similarly, \citet{hervert2018prediction} employed a Bayesian approach combined with historical match data to forecast the results of soccer matches. The methodology involved ranking teams based on performance and calculating probabilities adjusted with triangular distributions. The model used data from over 200,000 match results and the 2018 FIFA World Cup group stage matches, achieving an RPS of 0.2620, highlighting the model's high accuracy.

Expanding on AI applications, \citet{wang2024tacticai} developed TacticAI, a tool to optimize football tactics, particularly corner kicks, using geometric deep learning in spatio-temporal player tracking data. Validated on 7,176 corner kicks from the 2020-2021 Premier League seasons, the tool demonstrated high prediction accuracy and favorable expert assessments 90\% of the time. The dataset was provided by Liverpool FC, comprising spatio-temporal trajectory frames, event stream data, and player profiles.

\citet{goka2023prediction} presented a novel method for predicting shooting events using players' spatial-temporal relations through complete bipartite graphs. The study used Mask R-CNN for player detection and a graph convolutional recurring neural network (GCRNN) to capture latent features, resulting in an AP of 0.967 and an F1 score of 0.914 on a dataset of 400 video clips from the 2019 and 2020 Japan J1 League seasons.

In the realm of expected goals (xG) models, \citet{anzer2021goal} introduced an advanced xG model using an extreme gradient boosting algorithm, analyzing 105,627 shots from the German Bundesliga. The model achieved a superior accuracy with an RPS of 0.197, utilizing data from ChyronHego’s TRACAB system and event data from the German Bundesliga.

Focusing on player valuation, \citet{li2022machine} used machine learning to evaluate the market value of football players, using data from FBREF and CAPOLOGY. The study compared multiple linear regression models and random forest models, the latter achieving an R² value of 0.948. Key metrics included Akaike Information Criterion (AIC), Bayesian Information Criterion (BIC), RMSE, and R² values.

\citet{peters2022betting} explored the impact of lineups on football score predictions using historical player data from the English Premier League (2020-2022). The study employed various machine learning models, including SVR, which outperformed other techniques to predict the final scores. The dataset comprised 680 matches from the English Premier League, sourced from \url{www.fixturedownload.com} and \url{www.FBRef.com}.

Furthermore, \citet{deng2020analysis} analyzed soccer match data from 11 European countries (2008-2016) using logistic regression, decision trees, random forests, and a deep neural network (DNN), achieving a highest accuracy of 99\% with the DNN model. The dataset included detailed match events and betting odds from various bookmakers, sourced from Kaggle.

\citet{cwiklinski2021will} used machine learning to improve the building of football teams and player transfer decisions, employing Random Forest, Naïve Bayes, and AdaBoost algorithms. The study used Sofascore data, encompassing 4700 players from top European leagues over four seasons, achieving an accuracy of 0.82 and an F1 score of 0.83.

In predicting match outcomes in the English Premier League, \citet{ganesan2018english} applied SVM, XGBoost, and logistic regression, with XGBoost showing optimized performance. The dataset was sourced from football-data.co.uk, covering multiple seasons and various attributes such as team performance and venue.

\citet{andrews2021analysis} utilized logistic regression, SVM, and XGBoost to predict the outcomes of the EPL matches, with logistic regression emerging as the model with the best performance, achieving an F1 score of 0.6119. The study employed the CRISP-DM framework, using historical match data from the past 15 seasons.

\citet{stubinger2019machine} utilized an ensemble of machine learning algorithms to predict the outcome of matches using data from the five major European football leagues, covering 47,856 matches between 2006 and 2018. The ensemble model achieved a return of 1.58\% per match, outperforming individual models and naive betting strategies.

Using the Open International Soccer Database v2, \citet{hubavcek2022forty} evaluated statistical models and rating systems for predicting soccer match outcomes, with the Berrar ratings and Double Weibull models achieving the best performance. The dataset contained 218,916 match results from 52 leagues since the 2000/01 season.

\citet{mattera2023forecasting} employed score-driven models to predict binary outcomes in soccer matches, achieving high predictive accuracy with the generalized autoregressive score model (GAS). The study analyzed 13 seasons of matches from the English Premier League and Italian Serie A, with datasets sourced from football-data.co.uk.

\citet{liti2017predicting} predicted the outcomes of Serie A TIM matches tied at half-time using Naïve Bayes, LibSVM, and RBFClassifier. The dataset comprised 166 matches from the 2014-2015 and 2015-2016 seasons, with promising results shown by Naïve Bayes and RBFClassifier.

Focusing on goal-scoring opportunities, \citet{eggels2016expected} proposed a predictive model using logistic regression, decision trees, random forest, and AdaBoost. The dataset included ORTEC tactical data and Inmotio player tracking data, providing valuable insight into team performance and strategy.

\citet{fialho2019predicting} reviewed AI models for soccer prediction, highlighting Bayesian networks, logistic regression, ANN, SVM, and fuzzy logic systems. The Open International Soccer Database was used, containing over 216,743 match records, with various features that influence the accuracy of the prediction.

\citet{bunker2024machine} emphasized feature engineering and model evaluation, using gradient-boosted tree models such as CatBoost for prediction of soccer matches. The primary dataset was the Open International Soccer Database, showcasing the potential to integrate player- and team-level information.

\citet{geurkink2021machine} identified key variables predicting match outcomes in Belgian professional soccer using Extreme Gradient Boosting. The 576-game dataset was tracked by the SportVU system, achieving an accuracy of 89.6\%.

\citet{malamatinos2022predicting} applied machine learning models to predict outcomes in the Greek Super League, CatBoost achieving the highest accuracy of 67.73\%. The dataset covered six seasons (2014-2020), sourced from \url{www.football-data.co.uk} and \url{www.transfermarkt.de}.

\citet{yao2022goal} proposed a two-stage method for predicting in-game soccer outcomes using the Bernoulli distribution, achieving superior performance in predicting draws and final outcomes. The dataset included matches from the Chinese Super League (2012-2019).

%{\citet{aslan2007comparative} utilized neural networks to predict Serie A 2001-2002 season match results, with LVQB outperforming LVQA. The study highlighted the importance of input parameter selection for neural networks in soccer prediction.}

\citet{capobianco2019can} developed machine learning models using Random Forest for the Italian Serie A 2017-2018 season, achieving a precision of 0.857 and 0.879 for predicting outcomes and goals, respectively. The dataset included various match features from the reports.

\citet{hassan2020predicting} used a radial basis function neural network (RBFNN) for the 2018 FIFA World Cup, achieving 83.3\% accuracy for wins and 72.7\% for losses. The dataset included 57 matches, analyzed by FIFA using the TRACAB® system.

\citet{groll2019hybrid} combined random forests with Poisson ranking methods for international soccer match predictions, achieving substantial improvement over traditional models. The dataset included matches from the FIFA World Cups between 2002 and 2014.

\citet{ulmer2013predicting} applied machine learning models to the EPL seasons (2002-2003 to 2013-2014), with Linear SGD and Random Forest achieving the lowest error rates. The dataset was sourced from Football-Data, covering 3800 games for training and 760 for testing.

\citet{eryarsoy2019predicting} predicted Turkish Super League matches using Naïve Bayes, Decision Trees, and ensemble methods. The dataset included 3,060 matches, with the ensemble methods showing superior performance.

\citet{hubavcek2019learning} used relational and feature-based methods to predict soccer match outcomes, achieving the lowest RPS with gradient boosted trees. The dataset was from the Soccer Prediction Challenge 2017.

\citet{toda2022evaluation} introduced the VDEP method to evaluate soccer team defense, achieving a mean F1 score greater than 0.483. The dataset included player actions and positional data from 45 matches of the 2019 Meiji Yasuda Seimei J1 League season.

\citet{robberechts2021bayesian} used a Bayesian statistical framework to estimate the probability of wins, draws, and losses in soccer, demonstrating improved prediction accuracy over traditional methods. The dataset covered eight seasons of the top five European soccer leagues.

\citet{talattinis2019forecasting} utilized cost-sensitive classification models to predict EPL match outcomes, focusing on the Sharpe ratio as a performance metric. The dataset included historical data and odds from the bookmakers of 2010-2016, obtained from \url{www.football-data.co.uk}.

\citet{palinggi2019predicting} predicted EPL match outcomes using weather conditions as features, with SVM achieving the highest accuracy. The dataset covered the 2010/2011 to 2015/2016 seasons.

\citet{lee2022dnn} presented a DNN-based model to predict soccer team tactics, achieving up to 94.97\% accuracy for formations. The dataset included 11 seasons of Tottenham Hotspur player performance data, sourced from \url{www.whoscored.com}.

\citet{cho2018using} combined social network analysis and gradient boosting to predict the outcomes of Champions League matches, outperforming other classifiers. The dataset included passing distribution data.

\citet{shin2014novel} used virtual data from FIFA 2015 to predict the outcomes of soccer matches, achieving up to 80\% accuracy with a linear SVM model. The dataset included player features of the video game.

\citet{joseph2022time} applied time series methods to predict the outcomes of the EPL matches, with the LSTM model achieving the highest accuracy of 43.5\%. The dataset covered the EPL match results from 1993/94 to 2020/21 seasons.

\citet{stenerud2015study} developed a soccer prediction model using a Poisson process, achieving a small profit over five seasons. The dataset covered 14 seasons from 2000-2014, sourced from \url{www.football-data.co.uk}.

\citet{vaknin2021predicting} compared Poisson-based models and classification models to predict events related to scores, with classification models showing superior performance. The dataset included detailed match statistics and player ratings.

\citet{parker2023predictive} employed machine learning models to predict the outcomes of matches in the Mexican football league, achieving 81\% to 84\% accuracy. The dataset covered 2,612 games with 17 attributes per match.

\citet{parim2021prediction} analyzed UEFA Champions League group stage matches, identifying significant performance indicators. The dataset included 1920 match results over ten years, sourced from \url{www.WhoScored.com}.

\citet{rahimian2023pass} utilized Temporal Graph Networks to predict the outcomes of the soccer passes, achieving an F1 score of 0.95 and an AUC of 0.92. The dataset included 358,790 passes from 330 Belgian Pro League matches.

\citet{groll2015prediction} used a regularized Poisson regression model to predict the outcomes of the FIFA World Cup, achieving strong predictive power. The dataset included past World Cup results and various team-specific covariates.

\citet{du2022sports} developed a framework to predict the attendance of sports games, with XGBoost achieving the best performance. The data set covered 1200 Super League games from the China Football Association over three seasons.

\citet{stubinger2018beat} used machine learning models to predict the outcomes of football matches, achieving 75.62\% accuracy and a 5.42\% return per match with a Random Forest model. The dataset included 8,082 matches from the top five European leagues.

\citet{rahman2020deep} used deep neural networks to predict the outcomes of the FIFA World Cup 2018, achieving 63.3\% accuracy. The dataset included international football results and FIFA rankings.

\citet{berrar2019incorporating} introduced novel feature engineering methods for soccer prediction, with XGBoost achieving the best performance. The dataset was from the 2017 Soccer Prediction Challenge.

\citet{rico2023machine} analyzed 32 studies on machine learning in soccer, categorizing them into injury prediction, performance prediction, and talent identification. The methodologies used various machine learning algorithms, achieving accuracies greater than 66\% for injury prediction and varying accuracies for performance and talent prediction.

\begin{table*}[h!]
\centering
\caption{Summary of Basketball prediction models}\label{lbl_basketball}
\fontsize{6.5}{6.5}\selectfont
\renewcommand{\arraystretch}{1.6}
\begin{tabular}{|C{1.5cm}|C{1.5cm}|C{1.5cm}|C{1.5cm}|C{3.5cm}|C{3.5cm}|}
\hline
\textbf{Approaches} & \textbf{Work} & \textbf{Performance} & \textbf{Metrics} & \textbf{Features} & \textbf{Datasets} \\ \hline

\textbf{XGBoost} & \citet{chen2021hybrid} & 91.82\% & Accuracy & Defensive rebounds, 2P FG\%, FT\%, Offensive rebounds, Assists, 3P FG attempts & NBA 2018-2019 season (2460 game data points) \\ \hline

\textbf{Logistic Regression} & \citet{cao2012sports}, \citet{lin2014predicting}, \citet{horvat2020impact}, \citet{houde2021predicting}, \citet{sukumaran2022application} & 69.67\%, 68.75\%, 60.82\%, 65.1\%, 93.20\% & Accuracy & Comprehensive NBA statistics data mart, Points scored, FG attempts, Defensive rebounds, Assists, Turnovers, Record, Team assists, Steals, Personal fouls, FG attempts, Rebounds, Win percentage, FG\%, 3P\%, FT\%, Rebounds, Assists, Turnovers, Steals, Blocks, Plus-minus, Offensive rating, Defensive rating, True shooting percentage & NBA data from 5 regular seasons for training, 1 for scoring, NBA games from 1991-1998, NBA Enhanced Box Score and Standings (2012-2018) from Kaggle, Games from three NBA seasons (2018-2021) \\ \hline

\textbf{Random Forest} & \citet{lin2014predicting}, \citet{zhang2021modeling}, \citet{zhao2023enhancing}, \citet{cai2019hybrid}, \citet{houde2021predicting} & 65.15\%, 64.63\%, 71.54\%, 84\%, 65.1\% & Accuracy & Points scored, FG attempts, Defensive rebounds, Assists, Turnovers, Record, Team strength, Home court advantage, Player tiredness, Team assists, Steals, Personal fouls, FG attempts, Rebounds, Two-point shots, Three-point shots, Free throws, Attack, Defense, Assists & NBA games from 1991-1998, NBA official website (2016-2021), Professional Transactions Archive, Spotrac/ESPN, NBA Enhanced Box Score and Standings (2012-2018) from Kaggle, NBA 2018-2019 season (2460 game data points) \\ \hline

\textbf{SVM} & \citet{cao2012sports}, \citet{jain2017machine}, \citet{li2021data} & 69.67\%, 89.26\%, 73.95\% & Accuracy & Comprehensive NBA statistics data mart, 33 condition attributes including biometric data, college stats, draft order, Two-point shots, Three-point shots, Free throws, Defensive rebounds, Assists, Steals, Turnovers, Personal fouls & NBA data from 5 regular seasons for training, 1 for scoring, NBA data from 800 games in 2015-2016 regular season, NBA data from 2011-2012 to 2014-2015 seasons \\ \hline

\textbf{Neural Networks} & \citet{osken2022predicting} & 74.33\%, 76\% & Accuracy & Field goals, 3-pointers, Free throws, Rebounds, Assists, Steals, Blocks, Turnovers, Personal fouls, Player efficiency rating, True shooting percentage, Effective field goal percentage & NBA 2007-2008 season data from ESPN, NBA regular season data from \url{www.basketballreference.com} and shot selection data from \url{www.nbaminer.com} \\ \hline

\textbf{Naive Bayes} & \citet{lin2014predicting}, \citet{miljkovic2010use}, \citet{jain2017machine} & 65.15\%, 67\%, 89.26\% & Accuracy & Points scored, FG attempts, Defensive rebounds, Assists, Turnovers, Record, Field goals, Three-pointers, Free throws, Rebounds, Assists, Turnovers, Steals, Blocks, Fouls, Points per game, Standings attributes like total wins/losses, home/away records, current streaks, 33 condition attributes including biometric data, college stats, draft order & NBA games from 1991-1998, NBA data from the 2009-2010 season, NBA data from 800 games in 2015-2016 regular season \\ \hline

\textbf{Ensemble Methods} & \citet{cai2019hybrid}, \citet{sukumaran2022application} & 84\%, 84\% & Accuracy & Two-point shots, Three-point shots, Free throws, Attack, Defense, Assists, Multiple features evaluated across various studies & CBA dataset (2016-2017 season), Various datasets evaluated \\ \hline

\textbf{Poisson Factorization} & \citet{ruiz2015generative} & - & Negative log-likelihood, Profitability & Team-specific and conference-specific attack and defense coefficients & Regular season results from over 5000 games to predict outcomes of the 2014 NCAA tournament games, data sourced from Kaggle and OddsPortal \\ \hline

\textbf{Heteroscedastic Models} & \citet{manner2016modeling} & - & Win probabilities & Team strengths, Home court advantage, Back-to-back game effects & Eight NBA seasons' data from 2006-2014 \\ \hline

\textbf{Markov Models} & \citet{vstrumbelj2012simulating} & - & Probabilistic forecasts & Possession-based game progression & Play-by-play data from the NBA 2007-08 and 2008-09 seasons \\ \hline

\end{tabular}
\end{table*}

\begin{figure*}
    \centering
    \includegraphics[width=0.79\textwidth]{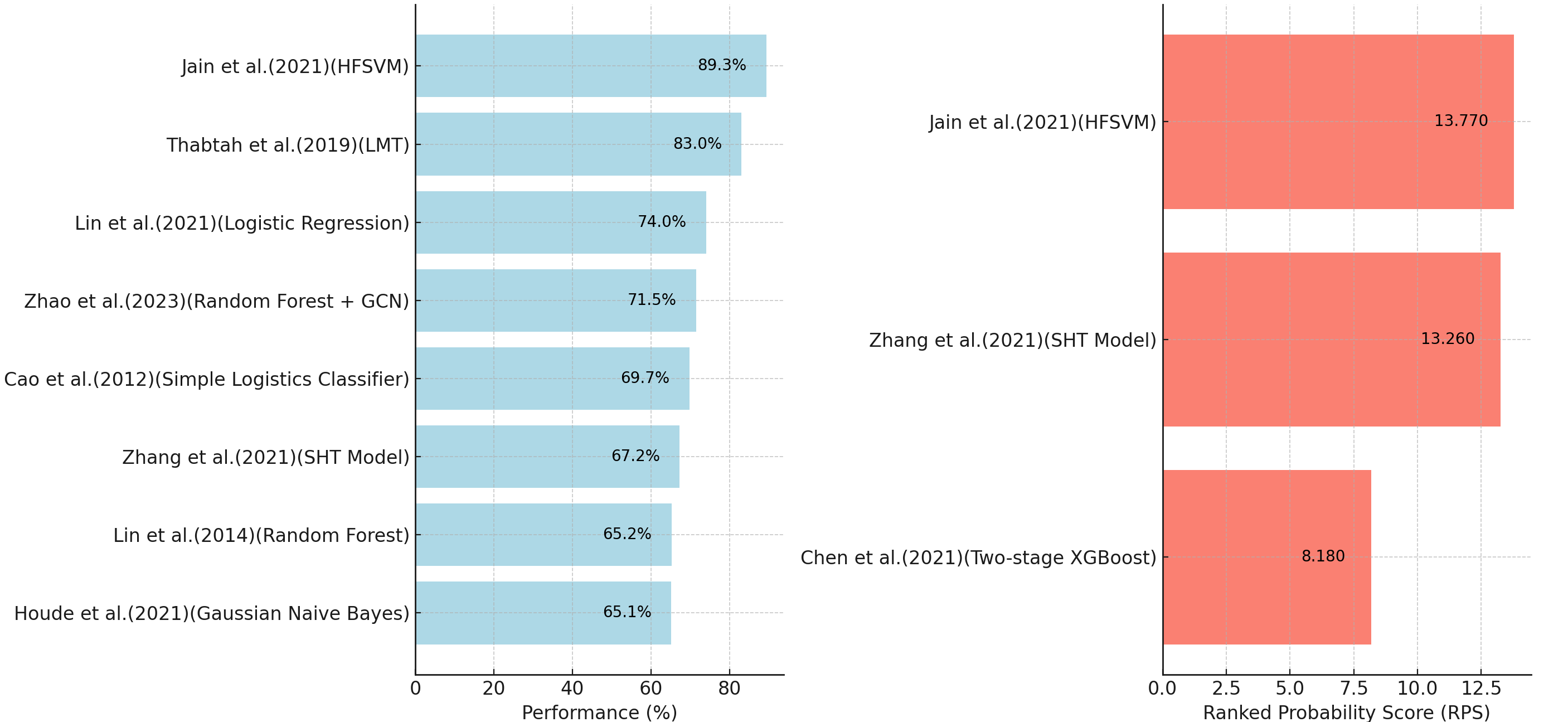}
    \caption{The best performances in basketball analytics based on accuracy, F1 and RPS}
    \label{fig:enter-label-soccer-performance-bk}
\end{figure*}

\citet{arndt2016predicting} used multitask regression to predict soccer player performances, evaluating models on five seasons of the German Bundesliga. The dataset included game event feeds and qualitative player grades.

\citet{elmiligi2022predicting} combined machine learning algorithms and statistical analysis to predict the outcomes of the match, achieving 46.6\% accuracy with a hybrid model. The dataset included 205,182 match results from the 2000/2001 to 2016/2017 seasons.

\citet{bilek2019predicting} analyzed situational variables and performance indicators in the EPL, highlighting the scoring of the first goal as the most influential factor. The dataset covered the 2017-2018 season.

\citet{pelechrinis2019evaluating} evaluated player contributions by measuring the expected impact of passes, using a Wyscout data set that covers 9,061 matches in major European leagues.

\citet{beal2021combining} combined machine learning and NLP with human insight to predict the outcomes of the EPL match, achieving 63.18\% accuracy. The dataset included match previews from The Guardian.

\subsection{Basketball \includegraphics[height=0.3cm]{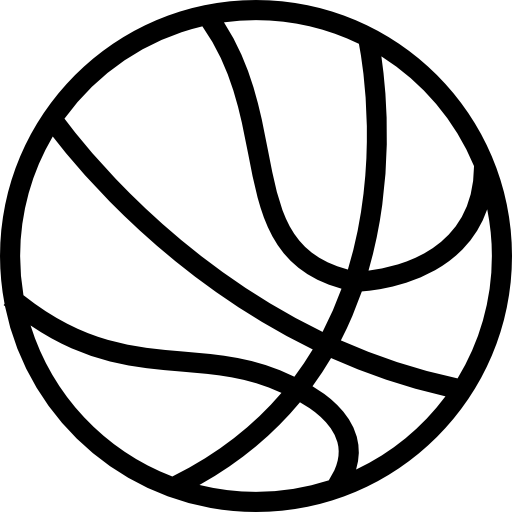}}\label{lbl_basketball_sec}

\begin{figure}[h!]
\fontsize{8}{8}\selectfont
\begin{tikzpicture}
    \begin{axis}[
        ybar,
        symbolic x coords={2012,2014,2015,2016,2018,2020,2021,2022,2023},
        xtick=data,
        x tick label style={rotate=45, anchor=east},
        ymin=0,
        xlabel={},
        ylabel={Number of articles},
        bar width=9.5pt,
        width=7.8cm,
        enlarge x limits=0.15,
    ]
        \addplot coordinates {(2012,1) (2014,2) (2015,1) (2016,2) (2018,2) (2020,1) (2021,3) (2022,2) (2023,3)};
    \end{axis}
\end{tikzpicture}
\caption{Histogram showing the number of articles per year in Basketball betting.}
    \label{fig:articles_per_year_basketball}
\end{figure}
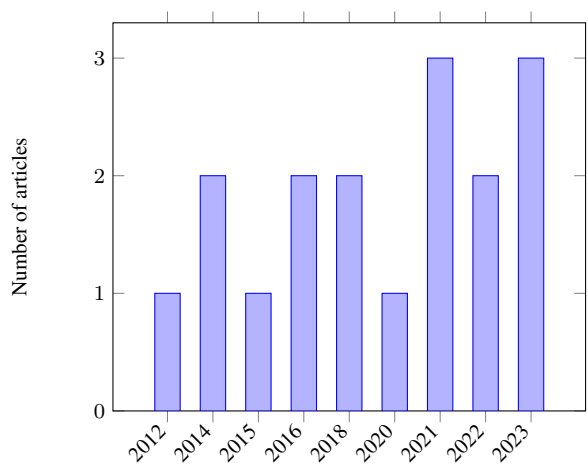

We synthesize the findings of various studies that developed machine learning models to predict the outcomes of basketball games, focusing on NBA games, NCAA tournaments, and international leagues. These studies used various methodologies and datasets to improve the accuracy of their predictive models, demonstrating the potential of machine learning in sports analytics (Table \ref{lbl_basketball} and Figures \ref{fig:enter-label-soccer-performance-bk} and \ref{fig:articles_per_year_basketball}).

Several studies utilized data mining techniques and machine learning algorithms to predict NBA game outcomes. \citet{chen2021hybrid} integrated five data mining techniques, including extreme learning machine, multivariate adaptive regression splines, k-nearest neighbors, eXtreme gradient boosting (XGBoost), and stochastic gradient boosting. The authors used data from the 2018-2019 NBA season and achieved the best performance with a two-stage XGBoost model using a four-game lag, resulting in a mean absolute percentage error (MAPE) of 0.0818. Similarly, \citet{cao2012sports} focused on predicting the outcomes of NBA games using algorithms such as the Simple Logistics Classifier, ANN, SVM and Naïve Bayes, the Simple Logistics Classifier achieving the highest accuracy of 69.67\%. Their dataset spanned five regular NBA seasons for training and one season for scoring.

In another study, \citet{lin2014predicting} employed box score statistics from NBA games between 1991 and 1998 to develop machine learning models to predict game winners, using algorithms such as Logistic Regression, SVM, AdaBoost, Random Forest and Gaussian Naïve Bayes. Their Random Forest algorithm achieved a prediction accuracy of 65.15\%, with logistic regression performing better in the final quartile of the season with an accuracy of 68.75\%. \citet{yeh2022evaluating} developed tools to measure the quality of continuously updated probabilistic forecasts, using Monte Carlo simulations on ESPN's real-time probabilistic forecasts of NBA games. They found ESPN's forecasts were generally well-calibrated, with a Brier score of 0.075, outperforming several naive models.

Further extending these findings, \citet{horvat2020impact} applied seven classification algorithms to predict NBA game outcomes using a dataset covering nine NBA seasons from 2009/2010 to 2017/2018. The k-Nearest Neighbors algorithm achieved the highest prediction accuracy of 60.82\% using up-to-date data. \citet{zhang2021modeling} constructed a multidimensional linear model (SHT) incorporating team strength, home court advantage, and player fatigue, achieving a win-loss accuracy of 64.63\% and an RMSE of 13.1185, with datasets from the NBA official website and other sources.

Furthermore, \citet{zhao2023enhancing} used graph convolutional networks (GCN) combined with Random Forest (RF) to predict the outcomes of NBA games, achieving an average prediction accuracy of 71.54\% using the ``NBA Enhanced Box Score and Standings (2012-2018)'' dataset from Kaggle. \citet{park2014prediction} investigated prediction models for NBA games using an extension of the Elo rating system, incorporating factors such as home/away performance and recent game outcomes, achieving improved fit and profitability in betting simulations.

The effectiveness of these models was further demonstrated by \citet{hu2004forecasting}, who utilized the weighted likelihood (WL) method to predict the outcome of NBA playoff games, specifically focusing on the 1996-97 finals between the Chicago Bulls and the Utah Jazz. The WL method showed competitive accuracy compared to logistic regression-based models, with datasets that included NBA game outcomes from the 1996-1997 season. \citet{zimmermann2016basketball} explored various models to predict the outcomes of basketball games using NCAAB and NBA data, finding that Naïve Bayes performed best for NBA predictions, particularly for playoff series.

In terms of predicting the success of NBA draft prospects, \citet{kannan2018predicting} evaluated logistic regression, SVM, and random forest models using a dataset from the 2009-2014 NBA draft combines, with the random forest model outperforming others. \citet{ruiz2015generative} presented a probabilistic model using modified Poisson factorization to predict NCAA tournament game outcomes, demonstrating superior profitability and prediction accuracy.

Studies also focused on in-game statistics and team performance metrics to improve prediction models. \citet{jones2016predicting} developed models using field-goal shooting percentage, three-point shooting percentage, free-throw shooting percentage, offensive rebounds, assists, turnovers, and attempted free throws, achieving accuracy of 88-94\%. \citet{lampis2023predictions} used machine learning algorithms on data from four European basketball tournaments, achieving accuracy up to 78\% in the Greek league.

\citet{lu2021improving} introduced adaptive weighted features to improve prediction accuracy in NBA games, achieving the best performance with stochastic gradient boosting (SGB) and a root mean square error (RMSE) of 11.5586. \citet{osken2022predicting} used clustering and artificial neural networks (ANNs) to predict NBA game outcomes, achieving prediction accuracies of approximately 76\%.

Other studies focused on the predictions for the NCAA tournament. \citet{ni2023comparative} used big data and machine learning classification models, with Random Forest achieving a prediction accuracy of 85.71\%. \citet{bennett2018comparing} evaluated neural networks, improved decision trees, and Naïve Bayes models to predict NCAA March Madness outcomes, with the neural network model performing best with a prediction accuracy of 64\%.

\begin{table*}[h!]
\centering
\caption{Summary of machine learning approaches for Tennis match prediction}
\label{lbl_tennis}
\fontsize{6.5}{6.5}\selectfont
\renewcommand{\arraystretch}{1.6}
\begin{tabular}{|>{\centering\arraybackslash}m{1.5cm}|>{\centering\arraybackslash}m{2cm}|>{\centering\arraybackslash}m{2cm}|>{\centering\arraybackslash}m{1.5cm}|>{\centering\arraybackslash}m{2.5cm}|>{\centering\arraybackslash}m{3cm}|}
\hline
\textbf{Approaches} & \textbf{Work} & \textbf{Performance} & \textbf{Metrics} & \textbf{Features} & \textbf{Datasets} \\
\hline
\textbf{Logistic regression} & \citet{sipko2015machine}, \citet{cornman2017machine}, \citet{wilkens2021sports}, \citet{gao2021random} & \textbf{Logistic Regression:} 70\% accuracy \cite{cornman2017machine}, 83.18\% accuracy \cite{gao2021random} & Probability of winning a point, ROI, logistic loss & Historical match and player data, point-level data, match-specific data & Wimbledon, OnCourt System, Jeff Sackmann's dataset, Tennis-Data.co.uk, Match Charting Project \\
\hline
\textbf{Artificial Neural Networks (ANN)} & \citet{sipko2015machine}, \citet{de2017using}, \citet{candila2020neural}, \citet{solanki2022prediction}, \citet{wilkens2021sports} & \textbf{ANN:} 82\% accuracy (Solanki), 70\% accuracy \cite{wilkens2021sports}  & Accuracy, F1 score, ROI, Brier Score & Statistical data, environmental data, player performance, serve statistics, recent performance & OnCourt System, ATP official website, Tennis-Data.co.uk, Jeff Sackmann's dataset, betting websites \\
\hline
\textbf{Random Forest} & \citet{gao2021random}, \citet{cornman2017machine}, \citet{solanki2022prediction}, \citet{wilkens2021sports} & \textbf{Random Forest:} 83.18\% accuracy \cite{gao2021random}, 73.5\% accuracy \cite{cornman2017machine}, 72\% accuracy \cite{solanki2022prediction}, 69\% accuracy \cite{wilkens2021sports} & Accuracy, ROI, precision, recall, F1 score & Serve strength, player height, age, match statistics, recent performance & ATP, Match Charting Project, Jeff Sackmann's dataset, Tennis-Data.co.uk \\
\hline
\textbf{Support Vector Machine (SVM)} & \citet{cornman2017machine}, \citet{ghosh2019comparison}, \citet{wilkens2021sports} & \textbf{SVM:} 70\% accuracy \cite{cornman2017machine}, 69\% accuracy \cite{wilkens2021sports} & Accuracy, RMSE, kappa statistic, recall, precision, F1 score & Player rankings, match statistics, recent performance & Jeff Sackmann's dataset, Tennis-Data.co.uk, UCI Tennis Match Statistics dataset \\
\hline
\textbf{Markov Chain Model} & \citet{knottenbelt2012common} & \textbf{Markov Chain:} 3.8\% ROI & Percentage of first serves, points won on serve, ROI & Serving and receiving statistics, common opponent statistics & ATP, bookmakers' odds \\
\hline
\textbf{Decision Tree (DT)} & \citet{ghosh2019comparison} & \textbf{Decision Tree:} 99.14\% accuracy & Accuracy, RMSE, kappa statistic, recall, precision, F1 score & Player rankings, match statistics & UCI Tennis Match Statistics dataset \\
\hline
\textbf{Fuzzy Inference System} & \citet{de2017using} & \textbf{Fuzzy Inference System:} Results not specified & Accuracy & Ranking points, historical performance, surface-specific win ratio & ATP official website \\
\hline
\textbf{Gradient Boosting Machine} & \citet{wilkens2021sports} & \textbf{Gradient Boosting Machine:} 70.5\% accuracy & Accuracy, precision, recall, specificity, F1 score, AUC, log-loss, Brier score & Player performance metrics & Tennis-Data.co.uk \\
\hline
\textbf{Learning Vector Quantization (LVQ)} & \citet{ghosh2019comparison} & \textbf{LVQ:} Accuracy not specified & Accuracy, RMSE, kappa statistic, recall, precision, F1 score & Player rankings, match statistics & UCI Tennis Match Statistics dataset \\
\hline
\end{tabular}
\end{table*}

\begin{figure*}[h!]
    \centering
    \begin{minipage}{0.5\textwidth}
        \centering
    \centering
    \includegraphics[width=0.99\textwidth]{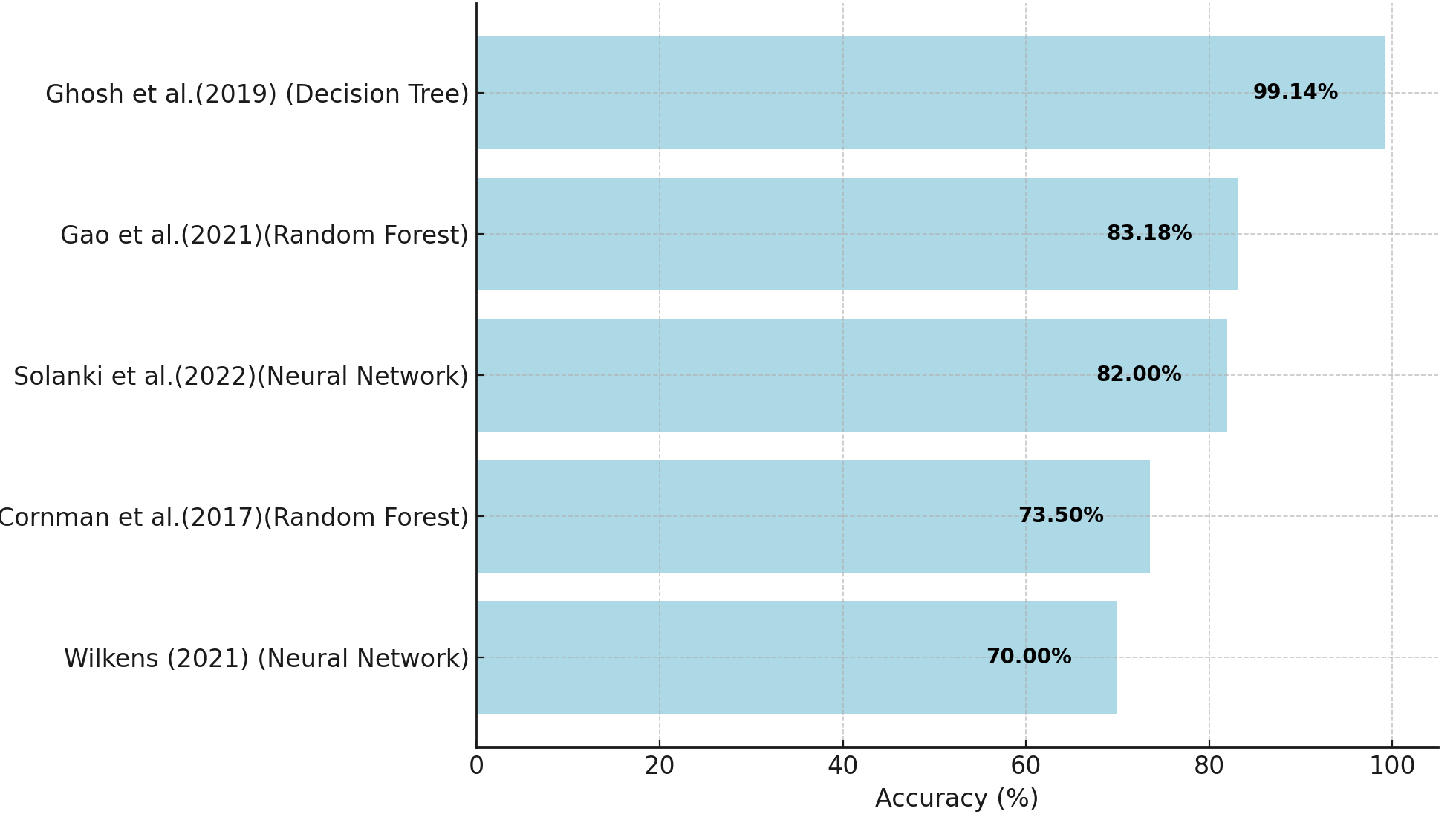}
    \caption{The best performances in Tennis analytics based on accuracy}
    \label{fig:enter-label-soccer-performance-tn}
    \end{minipage}\hfill
    \begin{minipage}{0.5\textwidth}
    \fontsize{8}{8}\selectfont
        \begin{tikzpicture}
    \begin{axis}[
        ybar,
        symbolic x coords={2012,2015,2017,2019,2020,2021,2022,2023},
        xtick=data,
        x tick label style={rotate=45, anchor=east},
        ymin=0,
        xlabel={},
        ylabel={Number of articles},
        bar width=9.5pt,
        width=7.8cm,
        enlarge x limits=0.15,
    ]
        \addplot coordinates {(2012,1) (2015,1) (2017,2) (2019,1) (2020,1) (2021,2) (2022,1) (2023,1)};
    \end{axis}
\end{tikzpicture}
\caption{Histogram showing the number of articles per year in Tennis betting.}
    \label{fig:articles_per_year_tennis}
    \end{minipage}
\end{figure*}

\subsection{Tennis \includegraphics[height=0.3cm]{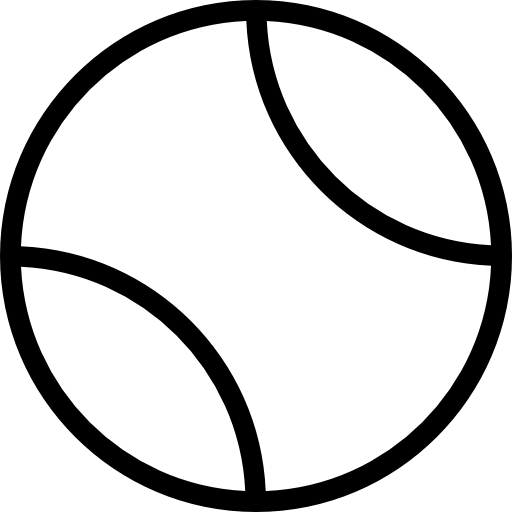}}\label{lbl_tennis_sec}

The prediction of tennis match outcomes has been extensively studied through various statistical and machine learning models, to take advantage of historical data and player statistics to make accurate predictions. Several methodologies have been developed, each demonstrating distinct approaches and performance metrics (Table \ref{lbl_tennis} and Figures \ref{fig:enter-label-soccer-performance-tn} and \ref{fig:articles_per_year_tennis}).

%{\citet{klaassen2003forecasting} proposed the \textit{TENNISPROB} model, which utilized a logit model to estimate the initial probability \( p_a \) of player A winning, based on world ranking differences. This model assumed points were independent and identically distributed (i.i.d.) and created dynamic probability profiles throughout the match by combining initial and match-specific probabilities. The model's efficacy was shown through historical Wimbledon finals, highlighting its ability to predict winning probabilities as matches progressed.}

%{Similarly, \citet{barnett2005combining} utilized a Markov chain model in Excel, which combined serving and receiving statistics, adjusted for match conditions such as surface type. This model was demonstrated using the 2003 Australian Open quarter-final match between Andy Roddick and Younes El Aynaoui, predicting a high probability of long sets due to the strong serving abilities of both players. Key metrics included the percentage of first serves in play and points won on serve, with the model achieving a high accuracy in predicting match outcomes.}

For instance, \citet{knottenbelt2012common} developed a hierarchical Markov model to predict the outcome of the matches by analyzing the proportion of service and return points won against common opponents. This model demonstrated a 3.8\% return on investment (ROI) when tested on a dataset of 2173 ATP matches from 2011. The efficacy of the hierarchical model was further validated using player statistics from their last 50 matches, indicating the model's potential to enhance returns from existing stochastic models. 

%{In another study, \citet{somboonphokkaphan2009tennis} employed a Multi-Layer Perceptron (MLP) neural network, incorporating statistical and environmental data. Three models---StatEnv, AdvancedStatEnv, and TimeSeries---were tested, with the TimeSeries model achieving the highest accuracy due to the inclusion of recent performance data. The StatEnv model showed a 75.59\% accuracy, while the AdvancedStatEnv model reached 79.53\% accuracy, underscoring the importance of integrating comprehensive datasets for improved predictions.}

Further advances were made by \citet{sipko2015machine}, who used logistic regression and artificial neural networks to predict match outcomes. By extracting 22 features from historical data, including factors such as fatigue and injury of the players, the models achieved a 4.35\% ROI, outperforming previous models. This approach highlighted the significance of detailed feature engineering in improving model performance.

\citet{cornman2017machine} utilized a dataset of 46,114 matches, merging data from Jeff Sackmann on GitHub and Tennis-Data.co.uk. Testing various models including logistic regression, SVM, neural networks and random forests, they found the random forest model to be the most effective, achieving training accuracy of 73.5\% and cross-validation accuracy of 69.7\%. This model yielded a profit of 3.3\% per match, emphasizing its potential in betting scenarios.

\citet{de2017using} explored different soft computing techniques such as ANN, Fuzzy Inference System, and Strength Equation. The ANN demonstrated the highest accuracy, particularly in three-set matches and Grand-Slam events. The dataset included ATP matches from the 2014 and 2015 seasons, and the combined voting system provided robust predictions.

\citet{gao2021random} employed a random forest model using ATP match data from 2000 to 2016. The model achieved 83.18\% prediction accuracy, significantly outperforming traditional betting odds. Serve-related variables were found to be the most impactful, suggesting that serve strength is a critical factor in match outcomes.

\citet{ghosh2019comparison} compared the performance of the Decision Tree (DT), Learning Vector Quantization (LVQ), and Support Vector Machine (SVM) classifiers on UCI benchmark datasets from Grand Slam tournaments. The Decision Tree model achieved an average accuracy of 99.14\%, outperforming other classifiers and demonstrating its efficacy in predicting match outcomes.

In another approach, \citet{candila2020neural} utilized ANNs to predict match outcomes, incorporating both previous and newly created variables. Their ANN model outperformed traditional models and achieved superior returns on investment, highlighting the economic profitability of neural networks in betting strategies.

\citet{solanki2022prediction} employed Neural Network, Linear Regression, and Random Forest algorithms on a dataset spanning tennis matches from 2000 to 2021. The Neural Network model achieved the highest accuracy of 82\%, demonstrating the effectiveness of machine learning in predicting the outcomes of tennis matches.

\citet{wilkens2021sports} analyzed various machine learning models in a dataset of ATP and WTA singles matches from 2010 to 2019. Despite no significant performance advantage over bookmaker-implied models, the study highlighted the robustness of machine learning models to capture match outcomes.

Finally, \citet{friligkos2023framework} utilized a dataset from the ATP Tour, developing a model with 17 features per player. The performance of the model was evaluated using holdout- and cross-validation methods, achieving a maximum accuracy of 71.95\% with holdout validation. The study introduced a web platform, Tennis Crystal Ball, for real-time match predictions.

\begin{table*}[h!]
\centering
\caption{Summary of approaches in Cricket prediction models}
\fontsize{6.5}{6.5}\selectfont
\renewcommand{\arraystretch}{1.6}
\begin{tabular}{|C{1.5cm}|C{2cm}|C{2cm}|C{1.5cm}|C{2.5cm}|C{2.5cm}|}
\hline
\textbf{Approaches} & \textbf{Work} & \textbf{Performance} & \textbf{Metrics} & \textbf{Features} & \textbf{Datasets} \\
\hline
\textbf{Decision Trees} & 
 \citet{kumar2018outcome},
\citet{vistro2019cricket},
\citet{bharadwaj2024player}
 & 

Decision Tree (55.1\%),
Decision Tree (94.87\%),
Decision Tree(97.45\%)
 & 
Precision, Recall, F1 Score, Accuracy & 
Past performance, ground, innings, venue, player performance & 
ESPNcricinfo \\ 
\hline

\textbf{MLP Networks} & 
\citet{kumar2018outcome},
 \citet{lamsal2018predicting}
 & 

MLP (57.4\%),
 MLP (71.66\%)
 & 
Precision, Recall, F1 Score, Accuracy & 
Past performance, ground, innings, venue, player performance, team strength & 
ESPNcricinfo, IPL official website \\ 
\hline

\textbf{Random Forest} & 

\citet{passi2018increased},
\citet{kamble2021cricket},
\citet{nimmagadda2018cricket}

 & 

Random Forest (92.25\%),
Random Forest (91\%),
Random Forest (improved accuracy)
 & 
Precision, Recall, F1 Score, AUROC, RMSE, Accuracy & 
Runs, wickets, overs, player statistics & 
ESPNcricinfo, Cricinfo website \\ 
\hline

\textbf{SVM} & 
\citet{jayanth2018team},
 \citet{manivannan2019convolutional},
 \citet{gour2024utilizing}
 & 

SVM with RBF kernel (75\%),
 SVM (feature encoding),
 SVM (high accuracy)
 & 
Precision, Recall, Accuracy & 
Player strengths, weaknesses, performances & 
\url{www.Howstat.com}, \url{www.espncricinfo.com} \\ 
\hline

\textbf{AdaBoost} & 

\citet{shenoy2022prediction}
 & 

AdaBoost (56.63\%),
 AdaBoost (62\%)
 & 
Accuracy & 
Batting and bowling statistics & 
\url{www.ESPNcricinfo.com} \\ 
\hline

\textbf{Linear Regression} & 

\citet{singh2015score},
 \citet{nimmagadda2018cricket}
 & 

Linear Regression (less error),
 Multiple Linear Regression (improved accuracy)
 & 
Error rates, Accuracy & 
Current run rate, number of wickets, venue, target score & 
ESPN Cricinfo, Cricinfo website \\ 
\hline

\textbf{Logistic Regression} & 

 \citet{nimmagadda2018cricket}
& 
 Logistic Regression (improved accuracy)
 & 
Error rates, Accuracy & 
Current run rate, number of wickets, venue & 
Cricinfo website \\ 
\hline

\textbf{K-means Clustering} & 

 \citet{shenoy2022prediction},
 \citet{sumathi2023cricket}
 & 

 K-means (62\%),
 K-means (high clustering accuracy)
 & 
Accuracy & 
Batting and bowling statistics, player performance characteristics & 
\url{www.ESPNcricinfo.com}, Kaggle \\ 
\hline

\textbf{XGBoost} & 

 \citet{vistro2019cricket},
 \citet{lamsal2018predicting}
 & 

 XGBoost (94.23\%),
 Extreme Gradient Boosting (high accuracy)
 & 
Accuracy & 
Player performance, weather conditions, venue, team strength & 
IPL official website, ESPNcricinfo \\ 
\hline

\textbf{CNN} & 

 \citet{manivannan2019convolutional}
 & 

 CNN (70\%)
& 
Accuracy & 
Player performances, team performance & 
\url{www.espncricinfo.com} \\ 
\hline
\label{cricket_review}
\end{tabular}
\end{table*}

\begin{figure*}
    \centering
    \includegraphics[width=0.99\textwidth]{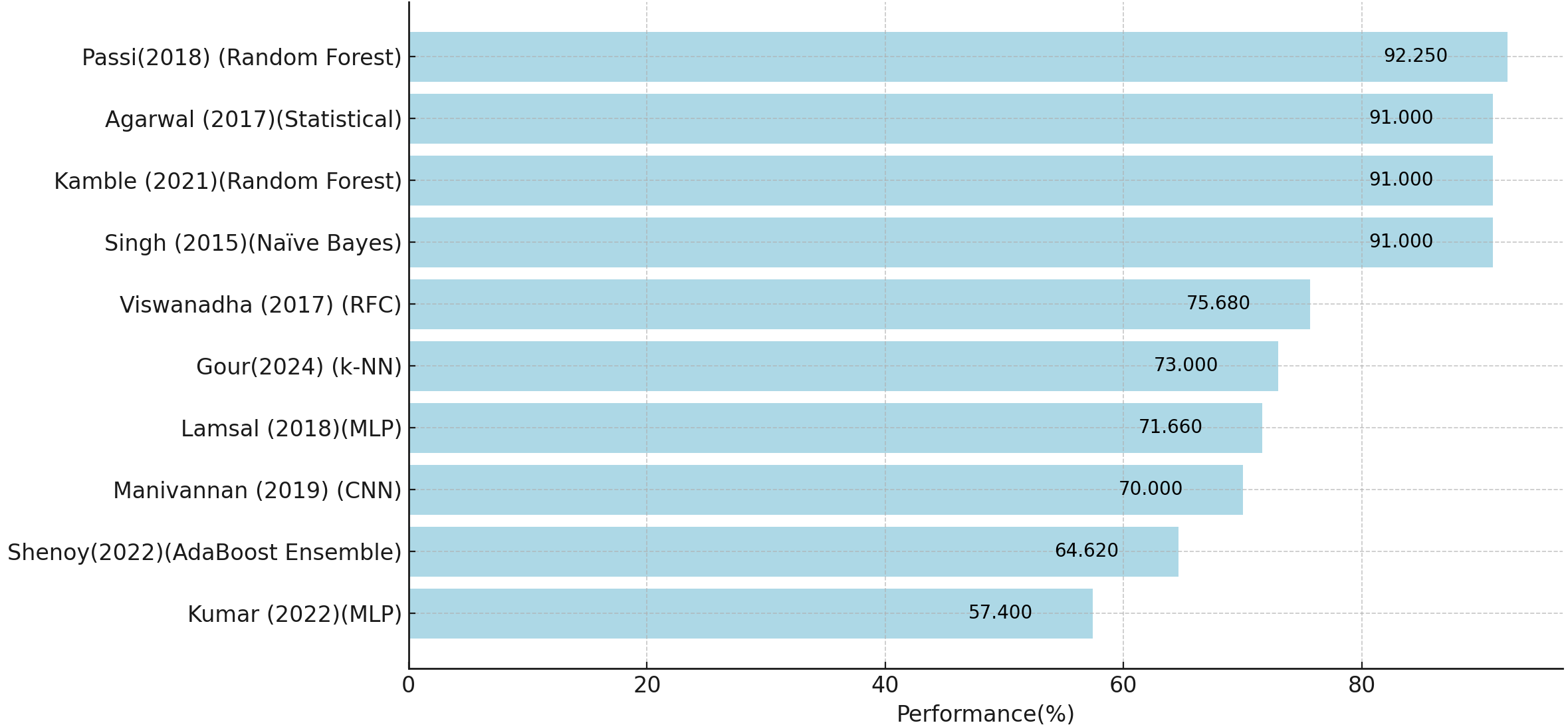}
    \caption{The best performances in Cricket analytics based on accuracy}
    \label{fig:enter-label-soccer-performance-crt}
\end{figure*}

\subsection{Cricket \includegraphics[height=0.3cm]{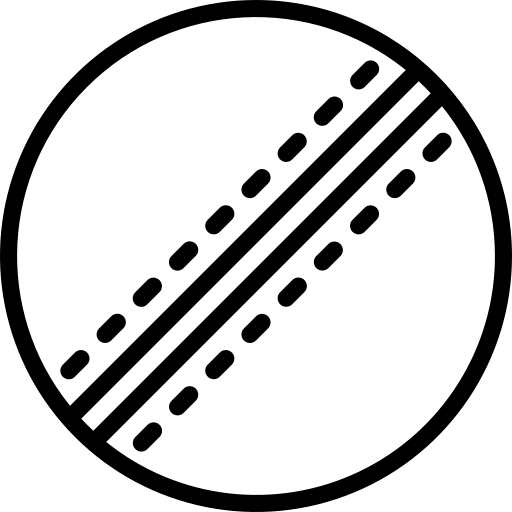}}\label{lbl_cricket}

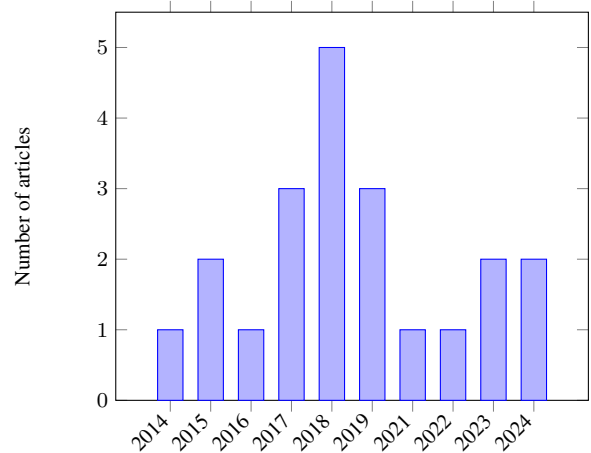
\begin{figure}[h!]
\fontsize{8}{8}\selectfont
\begin{tikzpicture}
    \begin{axis}[
        ybar,
        symbolic x coords={2014, 2015, 2016, 2017, 2018, 2019, 2021, 2022, 2023, 2024},
        xtick=data,
        x tick label style={rotate=45, anchor=east},
        ymin=0,
        xlabel={},
        ylabel={Number of articles},
        bar width=9.5pt,
        width=7.8cm,
        enlarge x limits=0.15,
    ]
        \addplot coordinates {(2014,1) (2015,2) (2016,1) (2017,3) (2018,5) (2019,3) (2021,1) (2022,1) (2023,2) (2024,2)};
    \end{axis}
\end{tikzpicture}
\caption{Histogram showing the number of articles per year in Cricket betting.}
    \label{fig:articles_per_year_cricket}
\end{figure}

In recent years, the application of machine learning techniques to predict outcomes in cricket has garnered significant attention. Several studies have explored different machine learning models to improve accuracy prediction for various cricket formats, including One Day Internationals, Twenty20 matches, and the Indian Premier League (Table \ref{cricket_review} and Figures \ref{fig:enter-label-soccer-performance-crt} and \ref{fig:articles_per_year_cricket}).

\citet{kumar2018outcome} conducted a study on the prediction of the outcomes of One Day International (ODI) cricket matches using Decision Trees and multilayer perceptron networks (MLP), using a dataset of 393 ODI matches played from January 5, 1971, to October 29, 2017, sourced from ESPNcricinfo. The dataset included features such as the past performance of the teams, the ground, innings and the venue (home / away / neutral). The dataset was cleaned to exclude tied matches, matches affected by rain and matches involving special teams. The study used both classifiers to analyze these features and found that the MLP network slightly outperformed Decision Trees in predicting match outcomes. The MLP achieved an accuracy of 57.4\% compared to the Decision Tree's 55.1\%. The precision, recall and F1 scores were calculated for both models: MLP had precision scores of 0.791 for India, 0.760 for Australia, and 0.767 for Pakistan, while Decision Trees had higher precision scores of 0.859 for India, 0.830 for Australia, and 0.789 for Pakistan. The study highlighted the potential of machine learning models to predict the outcomes of cricket matches and developed a desktop application called CricAI for practical implementation.

Similarly, \citet{shenoy2022prediction} employed multiple machine learning approaches to predict the results of T20 cricket matches, using data from 5,390 matches sourced from \url{www.ESPNcricinfo.com}. The extracted features included various bat and bowling statistics. The first approach involved normalizing 16 features for each player and experimenting with different aggregation levels, with the best accuracy (56.63\%) achieved using AdaBoost with aggregated player statistics. The second approach developed a new rating system that adjusted for the strength of opponents, employing stochastic gradient descent to optimize player ratings, achieving a maximum accuracy of 64. 62\% using a set of classifiers. The third approach used clustering of k-means to generate player ratings based on batting and bowling features, followed by a binary classifier, with AdaBoost achieving the highest accuracy at 62\%.

\citet{passi2018increased} employed supervised machine learning techniques to predict the performance of cricket players in one-day international (ODI) matches. They focused on predicting the number of runs a batsman would score and the number of wickets a bowler would take, treating both as classification problems. They used Naïve Bayes, Decision Trees, Random Forest, and Multiclass Support Vector Machine (SVM) classifiers, with Random Forest achieving the highest accuracy for both tasks. The study used data scraped from \url{www.cricinfo.com}, covering matches played between January 14, 2005, and July 10, 2017, for batting, and from January 2, 2000, to July 10, 2017, for bowling. The Random Forest classifier predicted runs with an accuracy of 90.74\% and wickets with an accuracy of 92.25\%, outperforming the other algorithms. Key performance metrics included precision, recall, F1 score, AUROC, and RMSE, with Random Forest consistently scoring highest on these metrics.

In addition, \citet{singh2015score} presented a model to predict scores and winning probabilities in one-day international (ODI) cricket matches using linear regression and Naïve Bayes classifiers. For the first innings, the model predicted the final score based on the current run rate, the number of wickets dropped, the venue and the batting team, with 5-over intervals recorded from matches played between 2002 and 2014. The Linear Regression classifier demonstrated less error compared to the traditional Current Run Rate method, making it more accurate in scoring prediction. In the second innings, the Naïve Bayes classifier estimated the probability of winning using similar attributes along with the target score, achieving an increase in accuracy from 68\% at the start to 91\% by the 45th over. The datasets for this study were obtained from ESPN Cricinfo, encompassing non-curtailed ODI matches of the top eight cricketing nations.

\citet{vistro2019cricket} applied various machine learning models to predict the winner of the Indian Premier League (IPL) matches using historical data from the 2008 to 2017 seasons. The authors used models that include Decision Tree, Random Forest, and XGBoost classifiers, using features such as player performance, weather conditions, and venue specifics. The dataset consisted of detailed match data from the IPL seasons, processed using the SEMMA methodology for data preparation and model assessment. The initial implementation of the Decision Tree classifier achieved an accuracy of 76.9\%, which improved to 94.87\% after parameter tuning. The Random Forest classifier initially provided 71\% accuracy, later enhanced to 80\% with parameter adjustments. XGBoost, without any tuning, delivered an accuracy of 94.23\%. These results demonstrate the potential of machine learning models in predicting match outcomes in cricket, highlighting the Decision Tree and XGBoost models as particularly effective.

Building on this, \citet{jayanth2018team} presented a supervised learning approach using Support Vector Machine (SVM) models with linear, nonlinear poly and RBF kernels to predict the outcomes of cricket matches based on players' strengths and weaknesses. The study utilized data from the 2011 Cricket World Cup, specifically \url{www.Howstat.com}, to create a player ranking index derived from batting and bowling statistics. The model divided the team into six divisions and calculated the features by subtracting the average ranking of the players in each division from the corresponding division of the opponent team. The experimental results indicated that the SVM with the RBF kernel outperformed others, achieving an accuracy of 75\%, a precision of 83. 5\%, and a recall rate of 62.5\%. Furthermore, the study proposed a system that recommends players for specific roles using k-means clustering and k-nearest neighbor (k-NN) classifiers, finding five similar players based on historical performance data.

Similarly, \citet{nimmagadda2018cricket} developed a predictive model for T20 cricket matches, particularly focusing on the Indian Premier League (IPL). They utilized Multiple Linear Regression to predict the first innings score by considering variables like the current run rate, the number of wickets fallen, and the venue of the match. For the second innings, Logistic Regression was used to predict outcomes and a random forest algorithm was applied to predict the winner of the match. The dataset for model training and validation was sourced from the Cricinfo website. The model demonstrated improved accuracy over traditional methods by incorporating additional predictive factors beyond just run rate, such as player performance statistics and match venue. The results showed that these combined methods provided a more reliable prediction of match outcomes, with the model performance validated through various statistical metrics such as accuracy and error rates.

Consequently, \citet{kamble2021cricket} developed a dual model system to predict the results and scores of cricket matches in ODI matches using machine learning. The first model predicted a team's final score after 50 overs based on the current match situation, while the second predicted win percentages for both teams before the match starts based on player selection and past performance data. The data set used spans the last five years for score predictions and the last 17 years for win predictions, sourced from ESPN. Various algorithms, including Naïve Bayes, Random Forest, multiclass SVM, and decision tree classifiers, were employed. The Random Forest classifier demonstrated the highest accuracy for both score prediction and win probability estimation. The key metrics considered included the number of wickets, current runs, overs, and player statistics such as run rate, strike rate, and bowling economy. The results indicated a significant improvement in predictive accuracy, with the Random Forest model achieving 91\% accuracy by the 42nd over.

Similarly, \citet{agarwal2017cricket} proposed a statistical modeling approach to predict the best cricket team lineup using Hadoop and Hive, focusing on Indian players. The model factored in various parameters such as players' overall stats, recent performances, opposition-wise stats, location-wise stats, and the last five performances. Batting and bowling scores were calculated with specific weights assigned to these factors, with a significant emphasis on recent performances. The model used these scores to predict the lineup of the team, achieving an accuracy of up to 91\% compared to the actual results. The data for the model were crawled from \url{www.sports.ndtv.com}, \url{www.thatscricket.com}, and \url{www.espncricinfo}. The study tested the model on matches from the England tour of India (ODI series) and the Australia tour of India (Test series) in early 2017, demonstrating the algorithm's effectiveness in predicting 14 out of 16 team members accurately for India vs. England and 13 out of 16 for India vs. Australia.

Furthermore, \citet{sumathi2023cricket} used machine learning techniques to predict and evaluate the performance of cricket players. Using a Kaggle dataset containing 2500 records and 15 attributes, the authors preprocessed the data to remove noise and missing values, resulting in 304 records for analysis. The study implemented three primary ML models: linear regression, K-means clustering, and random forest. Linear regression predicted individual player performance, regressing player attributes with a linear function to select relevant features. K-means clustering then grouped players into clusters based on similar performance characteristics, forming 13 clusters with cluster one containing the highest performing players. Finally, the random forest model validated the accuracy of these clusters and ranked players, producing a final list of the top 20 players. The experimental results, visualized through various graphs, demonstrated the effectiveness of the proposed system in accurately predicting player performance and aiding in team formation. The study's key metrics included player span, number of matches and innings, total runs, highest score, average run rate, and strike rate. The dataset used was named ``ICC Cricket'' from Kaggle.

In a similar vein, \citet{hatharasinghe2019data} explored various methodologies for predicting cricket match outcomes and examined two main approaches: using historical cricket data and using collective knowledge from social networks. The models evaluated included classification using team and player data, match simulation, and team composition-based methods. For example, \citet{kampakis2015using} used Naïve Bayes classifiers in English County Cricket data, achieving accuracy that exceeded the benchmarks of the betting industry. \citet{singh2015score} employed linear regression and Naïve Bayes classifiers on ODI data, achieving up to 91\% accuracy by the 40th over. \citet{sankaranarayanan2014auto} developed simulation models that provided the highest accuracy reported for ODIs. \citet{jhanwar2016predicting}'s team composition-based model achieved a 71\% accuracy by modeling individual player performances. \citet{mustafa2017predicting} demonstrated the use of social media sentiment analysis, achieving a theoretical profit of 67\% in CWC15 predictions. The datasets used included cricket data from ODI, IPL and English County Cricket data, with key metrics such as batting averages, strike rates, and sentiment scores playing crucial roles in the models.

Expanding on this concept, \citet{viswanadha2017dynamic} presented a dynamic model to predict the winner of the IPL Twenty20 cricket matches at the end of each match in the second innings, incorporating the context of the game and relative strengths of the team. The model evaluated various supervised learning algorithms, with the Random Forest Classifier (RFC) achieving the best results. Key metrics included runs remaining, wickets remaining, balls remaining, and relative team strengths, calculated based on individual player performance using both career and recent statistics. The dataset included career statistics from ESPN Cricinfo and ball-by-ball data from Cricsheet for IPL seasons 3-10. The RFC model showed an accuracy ranging from 65.79\% at the beginning of the second innings to 84.15\% by the 19th over, with an overall accuracy of 75.68\%.

\citet{ayub2023camp} introduced the Context-Aware Metric of player Performance (CAMP) for quantifying cricket players' contributions to matches, specifically limited over matches between 2001 and 2019. The CAMP model incorporated various contextual factors such as opponent strength, game situations, and player quality, using data mining techniques to provide a comprehensive performance metric. The empirical evaluation demonstrated that CAMP's ratings aligned with Man-of-the-Match decisions in 83\% of the 961 matches analyzed, outperforming the traditional Duckworth-Lewis-Stern (DLS) method. The dataset used for this study comprised ball-by-ball data from 961 One Day International (ODI) matches.

Complementing these findings, \citet{manivannan2019convolutional} proposed two novel approaches to predict the outcome of cricket matches by modeling team performance based on player performance. The first approach utilized feature encoding, assuming different categories of players and modeling teams as sets of player-category relationships, and then trained a linear SVM classifier on these relationships. The second approach employed a shallow Convolutional Neural Network (CNN) with four layers to learn an end-to-end mapping between player performances and match outcomes. The dataset, collected from \url{www.espncricinfo.com}, included 2,581 players and 474 ODI matches from May 2013 to October 2017. The experiments showed that the CNN-based approach achieved over 70\% accuracy, outperforming the feature encoding-based method and the baseline approaches. Various player attributes such as the number of matches played, runs scored, and captured wickets were used as features. The CNN architecture was designed to avoid capturing player-order information, focusing instead on extracting discriminative features through convolution and average pooling layers. Adding temporal information on player performances improved the model, considering performances from all previous seasons with higher weights for recent seasons.

\citet{bharadwaj2024player} predicted cricket player performance using machine learning, focusing on batting and bowling performance. They used four supervised learning algorithms: Naïve Bayes, Decision Tree, Random Forest, and SVM. The data was sourced from ESPNcricinfo, including various attributes of the players such as runs scored, wickets taken, and performance against specific opponents and venues. Decision Tree and Random Forest were the most accurate, achieving the highest prediction accuracy with Decision Tree scoring up to 97.45\% for batting and 94.90\% for bowling. The primary dataset used was ESPNcricinfo. The study highlighted the effectiveness of ensemble methods like Random Forest for handling high-entropy data and complex patterns in sports analytics. The metrics used included accuracy percentages for different training-test splits and weighted attribute values determined through the Analytic Hierarchy Process (AHP).

Continuing in a similar vein, \citet{gour2024utilizing} developed a predictive model for the outcomes of the IPL-T20 cricket matches using machine learning techniques, examining player statistics, match dynamics, and historical data. The study used SVM, Random Forest, Logistic Regression, Decision Tree, and k-NN models to predict player performance, using form, fitness, and historical data as features. The dataset comprised 340 matches and profiles of 154 players sourced from platforms such as Bet365, Cricbuzz, and ESPN. Their multi-model approach achieved the highest accuracy of 0.73, and the k-NN algorithm (k = 4) proved the most effective among individual models. The key metrics evaluated included accuracy, precision and recall, with relative team strength, player form, and opponent pairings identified as crucial predictive features.

In parallel, \citet{lamsal2018predicting} predicted the outcomes of the Indian Premier League (IPL) matches using machine learning. They identified seven key factors influencing match outcomes, including team strength, which was calculated based on the performance of top players using a multivariate regression model. Data was sourced from the IPL's official website, resulting in a dataset containing 634 matches. The models tested included Naïve Bayes, Extreme Gradient Boosting, Support Vector Machine, Logistic Regression, Random Forests, and Multilayer Perceptron (MLP). The MLP model, a three-hidden-layer neural network, outperformed others with an accuracy of 71.66\%, correctly predicting the outcome of 43 out of 60 matches in the 2018 IPL season. The metrics used to evaluate performance included precision, recall and the F1 score, with the MLP achieving a weighted mean F1 score of 0.72.

\begin{table*}[h!]
    \centering
    \caption{Summary of approaches in predicting American Football play calls and outcomes}
    \label{lbl_nfl}
    \centering
    \fontsize{6.5}{6.5}\selectfont
    \renewcommand{\arraystretch}{1.6}
    \begin{tabular}{|C{2cm}|C{2cm}|C{2cm}|C{1.5cm}|C{2.5cm}|C{2.5cm}|}
    \hline
\textbf{Approaches} & \textbf{Work} & \textbf{Performance} & \textbf{Metrics} & \textbf{Features} & \textbf{Datasets} \\
\hline
        \textbf{Hidden Markov Models (HMMs)} & \citet{otting2021predicting} & Out-of-sample prediction accuracy: 71.5\% (highest: 77.9\%, lowest: 60.2\%) & Accuracy, Precision, Recall & Game location, Yards to go, Down number, Formation, Score difference, Field position & Kaggle play-by-play dataset (2009-2017) \\ \hline
        \textbf{CNN-LSTM} & \citet{cheong2021prediction} & Reduced prediction errors in defender movements & RMSE, Trajectory metrics & Sensor data from RFID tags, Player movements & NFL Next Gen Stats (2018-2019) \\ \hline
        \textbf{LSTM, GRU, ANN, MLSTM-FCN} & \citet{skoki2021ml} & Best recall: 0.884 (LSTM), Precision: 0.075, F1 score: 0.138, AUC: 0.807 & Recall, Precision, F1 score, AUC & GPS tracking data, Time-series features & NFL Big Data Bowl (2018) \\ \hline
        \textbf{Logistic Regression, SVM, Neural Networks} & \citet{noldin2020predicting} & Highest accuracy: 87.9\% (SVM) & Accuracy & Game time, Distance to goal line, Score differential, Passing percentage & NFL play-by-play data (2009-2019) \\ \hline
        \textbf{Neural Networks} & \citet{joash2020predicting} & Accuracy: 75.3\%, False negative rate: 10.6\% & Accuracy, False negative rate & Play-by-play data, Madden NFL ratings & NFL regular seasons (2013-2017) \\ \hline
        \textbf{LSTM Recurrent Neural Network} & \citet{yurko2020going} & Superior performance in predicting expected yards & Expected Points (EP), Win Probability (WP) & Player positions, Trajectories, Ball location & NFL Big Data Bowl (2017) \\ \hline
        \textbf{Elastic Net, Random Forest, XGBoost} & \citet{patel2023predicting} & Best performance: XGBoost, Validation accuracy: 58.5\% & RMSE, Accuracy & Game statistics, ESPN power ranks, Elo ratings & NFL seasons (2016-2021) \\ \hline
        \textbf{Logistic Regression, Random Forest, SVM} & \citet{juuri2023predicting} & Best accuracy: 95.61\% (SVM) & Accuracy, RMSE, AUC & Team statistics, Turnovers, Fourth-down attempts & Pro Football Focus (latest two seasons) \\ \hline
        \textbf{Gaussian Process Model} & \citet{warner2010predicting} & 64.36\% game winner prediction & Accuracy, Error rate & Offensive and defensive stats, Temperature difference & Pro-Football-Reference (2000-2009) \\ \hline
        \textbf{Feedforward Neural Network} & \citet{suyerpredicting} & Varying degrees of accuracy & Loss functions, Error terms & Game statistics (total first downs, offensive yards) & \url{www.NFLsavant.com} (various seasons) \\ \hline
        \textbf{Mixture Models} & \citet{dutta2020unsupervised} & High accuracy in identifying coverage types & Probabilistic assignments & Variance in coordinates, Distance to nearest players & NFL Big Data Bowl (2017) \\ \hline
        \textbf{Logistic Regression, CCA} & \citet{sinha2013predicting} & Over 55\% accuracy for WTS predictions & Accuracy, Precision, Recall, F-measure & Tweet features, Game statistics & Twitter and \url{www.NFLdata.com} (2010-2012) \\ \hline
        \textbf{Feedforward Artificial Neural Network} & \citet{burke2019deepqb} & Targeted receiver prediction: 60\%, Completion rate: 74\% & Accuracy, Completion rate & Player positions, Velocities, Play metadata & NFL Next Gen Stats (2016-2017) \\ \hline
        \textbf{C4.5, ANN, Random Forest} & \citet{roumani2023sports} & Best performance: Random Forest, Accuracy: 84\% & Recall, AUC, Specificity & Regular season data, Oversampling with SMOTE & NFL seasons (2002-2019) \\ \hline
        \textbf{Multiple Linear Regression} & \citet{morgan2024forecasting} & Accuracy: 64.41\% for game winners, 56.78\% against spread & Adjusted R², Win percentage & Historical performance data, Significant predictors & NFLfastR package (2002-2023) \\ \hline
    \end{tabular}
    \label{nfl_review}
\end{table*}

\begin{figure*}
    \centering
    \includegraphics[width=0.779\textwidth]{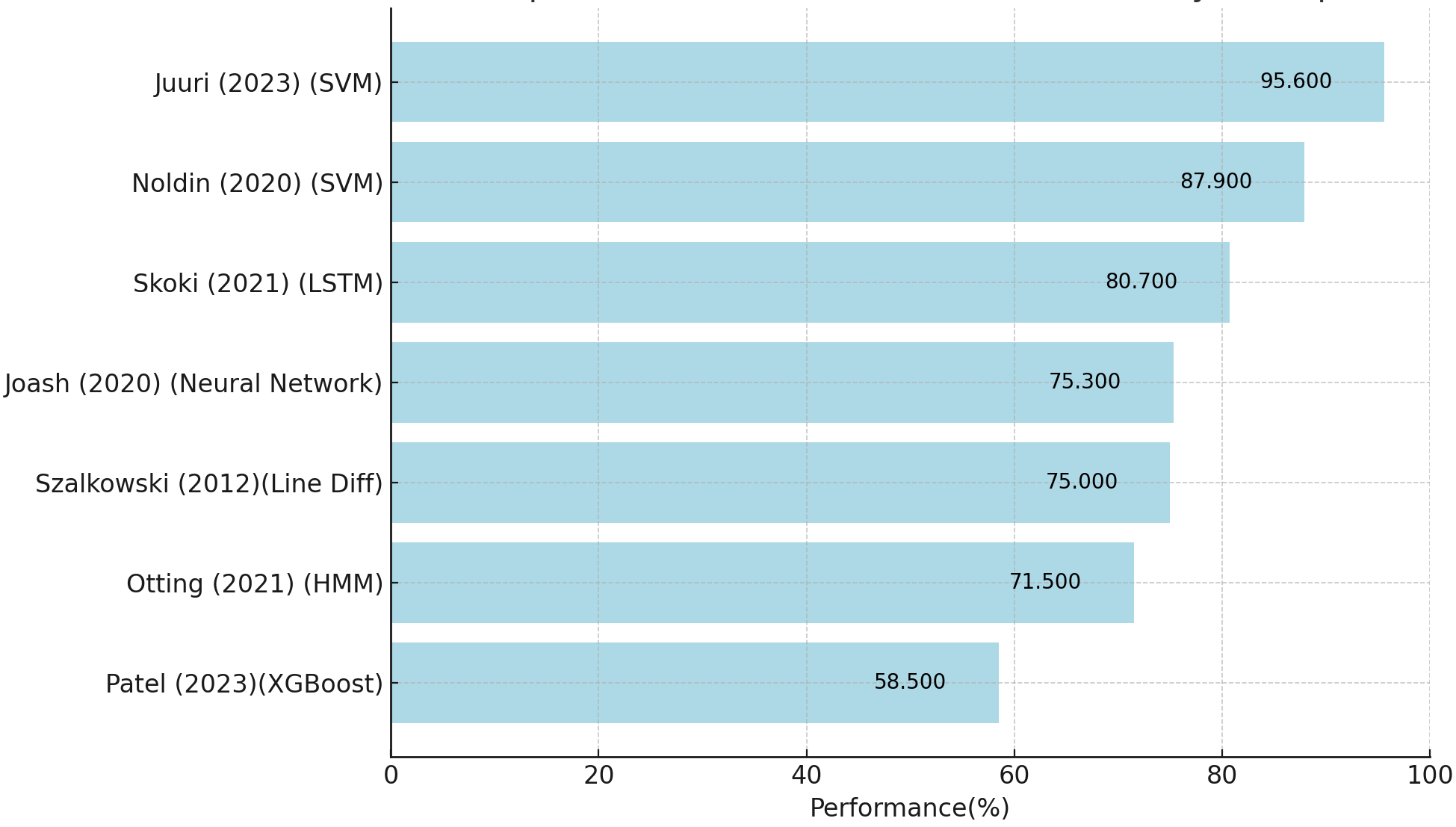}
    \caption{The best performances in American football analytics based on accuracy}
    \label{fig:enter-label-soccer-performance-nfl}
\end{figure*}

\subsection{American Football \includegraphics[height=0.3cm]{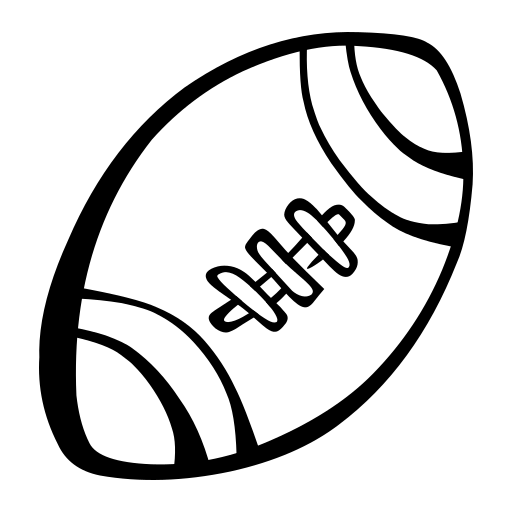}}\label{lbl_american_football}

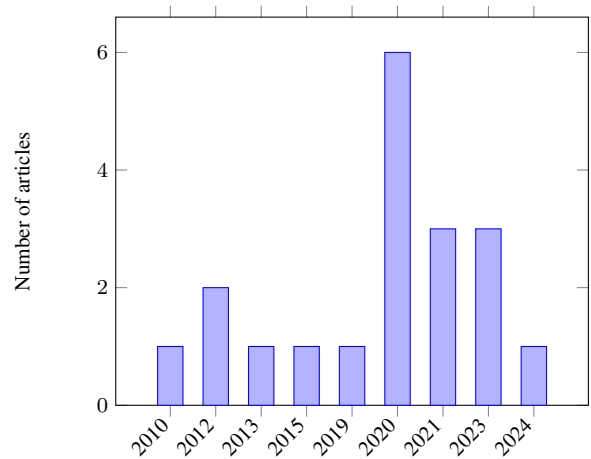
\begin{figure}[h!]
\fontsize{8}{8}\selectfont
\begin{tikzpicture}
    \begin{axis}[
        ybar,
        symbolic x coords={2010, 2012, 2013, 2015, 2019, 2020, 2021, 2023, 2024},
        xtick=data,
        x tick label style={rotate=45, anchor=east},
        ymin=0,
        xlabel={},
        ylabel={Number of articles},
        bar width=9.5pt,
        width=7.8cm,
        enlarge x limits=0.15,
    ]
        \addplot coordinates {(2010,1) (2012,2) (2013,1) (2015,1) (2019,1) (2020,6) (2021,3) (2023,3) (2024,1)};
    \end{axis}
\end{tikzpicture}
\caption{Histogram showing the number of articles per year in American football.}
    \label{fig:articles_per_year_AmFoot}
\end{figure}

Predicting outcomes in American football has been widely explored through the use of statistical and machine learning models, utilizing historical game data and player performance statistics to improve accuracy. Several approaches have been developed, each showcasing different methods and evaluation metrics. These models assess factors such as team dynamics, player stats, game conditions, and advanced analytics to predict game outcomes with high accuracy (Table \ref{nfl_review} and Figures \ref{fig:enter-label-soccer-performance-nfl} and \ref{fig:articles_per_year_AmFoot}).

For instance, \citet{otting2021predicting} used hidden Markov models (HMMs) to predict NFL play calls. Using a comprehensive play-by-play dataset from Kaggle, which includes 289,191 observations from regular season matches between 2009 and 2017, the study focused on predicting plays for the 2018 season. HMMs were chosen due to their ability to account for the time series structure of play call data, improving predictive power by modeling the team's propensity to pass or run. Key covariates included game location (home/away), yards to go, number of down, formation (shotgun / no huddle), score difference, and field position. The model achieved an out-of-sample prediction accuracy of 71.5\%, which was significantly higher than similar studies without player-specific data. The prediction accuracy varied between teams, with the New England Patriots achieving the highest accuracy at 77.9\% and the Seattle Seahawks the lowest at 60.2\%. The metrics used for the evaluation included prediction accuracy, precision, and recall for both pass and run plays, highlighting the robust performance of the model compared to previous approaches.

In a related vein, \citet{cheong2021prediction} developed a model to predict the defensive player trajectories in NFL games using a Defender CNN-LSTM approach. The model processed sensor data from RFID tags on players and the ball, focusing on passing plays. It used LSTM networks to handle sequence prediction, combined with 1D Convolutional Neural Networks (CNN) for feature extraction. The model accounted for what-if scenarios by predicting trajectories based on changes in the targeted receiver. Traditional regression metrics such as RMSE were used, but new trajectory metrics were developed for better evaluation. The results showed that the Defender CNN-LSTM model outperformed other models, reducing prediction errors and improving the accuracy of predicting defender movements. The dataset used was NFL Next Gen Stats (NGS) data from the 2018 and 2019 seasons.

Similarly, \citet{skoki2021ml} presented a machine learning-based approach to predict Defensive Pass Interference (DPI) in the NFL using GPS tracking data from NFL Next Gen Stats for the 2018 season, acquired through the NFL Big Data Bowl 2021 on Kaggle. The study used highly imbalanced time series binary classification models that included Long-Short-Term Memory (LSTM), Gated Recurrent Unit (GRU), Attention Neural Network (ANN) and Multivariate LSTM-FCN (MLSTM-FCN). Despite various pre-processing steps and model adjustments, the models exhibited limited success with high recall but low precision, indicating many false positives. The LSTM model with 128 hidden neurons achieved the best recall of 0.884 but had a precision of only 0.075, leading to an F1 score of 0.138 and an AUC of 0.807. The ANN and GRU models had similar recall but varied in precision and F1 scores. The imbalanced nature of the dataset, with only 259 DPI events out of 17,703 actions (1.46\%), significantly challenged model performance, suggesting that GPS data alone are insufficient for accurate DPI prediction. The study concluded that integrating these models as a pre-processing step for video analysis could enhance the accuracy of the DPI prediction.

Moreover, \citet{noldin2020predicting} investigated the feasibility of predicting play types in American football using machine learning. The study mainly focused on binary classification between pass and run plays and extended to predicting pass depth and run location. The research used logistic regression, support vector machine (SVM), and neural network models, with the SVM achieving the highest prediction accuracy of 87.9\%. The essential predictions features were identified through an interview with Max Sommer, the coach's consultant, and included the remaining game time, the distance to the goal line, the distance to a new first down, the scoring differential and the passing percentage of the offensive team for each down. The models were trained separately for each down to capture the distinct characteristics of play-calling strategies. The dataset used for training was obtained from \url{www.NFL.com}'s publicly available play-by-play data from 2009 to 2019, scraped and made accessible by Ronald Yurko on GitHub.

Furthermore, \citet{joash2020predicting} developed a machine learning model to predict NFL plays (pass vs. rush) using data from the 2013–2017 NFL regular seasons. They compared several models, finding that a neural network achieved the highest accuracy of 75.3\% with a false negative rate of 10.6\%. To balance accuracy and interpretability, they created a decision tree model that retained 86\% of the neural network's accuracy (65.3\%) and was practical for in-game use. They extended their analysis to team-specific decision trees, with accuracies ranging from 64.7\% to 82.5\%, using play-by-play data and Madden NFL video game ratings. The study emphasized creating a model that coaches can easily implement, helping real-time decision making, and improving defensive strategies.

In a related study, \citet{yurko2020going} developed a framework for the continuous-time within-play evaluation of game outcomes in the NFL using player tracking data. They employed a long-short-term memory (LSTM) recurrent neural network to estimate the expected yards a ball carrier would gain from their current position, factoring in the locations and trajectories of all players on the field. Their model incorporated modular submodels to handle different aspects of within-play events. The primary dataset used was player and ball tracking data from the NFL’s ``Big Data Bowl'' competition, covering the first 6 weeks of the 2017 regular season. The key metrics used included expected points (EP) and win probability (WP), with continuous updates throughout each play. The results demonstrated the superior performance of the LSTM model in predicting the expected end-of-play yard line, significantly improving over baseline models.

In a different approach, \citet{patel2023predicting} utilized machine learning models to predict the outcomes of the NFL game in terms of point spreads. Patel compared three different models: Elastic Net Linear Regression, Random Forest, and XGBoost, using various features derived from game statistics, ESPN power ranks, and FiveThirtyEight Elo ratings. The dataset included historical spreads, team statistics and Elo ratings from the 2016 to 2021 NFL seasons, with a focus on a rolling average of four weeks for feature engineering. The XGBoost model with top correlated variables and their quadratic terms achieved the best performance, with a cross-validated RMSE close to the set spread and a validation accuracy of 58.5\%. The model predicted the 2021 season outcomes with 53.65\% accuracy, indicating profitability at 10 to 11 betting odds. Key metrics included RMSE and accuracy, with the most influential features being the current week's spread and related Elo ratings. The dataset utilized for this study was the combined data from Sports Odds History, ESPN, and FiveThirtyEight.

%{Additionally, \citet{boulier2003predicting} evaluated the predictive accuracy of power scores for NFL games from 1994 to 2000 using probit regression models. They compared these predictions with naive models, the betting market, and the New York Times sports editor's forecasts. The dataset utilized included power scores published in the New York Times, based on variables like won-loss records, point margins, home-field advantage, and the quality of opponents. Results indicated that the betting market was the most accurate predictor, followed by probit models based on power scores. Predictions based solely on power scores were correct about 60\% of the time, slightly better than the sports editor’s forecasts. Home-field advantage was a significant factor, improving the accuracy of predictions. The Brier scores for various methods indicated that while power scores have predictive value, they were not as effective as the betting market. Bootstrapping the sports editor’s forecasts showed that statistical models could slightly outperform his predictions, but the overall difference was minimal.}

\citet{szalkowski2012performance} analyzed the performance of NFL betting lines in predicting game outcomes using data from 2560 NFL games between 2002 and 2011. The model used involved examining the difference between the opening and closing betting lines (line difference) and comparing it with the margin of victory (MOV) in the games. The authors found that the line difference could predict divisional winners with at least 75\% accuracy for ``straight up'' predictions. Home teams only beat the spread 47\% most of the time, but betting on home underdogs produced a winning strategy of 53.5\%, surpassing the 52.38\% threshold needed to break even. Principal Component Analysis (PCA) of box score data revealed that the value of the betting line consistently ranked highly as a predictor. A chi-squared goodness-of-fit test showed that the line difference followed a normal distribution. The study concluded that while the betting line was a good predictor of straight-up wins, it was less accurate against the spread. The dataset used was a compilation of NFL box scores and betting lines for all games from 2002 to 2011.

In another study, \citet{juuri2023predicting} utilized data from Pro Football Focus covering scores and team statistics from the two most recent NFL seasons. The study employed three machine learning models: Logistic regression, Random Forest Classifier (RFC), and SVM to predict game outcomes. The Logistic Regression model achieved an accuracy of 85.96\%, highlighting the significance of variables such as turnovers and fourth-down attempts. The RFC model, which included all variables without multicollinearity issues, achieved an accuracy of 90.35\% and identified passer ratings and rushing attempts as crucial factors. The SVM model, which was robust against outliers, achieved the highest accuracy at 95.61\%, with significant variables including passing touchdowns and turnovers. The study concluded that machine learning can effectively predict the outcomes of NFL games, with the most influential variables differing between models.

Complementing these findings, \citet{uzoma2015hybrid} developed a model that combines linear regression and k-NN algorithms to predict the outcomes of the NFL game. They used data mining techniques to extract features from NFL games statistics, using RapidMiner and Java for backend processing. The hybrid model leveraged high attribute weights from the LR model to inform the k-NN model, focusing on unique features like bookmaker betting spreads and player performance metrics. Tested on NFL games from weeks 16 and 17 of the 2013 season, the model achieved a prediction accuracy of 80.65\%, outperforming existing systems with accuracies ranging from 59.4\% to 67.08\%. The dataset used for this study was the NFL games statistics for the 2013 season.

In conjunction with this, \citet{hsu2020using} utilized machine learning techniques and candlestick chart patterns derived from betting market data to predict the outcome of the NFL game. The research explored both classification (win/loss) and regression (winning/loss margin) models. Various machine learning methods such as ensemble learning, support vector machines, and neural networks were applied. The dataset consisted of 13,261 instances over 32 NFL seasons. The random subspace regression method achieved the highest accuracy at 68.4\%. The key metrics used for the evaluation included accuracy, precision, recall, F measure, root mean square error (RMSE), and mean absolute error (MAE). The study concluded that candlestick charts based on betting market data can effectively predict game outcomes and suggested further exploration of these methods for individual team behavior analysis. The dataset utilized was sourced from \url{www.covers.com}.

Similarly, \citet{warner2010predicting} used a Gaussian process model to predict the margin of victory in NFL games and compared these predictions against the Las Vegas line. The dataset included a variety of offensive and defensive statistics from approximately 2,000 NFL games between 2000 and 2009. Additional features considered included the temperature difference between competing teams' cities and the computed strength of a team. The Gaussian process model achieved prediction accuracy for the margin of victory with an error rate only 2\% higher than the Las Vegas line and successfully picked game winners 64.36\% of the time. The proposed betting scheme, based on these predictions, resulted in a win rate just under 51\%, falling short of the 52.4\% needed to break even in the NFL gambling system. The final feature set for the model included the winning percentages of the home and away teams and the calculated strengths. The study highlighted that while the model showed promise, Las Vegas line-makers still had a slight edge in prediction accuracy. The primary dataset used for the training was obtained from \url{www.Pro-Football-Reference.com}.

On a different note, \citet{ajmeri2012using} introduced a computer vision and machine learning approach to classify NFL game film and develop a player tracking system using a manually captured dataset from Washington Redskins home games. The model utilized several machine learning techniques including Classification and Regression Trees (CART), Naïve Bayes, SVM, k-NN, and Logistic Regression to classify formations and player positions, achieving the highest accuracy with CART at 72.3\% for formation prediction. The metrics used included accuracy, precision, and recall, with specific results showing that CART achieved 86.5\% accuracy for the classification of the quarterback position. The system automated the tagging of formations, player routes, and speeds, enhancing scouting and game planning by analyzing player locations and movements. The results indicated an improvement in the efficiency of the scouting and provided detailed information on player performance and coaching tendencies, with potential applications that extend to various levels of football.

\citet{suyerpredicting} used a neural network model, specifically a feedforward neural network, constructed and executed with TensorFlow and Keras. The model was trained using extensive NFL data that encompass various game statistics, such as total first downs, offensive yards, and turnovers, obtained from \url{www.NFLsavant.com}. The study highlighted the significance of tweaking network parameters, such as activation functions, input/output nodes, and hidden layers, to improve prediction accuracy. Suyer employed a method of cross-validation to prevent overfitting and improve the model's generalization to new data. The model achieved varying degrees of accuracy across different prediction tasks, but specific accuracy rates and detailed results were not provided in the summarized content. The metrics used for the evaluation included typical neural network loss functions and error terms adjusted during the training process to optimize weights for the most accurate data analysis possible.

In addition, \citet{dutta2020unsupervised} proposed a model to automatically annotate the types of pass coverage of defensive backs using NFL player tracking data. The model utilized mixture models in an unsupervised learning framework to distinguish between man- and zone-coverage types. They generated a comprehensive set of features from the tracking data, such as the variance in the x- and y-coordinates, the mean distance to the nearest offensive and defensive players, and the variance in the direction of motion. The dataset used was the NFL's player and ball tracking data from the NFL's Big Data Bowl, which included game data from the first six weeks of the 2017 NFL season. The results showed that the mixture models could successfully identify coverage types with high accuracy, providing probabilistic assignments to each play.

Similarly, \citet{sinha2013predicting} explored the predictive power of social media data, specifically Twitter, in forecasting the outcomes of NFL games. They utilized tweets related to NFL teams and games from the 2010-2012 seasons, collected using Twitter's ``garden hose'' stream, combined with traditional game statistics from \url{www.NFLdata.com}. The authors employed logistic regression models to predict game winners, winners with the point spread (WTS), and over/under total points, experimenting with various feature sets including unigram features from tweets, statistical game features, and a combination of both through canonical correlation analysis (CCA) for dimensionality reduction. The results showed that Twitter data, particularly rate of change in tweet volume (rateS), could significantly improve the accuracy of the prediction, achieving a precision greater than 55\% for the WTS predictions, which is sufficient for profitability in the betting markets. The dataset used for this study is publicly available at \url{www.ark.cs.cmu.edu/football}, facilitating further research into the intersection of social media and sports analytics.

In an innovative approach, \citet{burke2019deepqb} presented a deep learning application using player tracking data from two full seasons of NFL football, specifically using the NFL's Next Generation Stats system. The DeepQB model suite employed a modular feedforward artificial neural network to predict and evaluate pass decisions and outcomes at the play level, incorporating features such as player positions, velocities, and orientations, as well as play metadata such as down and distance. The models included four variants: estimation of the probability of targeting each receiver, expected yards gained per receiver, probabilities of completion, incompletion, or interception, and an experimental variant using transfer learning to model individual quarterback decision-making. The key findings showed that the model accurately predicted the targeted receiver 60\% of the time and highlighted that when quarterbacks followed the prediction of the model, they achieved a completion rate 74\% compared to 55\% otherwise. Despite higher completion rates, targeting the predicted receiver resulted in a lower YPA (7.8 vs 8.0), suggesting that quarterbacks might favor safer and less optimal choices. The data used covered 45,501 pass attempts from the 2016 and 2017 NFL seasons for training and validation, with the 2018 season data used for testing.

\citet{roumani2023sports} compared the classification performance of the C4.5, Artificial Neural Network (ANN), and Random Forest (RF) models to predict the Superbowl winner using regular season data from the NFL, specifically covering the 2002-2019 seasons. The study tackled the issue of data imbalance—where only a small fraction of teams win the Superbowl each season—by applying the Synthetic Minority Oversampling Technique (SMOTE) to balance the training data. The models were evaluated on recall, AUC, accuracy, and specificity. The results showed that the Random Forest model outperformed the others with the highest recall (79\%), AUC (93\%), accuracy (84\%), and specificity (88\%) in the oversampling of 1800\%. The findings suggested that Random Forest is the most effective classifier for this task, highlighting its potential to develop decision support tools for NFL team managers and coaches. The data set used for this research was collected from \url{www.NFL.com}.

In a comprehensive study, \citet{morgan2024forecasting} utilized a multiple linear regression model to predict the outcomes of the NFL playoff games, focusing on historical performance data. The study employed a stepwise variable selection method to identify significant predictors and integrate them into the model. The dataset, sourced from the "NFLfastR" package, included game data from 2002-2023. The model achieved an accuracy of 64.41\% in predicting game winners and 56.78\% against spread. The key metrics used to evaluate the model included adjusted R², which for the 2023 NFL playoffs was 0.3168 for home scores and 0.3301 for away scores, and practical metrics such as correct win percentage and correct win percentage against the spread.

\begin{table*}[h]
    \centering
    \caption{Summary of approaches in predicting baseball outcomes}\label{lbl_baseball_tbl}
    \centering
    \fontsize{6.5}{6.5}\selectfont
    \renewcommand{\arraystretch}{1.6}
    \begin{tabular}{|C{2cm}|C{2cm}|C{2cm}|C{1.5cm}|C{2.5cm}|C{2.5cm}|}
    \hline
    \textbf{Approaches} & \textbf{Work} & \textbf{Performance} & \textbf{Metrics} & \textbf{Features} & \textbf{Datasets}\\ 
    \hline
        Deep Learning Models & \citet{lee2022prediction}, \citet{park2018deep}, \citet{kim2023baseball}, \citet{chun2021inter}, \citet{hanprediction}, \citet{kim2020study}, \citet{hughes2022regression}, \citet{sun2023performance} & EP2: 62.4\%, DNN Ensemble: RMSE 15.17\%, MAPE 14.34\%, KoBERT: Accuracy 0.7430, Inter-dependent LSTM: +12\%, Playoff Prediction DNN: 88.33\%, TensorFlow DL: 89\%, Multilinear \& Logistic Regression: 56.6\%, LSTM: Lower MAE \& RMSE compared to ZiPS & Accuracy, RMSE, MAPE, Precision, Recall, F1 score, AUC, Brier Score, MAE, RMSE & Pitch types, locations, game attributes, spectators, sentiment analysis, starting lineups, OPS, runs, pitching data, batting data & KBO (2015-2016, 2008-2017, 2020-2021, 2019), Kaggle (1999-2021), MLB (1961-2019) \\
        \hline
        Ensemble Models & \citet{park2018deep}, \citet{park2023machine} & DNN Ensemble: RMSE 15.17\%, MAPE 14.34\%, XGBoost: Precision 0.5-0.6 & RMSE, MAPE, Precision & Spectators, player performance & KBO (2008-2017), MiLB, MLB, KBO \\
        \hline
        Natural Language Processing (NLP) & \citet{kim2023baseball} & KoBERT: Accuracy 0.7430 & Accuracy & Sentiment analysis, traditional statistics & KBO (2020-2021), Naver Sports \\
        \hline
        Data Mining Techniques & \citet{oh2014using} & Random Forest (binary dataset): 84.14\% & Accuracy & Annual salary, earned run, strikeout, pitcher’s winning percentage, four balls & KBO \\
        \hline
        Machine Learning Models & \citet{seo2019win}, \citet{jang2014analyzing}, \citet{cui2020forecasting}, \citet{koseler2017machine}, \citet{hoang2015dynamic}, \citet{li2022exploring}, \citet{valero2016predicting}, \citet{elfrink2018predicting}, \citet{yaseen2022multimodal}, \citet{hamilton2014applying}, \citet{barbee2020prediction}, \citet{chang2021construction} &  SVM: 60\%, Logistic Regression: 77\%, XGBoost: 55.52\%, LR, SVM: 65.75\%, SVM: 93.4-94.18\%, k-NN: 95\%, Logistic Regression: 61.77\% & Accuracy, AUC, Precision, Recall, F1 score & Historical data, player performance, sabermetrics, game context, PITCH f/x data, player statistics & KBO, MLB (1930-2016, 2008-2012, 2015-2019), Lahman Database, Retrosheet, Baseball-Reference, Baseball Savant, Bet365, NPB \\
        \hline
        Heuristic Methods & \citet{sumitani2016predicting} & PFTI: Accuracy 20\% & Accuracy & Pitched frequency, intervals & NPB (2014-2015) \\
        \hline
        Probabilistic Models & \citet{yoshihara2020pitch} & STM: Identified distinct trends in pitch sequences & Semantic coherence, exclusivity, marginal effects & Pitch sequences, game situations, player characteristics & NPB (2016-2018) \\
        \hline
        Siamese Neural Networks & \citet{mun2023competenet} & CompeteNet: Accuracy 60\% & Accuracy, binary classification performance & Pitcher and batter features, team features & KBO (2012-2021) \\
        \hline
    \end{tabular}
    \label{tab:baseball_models}
\end{table*}

\begin{figure*}
    \centering
    \includegraphics[width=0.79\textwidth]{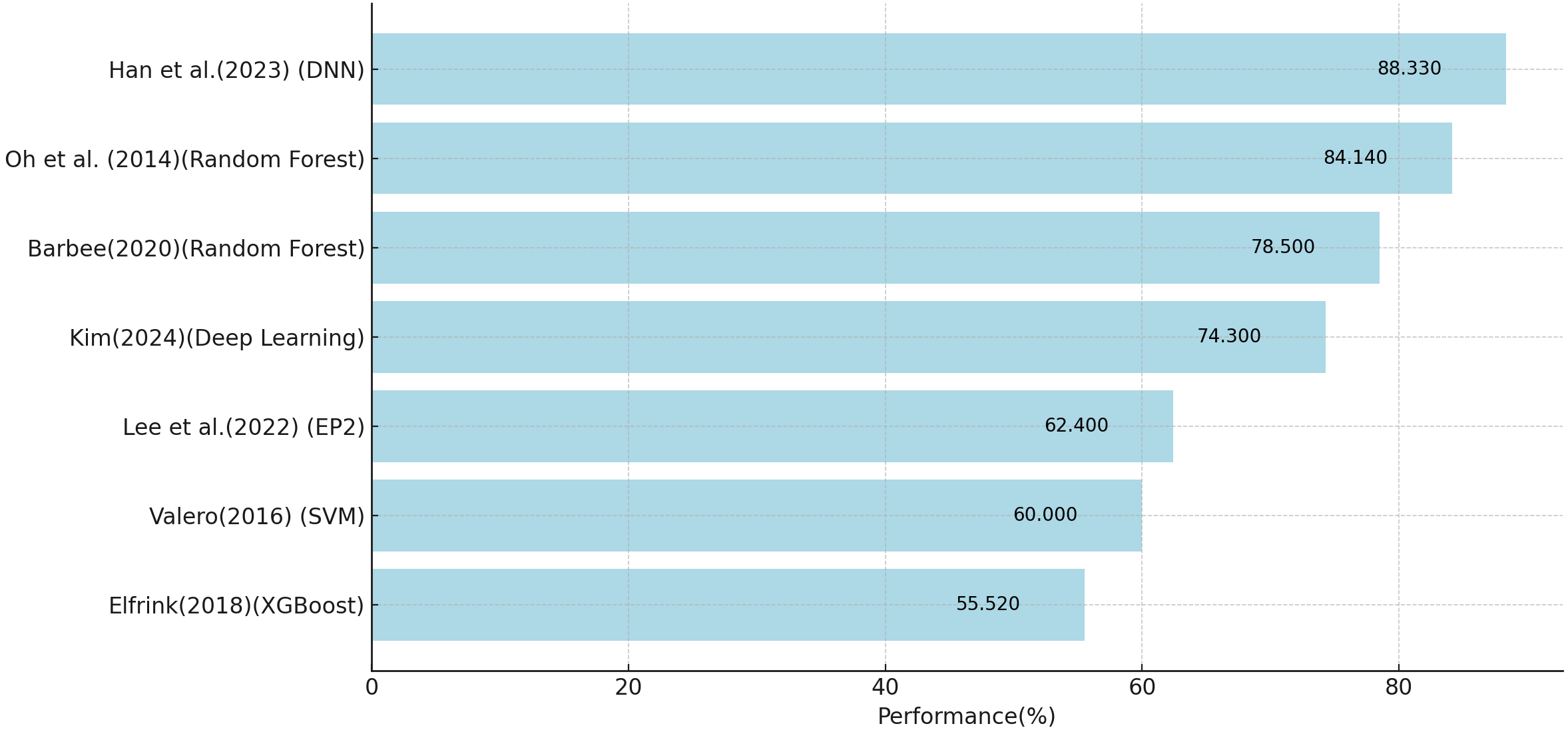}
    \caption{The best performances in Baseball analytics based on accuracy}
    \label{fig:enter-label-soccer-performance-baseball}
\end{figure*}

\subsection{Baseball \includegraphics[height=0.3cm]{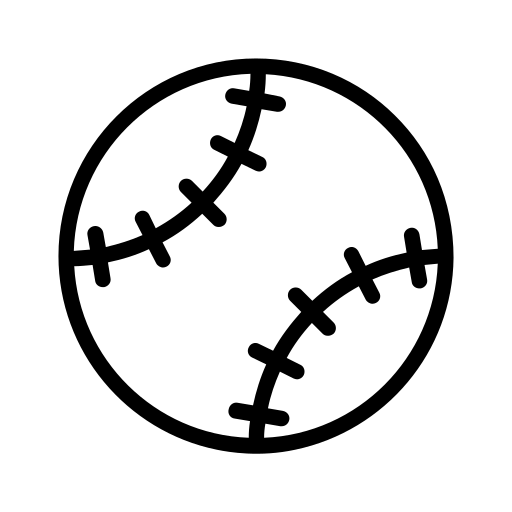}}\label{lbl_baseball_sec}

\begin{figure}[h!]
\fontsize{8}{8}\selectfont
\begin{tikzpicture}
    \begin{axis}[
        ybar,
        symbolic x coords={2014, 2015, 2016, 2017, 2018, 2019, 2020, 2021, 2022, 2023},
        xtick=data,
        x tick label style={rotate=45, anchor=east},
        ymin=0,
        xlabel={},
        ylabel={Number of articles},
        bar width=9.5pt,
        width=7.8cm,
        enlarge x limits=0.15,
    ]
        \addplot coordinates {(2014,2) (2015,1) (2016,2) (2017,1) (2018,3) (2019,2) (2020,1) (2021,2) (2022,2) (2023,2)};
    \end{axis}
\end{tikzpicture}
\caption{Histogram showing the number of articles per year in Baseball betting.}
    \label{fig:articles_per_year_baseball}
\end{figure}
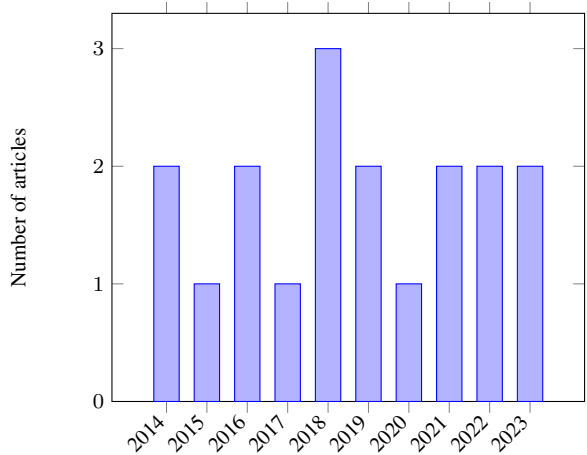

Predicting baseball game outcomes has been thoroughly explored using statistical and machine learning models, relying on historical data and player performance statistics to improve accuracy. Numerous approaches have been developed, each focusing on different methods and evaluation metrics. These models take into account factors such as team performance, individual player statistics, game conditions, and advanced analytics to produce more precise predictions of game outcomes (Table \ref{lbl_baseball_tbl} and Figures \ref{fig:enter-label-soccer-performance-baseball} and \ref{fig:articles_per_year_baseball}).

\citet{lee2022prediction} developed a system that predicted both pitch types and pitch locations in baseball, unlike previous research focusing only on pitch types. The study used an ensemble model of deep neural networks (DNNs) to handle the prediction of 34 classes, representing combinations of pitch types and locations. The dataset, collected from videos from the Korean Baseball Organization (KBO) league, included detailed attributes such as inning, progress (win/loss/draw), runner positions, ball/strike count and batter information. The final model, named EP2, achieved an accuracy of 62.4\%, significantly higher than the 8.2\% accuracy of naive prediction. The performance of the model varied with game situations, showing greater accuracy when the pitcher's team was losing, the count was unfavorable, or there were runners in scoring positions. The research demonstrated the potential for a more nuanced and practical prediction system in baseball analytics. The dataset used was from the KBO league, specifically analyzing games from 2015 and 2016.

Similarly, \citet{park2018deep} used a Deep Neural Network (DNN) to predict the daily number of spectators at the Gwangju-KIA Champions Field for Korean Baseball League games. The model included four types of DNN configurations with dropout and batch normalization to prevent overfitting. The data used for training and testing the models spanned from 2008 to 2017, with 80\% of the data used for training and 20\% for testing. The DNN models were evaluated using Root Mean Square Error (RMSE) and Mean Absolute Percentage Error (MAPE). The ensemble model, which averaged the predictions from multiple DNNs, demonstrated the highest accuracy, outperforming traditional multiple linear regression models by achieving RMSE and MAPE improvements of 15.17\% and 14.34\%, respectively. The dataset utilized for this research was derived from the official records provided by the KBO and additional sources from KIA Tigers' homepage.

In a related vein, \citet{kim2023baseball} proposed a baseball match prediction system that combined game record statistics with natural language processing (NLP) of news articles. They utilized the KoBERT model for binary sentiment classification and embedding of news articles, integrating these with traditional statistical data to account for external factors such as player conditions and team morale. The dataset included KBO game statistics from May 2020 to July 2021 and 40,674 news articles from Naver Sports. The traditional statistics-based model achieved an accuracy of 0.6508, while adding sentiment analysis improved accuracy to 0.7222, and using article embeddings further increased it to 0.7430, demonstrating the performance improvement from incorporating NLP.

Furthermore, \citet{oh2014using} explored various data mining techniques to predict the outcomes of Korean professional baseball games using historical data from the KBO website. They prepared three types of datasets: raw, ratio, and binary, and applied seven classification techniques: decision tree, random forest, logistic regression, neural network, support vector machine, linear discriminant analysis, and quadratic discriminant analysis. Among the 21 prediction scenarios (3 datasets × 7 techniques), the random forest model using the binary dataset achieved the highest accuracy of 84.14\%. The study highlighted that using ratio and binary datasets resulted in better prediction models than using raw data. The key winning factors identified included annual salary, earned run, strikeout, pitcher’s winning percentage, and four balls. This research is unique in its use of multiple data types and a variety of data mining techniques to predict wins and losses in Korean professional baseball games.

Furthermore, \citet{chun2021inter} proposed an interdependent LSTM to predict baseball game outcomes using only information from the starting lineup, addressing the issue of incomplete data in pre-game predictions. The model preprocesses historical data from the Korean Baseball Organization (KBO) by creating pairs of pre-game and post-game records, allowing the LSTM to learn dependencies between these events. This approach was contrasted with traditional methods that suffer from accuracy loss due to unknown substitutions. The interdependent LSTM used the pre-game data for odd-numbered cells and post-game data for even-numbered cells, capturing the transition patterns. Experiments using 720 KBO games from 2019 demonstrated that the proposed LSTM model achieved up to 12\% higher accuracy compared to conventional methods, including DNN. The dataset utilized for the study was from the KBO games in 2019.

Moreover, \citet{jang2014analyzing} utilized a k-NN algorithm to predict whether specific baseball players could be selected for the South Korean national team by analyzing historical player data. They used a dataset comprising past performance metrics (such as batting average, home runs, RBIs, walks, bases, mistakes, most hits, scores, and bats) of nine players who made the final roster for the 2010 Asian Games and five current players considered potential candidates. The model calculated the distance between the candidate players and the established team members to determine the likelihood of selection. The dataset was obtained from the KBO and Samsung Lions websites. The k-NN algorithm was chosen for its accuracy in classification tasks and demonstrated the potential to improve the selection process for players of national teams, contributing to the development of professional baseball in South Korea. Future work was planned to compare the performance of the k-NN algorithm with other machine learning methods such as SVM and decision trees.

In conjunction with this, \citet{seo2019win} proposed a model for predicting the outcomes of Korean professional baseball games using a deep learning technique. They used both basic baseball data and Sabermetrics data, which have a high correlation with scores, to train their supervised learning model. The model incorporated the Drop-Out algorithm and the ReLu activation function to improve the stability and accuracy of the predictions. The dataset used for training spans from 2006 to 2011, sourced from iSTAT. The results showed that their model predicted win rates 4.2\% more accurately than the traditional Pythagorean expectation method. The metrics utilized in evaluating the effectiveness of the model included actual win percentages, Pythagorean expectations, and the win percentages predicted by their model.

Following a similar pattern, \citet{hanprediction} developed a Deep Neural Network (DNN) model to predict the likelihood that KBO teams advance to the playoffs. Utilizing data from 1999 to 2021, sourced from Statiz and Kaggle, the model incorporated key features such as On-base Plus Slugging (OPS), runs, wild pitches, shutouts, and Grounded into Double Play (GDP) to predict playoff outcomes. Data were normalized and various neural network architectures were tested to optimize performance, ultimately using a model with a sigmoid accuracy function, softmax activation function, binary cross-entropy loss function and Adam optimizer. This configuration achieved an accuracy of 88.33\% when predicting the outcomes of the 2022 season. The study highlighted that variables such as OPS and wild pitches significantly impact predictions, and the high accuracy of the model demonstrated its potential utility for predicting playoff advancements in other baseball leagues. The dataset used was the KBO Batting and KBO Pitching data provided by Kaggle.

In parallel, \citet{kim2020study} developed artificial intelligence models to predict the winning-loss outcomes and the final league rankings of Korean professional baseball teams. The models utilized were k-NN, AdaBoost, and a customized deep learning model using TensorFlow. The predictions were based on data from the first, third and fifth innings of the games, with k-NN being the most accurate for the first inning and AdaBoost for the third and fifth innings. The deep learning model achieved an overall accuracy of 89\%. The dataset used consisted of game data from January 2016 to August 2020 for the Kia Tigers, with actual results used for games that did not involve the Kia Tigers. The study found that as the season progressed, the prediction error in rankings decreased, suggesting that the models improved performance over time. The metrics used to evaluate the models included training accuracy, test accuracy, and ranking error.

Complementing these findings, \citet{park2023machine} utilized machine learning models, including Logistic Regression (LR), Random Forest (RF), XGBoost, LightGBM, and CatBoost, to predict contract renewals for foreign players in KBO. They employed player performance data from Minor League Baseball (MiLB), Major League Baseball (MLB), and KBO, together with player image data processed using VGGFace. The results indicated that the performance of the players in KBO significantly affected the predictions, with precision rates ranging from 0.5 for pitchers and 0.533 for batters using XGBoost, and up to 0.6 for batters using LR with MLB data. The dataset utilized in the study included publicly available data from sources like Baseball-Reference, FanGraphs, and Baseball Savant.

Expanding on this concept, \citet{sumitani2016predicting} introduced a heuristic method to predict starting pitchers in baseball. The model leveraged four prediction approaches—PF (Pitched Frequency), PFA (Pitched Frequency After Pitcher), PFT (Pitched Frequency Per Team), and PFTI (Pitched Frequency Per Team using Interval). The model mimicked human prediction methods to overcome challenges due to limited data and variability in pitcher selection. Evaluations using data from the Nippon Pro-Baseball (NPB) central league (2014 and 2015) revealed that the PFTI model, which incorporated game intervals in its predictions, achieved the highest accuracy. The dataset comprised 6 teams' game records across two years, resulting in 143 or 144 pairs of pitchers and teams each year. Performance was measured by accuracy (AC), defined as the proportion of correctly predicted games. The PFTI model's average accuracy was approximately 20\%, outperforming other methods and comparable to human prediction accuracy, illustrating the potential effectiveness of heuristic rules in predicting starting pitchers. The dataset used was sourced from the "Professional Baseball Data Freak" website.

Similarly, \citet{kato2022pulled} utilized a machine learning approach to analyze fly balls in professional baseball, focusing on the aerodynamic effects on the probability of a flyout (P(flyout)) between the pull side (PS) and the opposite side (OS). Using radar-tracking data from the 2018-2019 Nippon Professional Baseball (NPB) seasons, they analyzed 25,413 fly balls, identifying that pulled low fly balls have greater variability in deflecting motions than their counterparts on the opposite side. The study employed the eXtreme Gradient Boosting algorithm to predict P(flyout) based on kinematic characteristics such as launch speed, vertical and horizontal launch angles, flight distance and hang time. The results demonstrated that the low fly balls on the PS had a lower P(flyout) (0.41) compared to OS (0.49) and the flyout zone for outfielders on the PS was significantly smaller (mean = 698 m²) than on the OS ($\geq$ 779 m²). This study confirmed that aerodynamic effects and variability in deflecting motions make catching pulled low fly balls more challenging. The dataset used was provided by NPB radar-tracking systems (TrackMan A/S, Denmark).

In a similar study, \citet{yoshihara2020pitch} applied a probabilistic topic model to analyze pitch sequences in Nippon Professional Baseball (NPB) games from 2016 to 2018, using data covering 799,507 pitches and 196,789 plate appearances. The model, specifically a structural topic model (STM), identified pitch sequence patterns in relation to pitcher/hitter characteristics, game situations, and hit results. The results revealed different trends: for example, sequences that aim to avoid sweet spots for high-average hitters and strategies to induce swings at bad pitches for home runs. Marginal effects showed pitchers' tendencies to jam hitters or avoid inside pitches based on outs and baserunner situations. The study also highlighted specific trends for individual pitchers, such as Takahiro Norimoto's pattern of tricking hitters with fastballs and sliders. Metrics used included the marginal effects of metadata on topic proportions, semantic coherence, and exclusivity of topics. The dataset from Data Stadium Inc. included metadata like pitch type, speed, location, ball/strike call, and player characteristics.

Furthermore, \citet{mun2023competenet} introduced CompeteNet, a prediction model using Siamese neural networks to predict win-loss outcomes in baseball games. The model learned the competitive relationship between a pitcher and a batter from two teams by encoding them into embedding vectors. The dataset comprised data from the Korean Baseball Organization (KBO) data, including 17 pitcher features, 16 batter features, and 3 team features. Extensive experiments demonstrated that CompeteNet outperformed traditional machine learning models, achieving a maximum accuracy of 60.00\%. The key metrics used included accuracy and binary classification performance, with the model showing significant improvements over baseline methods such as logistic regression, SVM, and MLP.

In Major League Baseball (MLB), \citet{elfrink2018predicting} investigated the prediction of MLB game outcomes using machine learning algorithms, particularly random forests and XGBoost, applied to historical data from the Retrosheet dataset (1930-2016). The study involved pre-processing data to extract relevant features such as game context and team performance statistics, both seasonal and recent. Two prediction models were tested: a binary classification to predict the winning team and a regression model to predict the difference in scores. Among the models tested, XGBoost achieved the highest accuracy of 55.52\% in predicting game outcomes as a binary classification problem using only season data, which outperformed random guessing and simple prediction of home teams. The study highlighted the potential of machine learning to predict game outcomes, but acknowledged that the results are not yet robust enough to surpass betting companies, suggesting further improvements in data quality, feature engineering, and model optimization.

Consequently, \citet{huang2021use} used machine learning and deep learning models to predict the outcomes of Major League Baseball (MLB) matches using data from the 2019 MLB season. The study compared the performance of a one-dimensional convolutional neural network (1DCNN), an ANN, and an SVM on two datasets: one with data from only the starting pitcher and the other including data from all pitchers. The highest prediction accuracies were 93.4\%, 93.91\%, and 93.90\% with the 1DCNN, ANN, and SVM models, respectively, before feature selection. After feature selection, the highest accuracies were 94.18\% and 94.16\% with the ANN and SVM models, respectively. The study demonstrated that using data from all pitchers resulted in higher accuracy compared to using only the starting pitcher's data. The metrics used included accuracy, precision, recall, and the F1 score, with five-fold cross-validation used to evaluate model performance. Data were sourced from \url{www.Baseball-Reference.com}.

In conjunction with these findings, \citet{li2022exploring} predicted the outcomes of Major League Baseball (MLB) games by exploring and selecting relevant features. They utilized game data from the 2015-2019 MLB seasons and tested four prediction models: one-dimensional convolutional neural network (1DCNN), ANN, SVM, and logistic regression (LR). The dataset, comprising 30 individual team datasets, included variables related to hits, pitcher performance, and scoring, which were normalized and split into training and testing sets. Feature selection was performed using Recursive Feature Elimination (RFE). The SVM model yielded the highest prediction accuracy of 65.75\% and an AUC of 0.6501 after feature selection, outperforming the other models. The study concluded that feature selection and optimized parameter combination significantly improved prediction performance, surpassing previous research in the field.

Following a similar pattern, \citet{valero2016predicting} compared the predictive capabilities of four machine learning techniques - k-NN, ANN, decision trees (DT), and SVM - to predict win-loss outcomes in regular season MLB games using ten years of sabermetrics statistics from publicly available datasets such as Retrosheet and Lahman Database. The research employed both classification and regression schemes, using a feature selection process to identify the most important predictors, including home field advantage, Log5 and Pythagorean expectation. The study found that SVMs produced the best predictive results, achieving nearly 60\% accuracy with a 10-fold stratified cross-validation methodology. Overall, the classification schemes outperformed the regression schemes, and while the accuracy of the model was an improvement over random guessing, it highlighted the inherent complexity of sports prediction problems.

\citet{cui2020forecasting} employed a regularized logistic regression elastic net model to predict the outcomes of the MLB game. The study used game-level data from Retrosheet and Sean Lahman's Baseball Database, including features like differences in on-base percentage (OBP), rest days, isolated power (ISO) and average baserunners allowed (WHIP) between home and away teams. The model, trained on data from 2001-2015 and tested on 9,700 games from 2016-2019, achieved a classification accuracy of 61.77\% and an AUC of 0.6706. Further refinements, such as the incorporation of game-level pitcher covariates and Bayesian hyperparameter optimization, are suggested to potentially improve predictive accuracy and model lift, with the strongest results reaching 64-65\% accuracy when evaluated monthly.

Furthermore, \citet{hughes2022regression} utilized a combination of multilinear and logistic regression models to predict the outcomes of baseball games. The player value model used multilinear regression to quantify player contributions based on statistics such as OPS, SB, RBI, K\% (Strikeout Rate), and BB\% (Walk Rate) for hitters, and ERA, K\%, BB\%, and opposing OPS for pitchers. These player values were then aggregated to assign team values. The logistic regression model predicted game outcomes by comparing aggregated team values, achieving a classification accuracy of 56.6\%, an AUC of 0.549, and a Brier score of 0.244. The Lahman Baseball Database and Baseball Savant were used for data sourcing. Despite a modest classification accuracy, the model showed promise in probabilistic predictions.

Moreover, \citet{barbee2020prediction} utilized binary logistic regression, random forests, neural networks, and Bayesian logistic regression to classify the successful pitches, specifically strikeouts, in MLB games. Using data from MLB Pitch datasets from 2015 to 2018, comprising 52 variables and 736,325 observations, the study evaluated the performance of each model. The results indicated that both the random forest model with the original data and the neural network model with a balanced data set outperformed the accuracy of the MLB umpires, achieving prediction accuracies that exceeded the umpire benchmark of 78.5\%. The performance metrics for the model included accuracy, sensitivity, and specificity, and the study used variable importance plots to determine the significance of various predictors such as pitch type, pitch speed, and pitch location.

In addition, \citet{chang2021construction} developed a model that outperformed the odds of online bookmakers by predicting the outcomes of the first five innings of MLB games. The study utilized a Markov process method and the runner advancement model (RAM) to estimate expected runs based on the batting lineup and pitcher statistics from historical data available on the MLB official website. Predictive performance was evaluated using two main metrics: Ranked Probability Score (RPS) and Expected Value (EV). The results indicated that the new implied probability method (NIP), which incorporated batter vs. pitcher career statistics, provided more accurate and profitable predictions compared to the basic normalization method (BN). The dataset used for the analysis comprised 70 MLB matches from September 2018, with betting odds sourced from the Bet365 online bookmaker. The NIP method demonstrated a prediction accuracy 75\% with an expected value of 10.15, indicating its potential advantage in sports betting markets.

Likewise, \citet{hamilton2014applying} utilized machine learning methods to predict the type of pitch in baseball using PITCH f/x data from MLB games during the 2008 and 2009 seasons. The authors applied classification techniques, including SVM and k-NN, and introduced a novel approach to feature selection tailored to individual pitchers and specific game situations. They used 18 features from the raw data and generated additional relevant features, achieving a prediction accuracy of approximately 80.88\% with k-NN and 79.76\% with SVM. Their model showed a 20.85\% improvement over naive guessing. The dataset utilized for this study was the PITCH f/x data, which contain detailed information about each pitch, including speed, break angle, and type.

In a similar vein, \citet{sun2023performance} investigated the use of long-short-term memory (LSTM) networks to predict the future performance of Major League Baseball (MLB) players, specifically focusing on the number of home runs (HR). The researchers compiled a dataset of 5,401 players from 1961 to 2019, sourced from \url{www.Baseball-Reference.com}, and utilized 21 features, including age, height, weight, and various performance metrics. They trained several LSTM models, as well as traditional machine learning models such as linear regression, support vector machines, random forests, and neural networks. LSTM models outperformed other methods, demonstrating a lower mean absolute error (MAE) and a lower root mean square error (RMSE) in the 2018 and 2019 predictions. The study compared these models against the widely used ZiPS projection system, finding that LSTM models provided more accurate predictions, particularly for players hitting fewer than 30 home runs. However, predicting players with more than 30 home runs remains a challenge. The dataset utilized in the study was \url{www.Baseball-Reference.com}.

Similarly, \citet{yaseen2022multimodal} developed a model to predict MLB playoff teams using 20 years of historical data and machine learning techniques. The study employed Logistic Regression and Support Vector Classifier (SVC) algorithms, utilizing features like Runs, Batting Average, Homeruns, Strikeouts, Innings Pitched, Earned Runs, and Earned Runs Average. The dataset was compiled from \url{www.Baseball-Reference.com}. The models achieved a prediction accuracy of 77\%, with recall of 59\%, resulting in an overall accuracy of 60\%. Despite the limitations of the model, the study suggested potential improvements with more advanced algorithms and additional features. The key metrics used for the evaluation included precision, recall, F1 score, and AUC-ROC, with the ROC AUC score being 0.76, indicating the model's ability to distinguish between true positives and false positives.

Moreover, \citet{koseler2017machine} reviewed the use of machine learning in baseball analytics, focusing on three primary problem classes: Binary Classification, Multiclass Classification, and Regression. SVM and k-NN are the algorithms that are the most used. In binary classification, SVMs significantly improved pitch prediction accuracy, achieving a prediction rate 70\%. In Multiclass Classification, bagged random forests of classification trees outperformed other methods in predicting pitch types. For regression tasks, Bayesian inference and linear regression with feature selection are common, with linear regression models predicting free agent performance effectively using WAR (Wins Above Replacement) as a key feature. The PITCHf/x dataset was extensively used in these studies to track and analyze pitch data.

Building on this, \citet{hoang2015dynamic} presented a machine learning approach to predict pitch types (fastball vs. nonfastball) using Linear Discriminant Analysis (LDA). The study utilized data from the PITCHf/x system, covering MLB games from 2008 to 2012, which includes more than 3.5 million observations with about 50 features each. The authors introduced an adaptive feature selection strategy that dynamically selected features based on different pitcher/count situations to mimic a pitcher's cognitive process. The approach showed an 8\% improvement in the accuracy of the prediction over previous studies, achieving an average accuracy of 78\%. The evaluation metrics used included accuracy and the area under the ROC curve (AUC), with cross-validation employed to prevent overfitting.
Finally, \citet{chun2021inter} proposed an inter-dependent LSTM model to predict Korean Baseball Organization (KBO) game outcomes using only starting lineup information. Traditional models struggle due to incomplete player lists and mid-game substitutions, but this model processes pairs of pre-game and post-game records to capture hidden patterns in time series data. Pre-game data includes starting lineups, while post-game data includes all participating players. The interdependent LSTM model alternates these inputs between odd and even cells to learn dependencies, enhancing prediction accuracy. Their experiments with 720 KBO 2019 game records demonstrated that this approach increased prediction accuracy by up to 12\% compared to existing models, addressing the challenge of missing in-game substitution information. The dataset used was derived from KBO games in 2019.

\begin{table*}[h!]
    \centering
    \caption{Summary of Horse Racing prediction models}
    \label{lbl_horse_racing}
    \centering
    \fontsize{6.5}{6.5}\selectfont
    \renewcommand{\arraystretch}{1.6}
    \begin{tabular}{|C{2cm}|C{2cm}|C{2cm}|C{1.5cm}|C{2.5cm}|C{2.5cm}|}
\hline
\textbf{Approaches} & \textbf{Work} & \textbf{Performance} & \textbf{Metrics} & \textbf{Features} & \textbf{Datasets} \\
\hline
\multicolumn{6}{|c|}{\textbf{Machine Learning Models}} \\
\hline
XGBoost & \citet{terawong2024xgboost}& XGBoost-trained bettor-agent: Outperformed original bettor-agents & Profitability & Synthetic data from bettor-agents employing simple strategies & Bristol Betting Exchange (BBE) ABM \\
\hline
Random Forest & \citet{gupta2024horse}, \citet{lessmann2010alternative} & RF: 93.1\% accuracy (Gupta et al.), RF: Higher profitability and predictive accuracy compared to traditional methods (Lessmann) & Accuracy, ROC, Profitability, NDCG & Information gain, Chi-square filtering, Kelly betting strategy & Horse Racing Company dataset (India), Hong Kong racetracks \\
\hline
\multicolumn{6}{|c|}{\textbf{Regression-Based Models}} \\
\hline
Linear Regression (LR) & \citet{gupta2024horse}, \citet{selvaraj2017predicting} & LR: Lower performance than RF (Gupta), LR: 89.58\% accuracy on the second dataset (Selvaraj) & Accuracy, ROC & Information gain, Chi-square filtering, Information gain ratio & Horse Racing Company dataset (India), Sporting Life, Kaggle \\
\hline
\multicolumn{6}{|c|}{\textbf{Classification-Based Models}} \\
\hline
Support Vector Machine (SVM) & \citet{gulum2018horse} & SVM: Improved accuracy with graph-based features (Gulum) & Rate of return, R-squared, t-values, AUC, Accuracy, Precision, Recall & Fundamental variables, market odds, Graph-based features (win-loss spread, node scores), basic race features & UK racetrack, Equibase website \\
\hline
k-Nearest Neighbors (k-NN) & \citet{gupta2024horse}, \citet{selvaraj2017predicting}  & k-NN: Lower performance than RF (Gupta), k-NN: 91.40\% accuracy on the second dataset (Selvaraj) & Accuracy, ROC & Information gain, Chi-square filtering, Information gain ratio & Horse Racing Company dataset (India), Sporting Life, Kaggle \\
\hline
Neural Network (NN) & \citet{borowski2021machine} , \citet{gulum2018horse} & NN: Competitive with LDA (Borowski), Improved accuracy with graph-based features (Gulum) & AUC, Net Profit Function, ROI, Correct Bet Ratio, Accuracy, Precision, Recall & Previous prizes won, jockey, trainer characteristics, Graph-based features (win-loss spread, node scores), basic race features & 3,782 races (Poland), Equibase website \\
\hline
\multicolumn{6}{|c|}{\textbf{Statistical Methods}} \\
\hline
Logistic Regression & \citet{selvaraj2017predicting}, \citet{gulum2018horse}, \citet{davoodi2010horse} & Logistic Regression: 89.58\% accuracy on the second dataset (Selvaraj), Improved accuracy with graph-based features (Gulum), Effective in predicting first-place horses (Davoodi) & Accuracy, Precision, Recall, Mean Squared Error (MSE) & Information gain ratio, basic race features, Graph-based features (win-loss spread, node scores), Horse weight, race type, trainer, jockey, number of horses, race distance, track condition, weather & Sporting Life, Kaggle, Equibase website, AQUEDUCT Race Track (New York) \\
\hline
\multicolumn{6}{|c|}{\textbf{Other Approaches}} \\
\hline
Fuzzy Logic & \citet{jogeeah2015using}, \citet{pudaruth2013horse} & Fuzzy Logic: 75\% accuracy (Jogeeah), Weighted Probabilistic Approach: 58\% accuracy (Pudaruth) & Prediction Accuracy, Precision, Recall, Number of winners predicted, profit per race & Horse speed, weight, past performance, Jockey performance, horse experience, odds, previous performances, draw positions, horse type by distance, weight, rating, stable reputation & Champ de Mars racecourse \\
\hline
\end{tabular}
\end{table*}

\begin{figure*}
    \centering
    \includegraphics[width=0.79\textwidth]{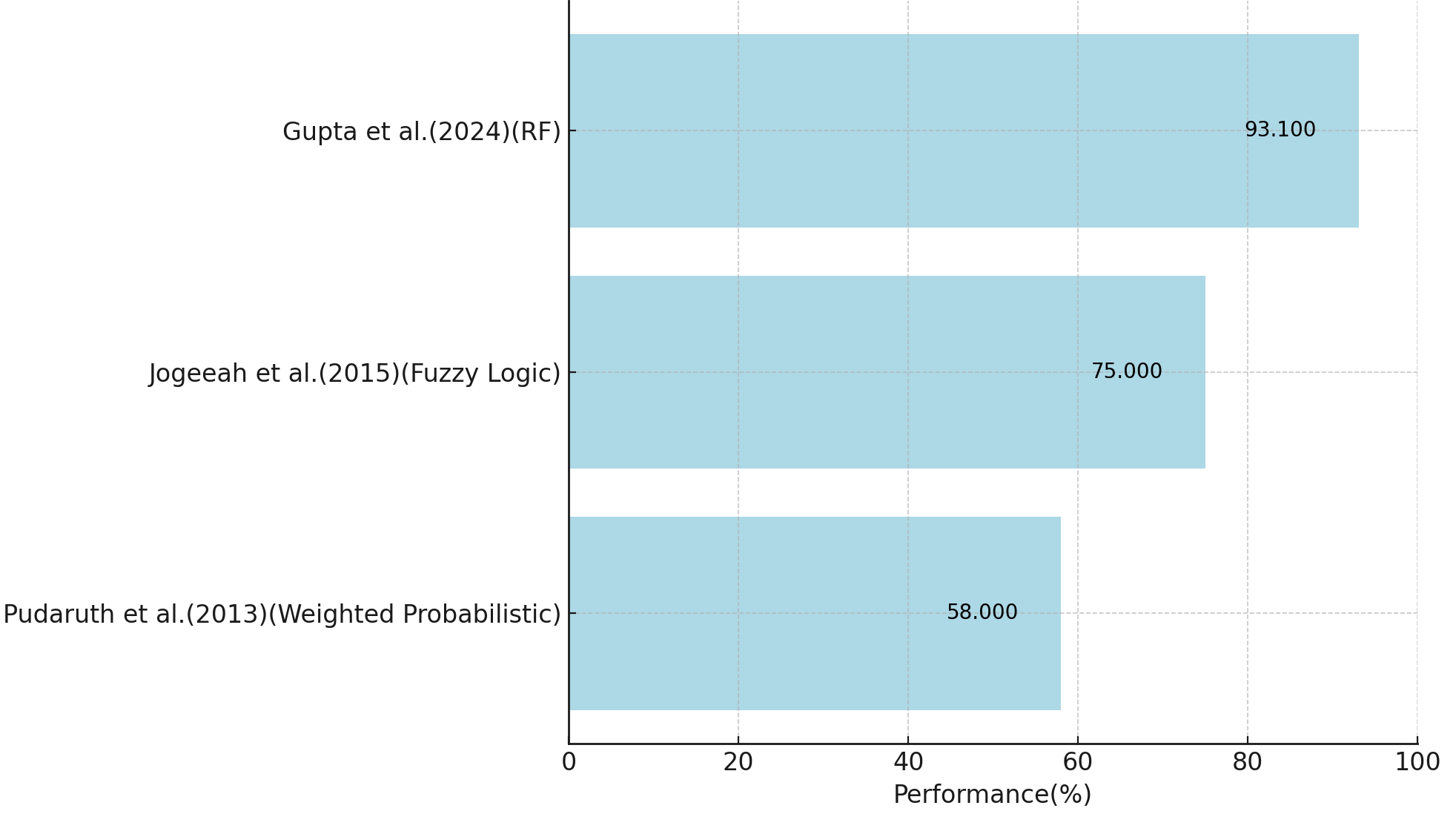}
    \caption{The best performances in Horse racing analytics based on accuracy}
    \label{fig:enter-label-soccer-performance-hrc}
\end{figure*}

\subsection{Horse Racing \includegraphics[height=0.3cm]{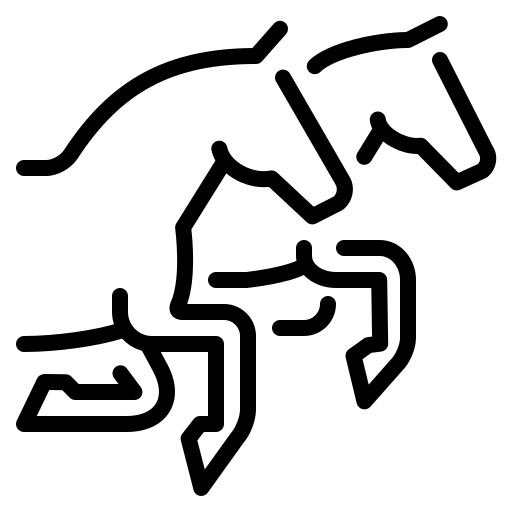}}\label{lbl_horse_racing_sec}

\begin{figure}[h!]
\fontsize{8}{8}\selectfont
\begin{tikzpicture}
    \begin{axis}[
        ybar,
        symbolic x coords={2010, 2013, 2015, 2017, 2018, 2021, 2024},
        xtick=data,
        x tick label style={rotate=45, anchor=east},
        ymin=0,
        xlabel={},
        ylabel={Number of articles},
        bar width=9.5pt,
        width=7.8cm,
        enlarge x limits=0.15,
    ]
        \addplot coordinates {(2010,2) (2013,1) (2015,1) (2017,1) (2018,1) (2021,1) (2024,3)};
    \end{axis}
\end{tikzpicture}
\caption{Histogram showing the number of articles per year in Horse Racing betting.}
    \label{fig:articles_per_year_Horse}
\end{figure}
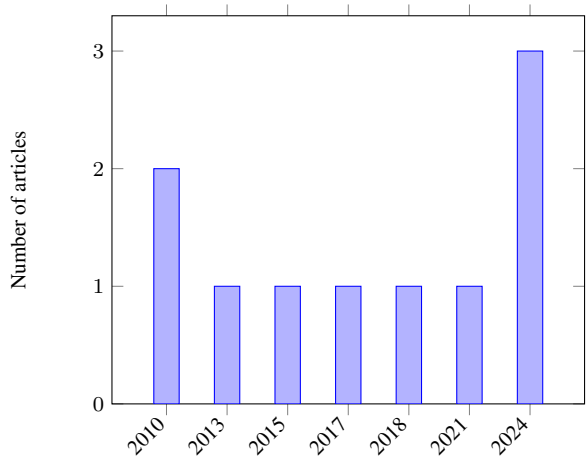

Horse racing analytics has advanced considerably over time, with many studies investigating different methods to predict race outcomes, assess horse performance, and analyze racing strategies. This section highlights key research contributions in the field, focusing on methodologies, datasets, and findings (Table \ref{lbl_horse_racing} and Figures \ref{fig:enter-label-soccer-performance-hrc} and \ref{fig:articles_per_year_Horse}).

\citet{terawong2024xgboost} employed XGBoost, a machine learning method, within the Bristol Betting Exchange (BBE) agent-based model (ABM) to develop profitable dynamic betting strategies for in-play betting during track-racing events such as horse races. They used BBE ABM to generate synthetic data from bettor-agents employing minimally simple strategies, which were fed into the XGBoost system to identify profitable betting patterns. The XGBoost-trained bettor-agent was then evaluated in various betting market scenarios within the BBE ABM. The primary metric for evaluation was profitability, and the XGBoost agent consistently outperformed the original bettor-agents used for training. The authors highlighted that their XGBoost agent learned profitable betting strategies that generalized well beyond training strategies, showcasing significant promise for automated betting systems. The dataset utilized was the simulated race records from the BBE ABM.

Similarly, \citet{gupta2024horse} used several machine learning models to predict the outcomes of horse races, leveraging a dataset of 14,750 rows and 56 attributes sourced from a well-known Horse Racing Company in India. The models included k-NN, Linear Regression (LR), Random Forest (RF), Gaussian Naïve Bayes (GNB), ADA Boost (BAG), and Bagging, with feature selection techniques such as information gain and Chi-square filtering applied to enhance predictive performance. The Random Forest model demonstrated superior accuracy at 93.1\% when trained on the original dataset, outperforming other models. Metrics such as accuracy and receiver operating characteristics (ROC) were used to evaluate the models, with RF showing the highest performance, particularly after applying feature selection methods such as information gain and recursive feature elimination (RFE). The study concluded that the Random Forest approach, combined with appropriate feature selection, offered a robust solution for horse race result prediction, providing valuable insights for the sports management and betting industries. Dataset used: Horse Racing Company dataset (India).

In a related vein, \citet{tondapu2024efficient} examined the UK horse racing market using Betfair's time series data. The market was characterized by short tails, rapidly decaying autocorrelations, and no long-term memory, indicating high information efficiency. The generalized Gaussian unconditional distribution with a light tail suggested that knowledge was quickly assimilated and reflected in prices. The analysis involved various statistical tests, such as KPSS and ADF, to confirm stationarity and assess market efficiency. The findings did not reveal strong tails in the return distributions and suggested a higher level of market efficiency compared to traditional financial markets. The dataset used in this study included 1,056,766 price change signals across 73 markets and 10 events, with messages transmitted every 50 milliseconds, sourced from Betfair's PRO package.

In addition, \citet{lessmann2010alternative} used a random forest classifier to predict race outcomes, adapting it to account for the competitive nature of horse racing. The study used a dataset of 1,000 races (12,902 horses) on Hong Kong race tracks. They demonstrated that the adapted random forest model yielded higher profitability and predictive accuracy compared to traditional methods like the conditional logit model (CL) and SVM. The random forest model achieved significant profits when used with a Kelly betting strategy, highlighting the model's ability to outperform the market's odds. The metrics used included the normalized discounted cumulative gain (NDCG) and the model's profitability in a betting context. The primary dataset utilized was from Hong Kong racetracks, covering races between January 2005 and December 2006.

Building on this, \citet{selvaraj2017predicting} focused on improving the accuracy of horse race outcome predictions using advanced data mining techniques. The study utilized two datasets: one web-scraped from Sporting Life and another from Kaggle. The methodology used CRISP-DM, which involves data understanding, preparation, and modeling. The key models used included k-NN, Linear Discriminant Analysis (LDA), and Logistic Regression, each applied after feature selection using the information gain ratio. The first dataset showed low accuracies (k-NN: 33.97\%, LDA: 33.33\%), highlighting the importance of relevant attributes. The second dataset, with more detailed attributes, achieved higher accuracies (k-NN: 91.40\%, LDA: 92\%, Logistic Regression: 89.58\%), emphasizing the need for a comprehensive feature selection for effective model performance.

\citet{borowski2021machine} compared six machine learning algorithms—Classification and Regression Tree (CART), Generalized Linear Model (Glmnet), Extreme Gradient Boosting (XGBoost), Random Forest (RF), Neural Network (NN), and Linear Discriminant Analysis (LDA)—to predict the outcomes of flat horse races in Poland and develop a profitable betting strategy. The dataset comprised 3,782 Arabian and Thoroughbred races from 2011-2020. The performance of the models was evaluated using the AUC, net profit function, ROI and the ratio of correctly placed bets. The best results were achieved with LDA and Neural Networks, with a correct bet ratio of 41\% for Win bets and more than 36\% for Quinella bets, indicating market inefficiencies. The most influential variables included the amount of previous prizes won by the horse, while jockey and trainer characteristics were less significant. The study found that profitable betting could be achieved under certain conditions, with the LDA model on the combined dataset of all horses showing the highest ROI for Win bets.

Expanding on this concept, \citet{pudaruth2013horse} developed a weighted probabilistic approach to predict the results of horse racing at Champ de Mars, Mauritius. The model analyzed factors such as jockey performance, horse experience, odds, previous performances, draw positions, horse type by distance, weight, rating, and stable reputation, each assigned a specific weight. Using data from the 2010 racing season, the model predicted race winners with a 58\% accuracy rate, outperforming professional tipsters who averaged a 44\% success rate. The metrics used included the number of predicted winners and the profit per race, the dataset comprising 240 races from 2010. The prediction system, designed in Java, allowed for automatic weight adjustments based on new race data, enhancing future prediction accuracy, and demonstrating significant profit potential for bettors.

In conjunction with this, \citet{jogeeah2015using} utilized a fuzzy logic model to predict race outcomes. The input of the model included variables such as horse speed, weight, and past performance, processed through a fuzzy inference system to handle the uncertainties and vagueness inherent in horseracing. The results demonstrated the effectiveness of the model, achieving a prediction accuracy of 75\%, a significant improvement over traditional statistical methods. The dataset utilized for this study was derived from historical race data from the Champ de Mars racecourse. The metrics used to evaluate the performance of the model included the accuracy, precision and recall of prediction, highlighting the capability of the model to identify winning horses accurately.

Following a similar pattern, \citet{gulum2018horse} utilized ANN and logistic regression models to predict horse race outcomes. The research focused on two approaches: using basic race features and integrating graph-based features derived from horse race data spanning from 2015 to 2017. The graph-based features included the win-loss spread and the node scores, calculated by creating a directed graph where the edges represented the race results between horses. These features were incorporated into the predictive models to assess improvements in accuracy. 
The dataset used in the study was sourced from the Equibase website and included comprehensive race details such as horse weights, jockeys, track conditions, weather, and previous race positions. The models were trained and tested on data from multiple races, with specific attention given to races from the 2016 and 2017 Kentucky Derby. The results showed that the incorporation of graph-based features improved the accuracy of the prediction compared to the use of only basic features. The models were particularly effective in predicting the outcomes of specific races when additional graph features reflecting specific conditions (e.g., track type, weather) were utilized. The study demonstrated that ANN and logistic regression models with graph-based features outperformed those with only basic features, highlighting the importance of a global perspective in horse race predictions. Metrics such as accuracy, precision and recall were used to evaluate model performance, and the results were compared with the baseline predictions provided by betting odds.

Lastly, \citet{davoodi2010horse} applied various ANN algorithms, including Back-Propagation (BP), Back-Propagation with Momentum (BPM), Quasi-Newton (BFGS), Levenberg-Marquardt (LM) and Conjugate Gradient Descent (CGD), to predict horse racing outcomes using data from the AQUEDUCT Race Track in New York, spanning 100 races in January 2010. Each ANN model predicted the finishing times of individual horses based on eight features: horse weight, race type, trainer, jockey, number of horses, race distance, track condition, and weather. The study found that BP had slightly better accuracy (77\%) in predicting first-place horses, but required longer training times, while LM was the fastest algorithm. The performance of the models was evaluated using the Mean Squared Error (MSE) metric. The experimental results showed that the CGD algorithm was better at predicting the last place horses and that BP and BPM were more effective at predicting the first place horses. In general, ANNs demonstrated significant potential for horse racing predictions with average accuracy across all algorithms.

\begin{table*}[h!]
    \centering
    \caption{Summary of Rugby prediction models}
    \label{lbl_rugby}
    \centering
    \fontsize{6.5}{6.5}\selectfont
    \renewcommand{\arraystretch}{1.6}
    \begin{tabular}{|C{2cm}|C{2cm}|C{2cm}|C{1.5cm}|C{2.5cm}|C{2.5cm}|} \hline
\textbf{Approaches} & \textbf{Work} & \textbf{Performance} & \textbf{Metrics} & \textbf{Features} & \textbf{Datasets} \\ \hline
\textbf{Multivariate Analysis} & \citet{fontana2017player} & Discriminant Function: 81\% (Int), 77\% (Nat) & MANOVA & Body mass, height, body fat, fat-free mass, SJ, CMJ, sprint times, VO2max & Italian Rugby Federation (FIR) youth draft camps \\ \hline

\multirow{5}{*}{\centering \textbf{Random Forest}} & \citet{welch2018training} & RF Classifier: AUC 0.98 (21-day model) & AUC & Total distance, high-speed distance, acceleration/deceleration load, high-metabolic-power distance, impulse, mechanical work & GPS tracking data from National Rugby League \\ \cline{2-6}
& \citet{scott2023classifying} & Full model: 83\% (relative), 64\% (isolated); Reduced model: 85\% (relative), 66\% (isolated) & Classification accuracy & Kicks from hand, metres made, clean breaks, turnovers conceded, scrum penalties & OPTA \\ \cline{2-6}
& \citet{xu2022machine} & Random Forest: 87.5\% accuracy & Mean decrease accuracy & Pass Forward, Penalty Kick, Tackle, Out, Scrum, Conversion, Yellow Cards, Red Cards, Possession Time & 14th China National Games rugby sevens \\ \cline{2-6}
& \citet{kvasnytsya2024team} & RF and Extra Trees Classifier: 92\% accuracy & Prediction accuracy & Possession, passes, tries, missed tackles & European Rugby Championships 15 \\ \cline{2-6}
& \citet{bennett2021predicting} & Group-phase model: 87.5\% accuracy & Classification accuracy & Tackle ratio, clean breaks, average carry, lineouts won, penalties conceded, missed tackles & OPTA \\ \hline

%\textbf{Simulation-based} & O'Brien et al. (2016) & Simulation: 10,000 runs & Prediction accuracy & World ranking points & IRB World Rankings and previous Rugby World Cups \\ \hline

\textbf{Dynamic Graph Analysis} & \citet{cintia2016haka} & Prediction accuracy: 83\% (meters gained), 77\% (match outcomes) & Connectivity, assortativity, clustering & Pass network connectivity & OPTA \\ \hline

\textbf{LSTM Model} & \citet{pituxcoosuvarn2022rugby} & LSTM: 74\% accuracy with augmented data & Prediction accuracy & Right knee, ankle, shoulder, eyes & OpenPose-extracted posture data \\ \hline

\textbf{Continuous-Time Model} & \citet{crewther2020longitudinal} & Temporal persistence, standardized effect & Time-dependent interrelationships & Salivary testosterone, training motivation & Not explicitly named \\ \hline

\textbf{Power Law Models} & \citet{howe2022modeling} & Linear relationships (R² = .967–.993) & Prediction errors & Mean speed, metabolic power, PlayerLoad™ & Wearable GNSS and accelerometers \\ \hline
\end{tabular}
\end{table*}

\begin{figure*}
    \centering
    \includegraphics[width=0.79\textwidth]{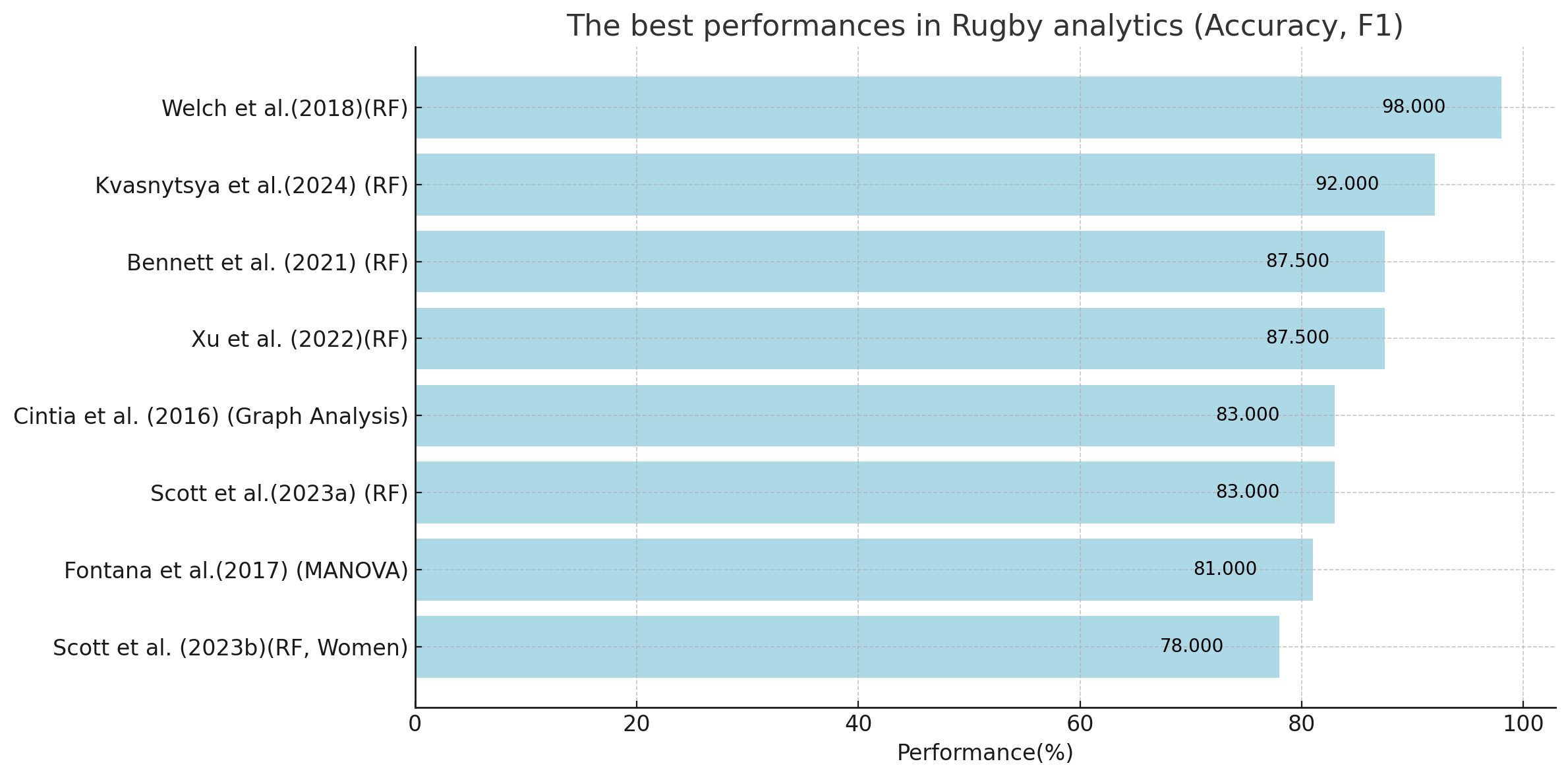}
    \caption{The best performances in Rugby analytics based on accuracy}
    \label{fig:enter-label-soccer-performance-rgb}
\end{figure*}

\subsection{Rugby \includegraphics[height=0.3cm]{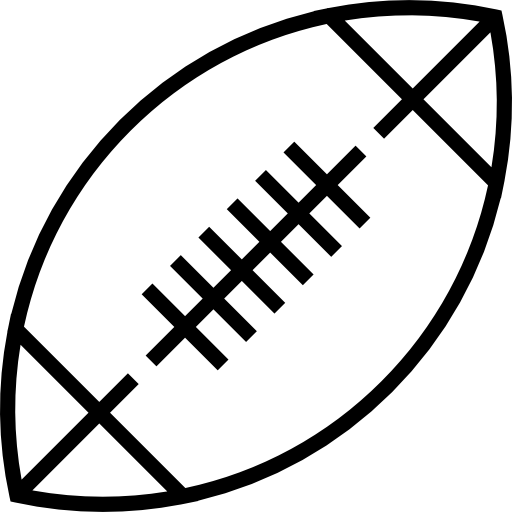}}\label{lbl_rugby_sec}

\begin{figure}[h!]
\fontsize{8}{8}\selectfont
\begin{tikzpicture}
    \begin{axis}[
        ybar,
        symbolic x coords={2012, 2016, 2017, 2018, 2020, 2021, 2022, 2023, 2024},
        xtick=data,
        x tick label style={rotate=45, anchor=east},
        ymin=0,
        xlabel={},
        ylabel={Number of articles},
        bar width=9.5pt,
        width=7.8cm,
        enlarge x limits=0.15,
    ]
        \addplot coordinates {(2012,1) (2016,2) (2017,1) (2018,1) (2020,1) (2021,1) (2022,3) (2023,2) (2024,1)};
    \end{axis}
\end{tikzpicture}
\caption{Histogram showing the number of articles per year in Rugby betting.}
    \label{fig:articles_per_year_rugby}
\end{figure}
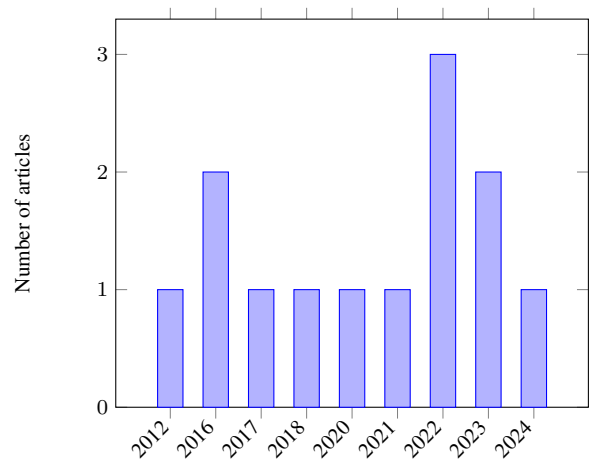

Rugby analytics has advanced significantly over time, with various studies exploring various methods for predicting match results, assessing player performance, and examining tactical approaches. This section highlights important research contributions in the field, focusing on methodologies, datasets, and outcomes (Table \ref{lbl_rugby} and Figures \ref{fig:enter-label-soccer-performance-rgb} and \ref{fig:articles_per_year_rugby}).

\citet{fontana2017player} conducted a study to determine if specific anthropometric and functional characteristics measured in youth rugby draft camps could accurately predict subsequent career progression in rugby union. They analyzed data from 531 male players (U16) who were classified according to their senior team representation at age 21-24 into International (Int) or National (Nat) categories. The study utilized multivariate analysis of variance (one-way MANOVA) to identify differences between groups using variables such as body mass, height, body fat, fat-free mass, squat jump (SJ), counter movement jump (CMJ), sprint times (t15m, t30m), and VO2max. A discriminant function (DF) combining these variables predicted group assignment with a hit rate of 81\% for Int and 77\% for Nat, with percentage of body fat and speed being the most influential predictors. The dataset used in this study was from the Italian Rugby Federation (FIR) youth draft camps.

Similarly, \citet{welch2018training} utilized GPS tracking data and machine learning algorithms to predict the risk of injury in professional rugby league players during a single season of the National Rugby League. The researchers collected training load (TL) data from 46 players in 4453 training sessions, focusing on six variables: total distance, high speed distance, acceleration / deceleration load, high metabolic power distance, impulse, and mechanical work. They employed supervised machine learning algorithms, specifically Random Forest (RF) and ANN classifiers, using a kernel-smoothed bootstrap sampling method to address the skewed dataset, which included 33 injuries. The RF classifier, particularly the 21-day window model, showed superior performance with an Area Under the Curve (AUC) of 0.98, compared to the ANN's highest AUC of 0.88 for the 14-day window. The results indicated that the sum of the high speed distance and the standard deviation of the total distance were significant predictors of the risk of injury.

In a related vein, \citet{scott2023performance} conducted a study to identify performance indicators (PIs) associated with match outcomes in the United Rugby Championship (URC), comparing the efficacy of isolated and relative datasets, and investigating whether reduced statistical models could maintain predictive accuracy. They selected 27 PIs from 96 matches in the 2020-21 URC season, using random forest classification on isolated and relative data to predict match outcomes. The full models correctly classified 83\% of matches for the relative dataset and 64\% for isolated data, with reduced models classifying 85\% and 66\%, respectively. The reduced relative model predicted 90\% matches in the 2021-22 season. The key PIs identified were hand kicks, meters made, clean breaks, turnovers conceded, and scrum penalties. The study concluded that relative performance indicators were more effective in predicting the outcomes of matches and reducing the features did not degrade accuracy, simplifying the practical application. The dataset utilized was from OPTA.

Expanding on this concept, \citet{xu2022machine} utilized a random forest model to evaluate the importance of nine performance indicators (Pass Forward, Penalty Kick, Tackle, Out, Scrum, Conversion, Yellow Cards, Red Cards, Possession Time) from the 14th China National Games rugby sevens dataset, consisting of 32 group-phase and 8 knockout-phase matches. The dataset was analyzed to convert these indicators into performance predictors. The random forest model, trained on group-phase data and tested on knockout-phase matches, achieved an 87.5\% accuracy in predicting match outcomes. Key metrics included mean decrease accuracy, with 'Conversion' and 'Possession Time' showing significant predictive power (mean decrease accuracy values of 27.35 and 16.56, respectively), indicating their strong correlation with winning matches.

Furthermore, \citet{kvasnytsya2024team} conducted a study to identify key performance indicators to predict match outcomes in the Rugby Union, using data from the European Rugby Championships 15 from 2021 to 2023, which included 41 matches. They analyzed 22 indicators for both home and away teams, such as possession, passes, tries, and missed tackles. The study utilized machine learning models, specifically the Random Forest Classifier and the Extra Trees Classifier, achieving a prediction accuracy of 92\%. The key findings included that the winning teams had significantly higher values in 15 indicators, such as tries (196.3\% higher), conversions (176.7\% higher), and offloads (126.3\% higher), while losing teams had higher missed tackles (-56.46\%). The model demonstrated that tries, conversions, offloads, and missed tackles were among the most significant predictors. The dataset used was from the official Rugby Europe website.

In addition, \citet{o2016predictive} compared 12 predictive models for the 2015 Rugby World Cup, using data from all previous tournaments and focusing on linear regression models. The most accurate model was one that used data from all seven previous tournaments, despite violating linear regression assumptions, and included World ranking points as a predictor variable. The model performed better than those considering only the most recent tournaments and was more effective without transforming the variables to satisfy the regression assumptions. The simulation based on this model, run 10,000 times, indicated that top-ranked teams underperformed while lower-ranked teams outperformed predictions, suggesting increasing competition depth. Additionally, the study found that having more recovery days than opponents provided a small but nonsignificant advantage of 4.1 points. The dataset used was from the International Rugby Board (IRB) World Rankings and previous Rugby World Cups.

Complementing these findings, \citet{cintia2016haka} employed a dynamic graph analysis framework to evaluate the performance of the rugby team by constructing multilayer networks representing passes and disruptions (tackles). The analysis used data from 18 rugby matches collected by Opta, covering the 2012 Tri-Nations championship, the 2012 New Zealand Europe tour, the 2012 Irish tour to New Zealand, and the 2011 Churchill Cup. Key metrics included connectivity, assortativity, number of strongly connected components, and clustering of the pass network, along with similar measures weighted by disruption events. The results indicated that high connectivity in the pass networks was correlated with more meters gained, highlighting the importance of maintaining multiple pathways for the ball. The study achieved a prediction accuracy of 83\% for the meters gained and 77\% for the outcomes of the matches, comparable to the predictions of the bookmakers, highlighting the robustness of the structural features of the network to predict rugby performance.

In parallel, \citet{bennett2021predicting} predicted the outcome of the matches using a random forest classification model based on performance indicators (PIs) from the 2015 Rugby World Cup group-phase and determined their relevance in the knockout-phase. The dataset, obtained from OPTA, included 26 PIs such as tackle ratio, clean breaks, average carry, lineouts won, penalties conceded, and missed tackles. The model identified 13 significant PIs to predict outcomes, with tackle ratio, clean breaks, and average carry being the most accurate standalone predictors, predicting 75\%, 70\%, and 73\% of the matches, respectively. The group-phase model correctly predicted 87.5\% of the knockout-phase matches, demonstrating its robustness across different competition phases.

Following a similar pattern, \citet{king2012playoff} developed a new simulation-based measure of playoff uncertainty to examine its impact on match attendance compared to other variants of playoff uncertainty. The model integrated match uncertainty, playoff uncertainty, past home team performance, and other control variables, using data from the Australian National Rugby League (NRL) for seasons 2004–2008. The results indicated that the probability of making the playoffs and the success of the home team significantly influenced the attendance of the match, more so than the uncertainty of the match itself. Alternative measures of playoff uncertainty, such as points behind the leader, captured similar effects on attendance. The simulation approach predicted match outcomes based on team strengths and home advantage, updating these factors throughout the season. Key findings included a 20\% higher attendance for teams certain of making the playoffs and a quadratic relationship between home win probability and attendance, although support for the uncertainty of outcome hypothesis (UOH) at the match level was modest. The study used NRL data, focusing on 924 regular season matches, and found that home team playoff probabilities and recent winning streaks were significant predictors of higher attendance, highlighting the importance of playoff contention in drawing crowds.

Furthermore, \citet{pituxcoosuvarn2022rugby} utilized a long- and short-term memory (LSTM) model to predict the success of rugby goal kicks based on posture data extracted from videos using OpenPose. The model aimed to provide feedback to players, focusing on key body parts such as the right knee, ankle, shoulder, and eyes. The dataset comprised 50 goal-kick videos from different angles, expanded to 4,000 data points using data augmentation techniques. The initial model trained on 50 data points had an accuracy of 48\%, which improved to 74\% with the augmented dataset. In addition, a domain knowledge-based model was tested using joints deemed crucial by expert advice and player surveys, achieving comparable accuracy. The dataset used in this study was OpenPose-extracted posture data from goal-kick videos.

Consequently, \citet{crewther2020longitudinal} investigated the bidirectional and time-dependent interrelationships between testosterone and training motivation in an elite rugby environment. Utilizing a continuous-time (CT) model, the researchers monitored 36 male rugby players on training, competition, and recovery days, collecting up to 40 pre-breakfast measures of salivary testosterone and training motivation (rated on a scale of 1-10). The CT model revealed that testosterone exhibited stronger temporal persistence compared to motivation. Cross-lagged effects indicated that training motivation positively influenced testosterone levels at later time points, peaking after 2.83 days (standardized effect = 0.25) before dissipating over longer intervals. In contrast, the influence of testosterone on subsequent motivation for training was positive, but not statistically significant. Furthermore, match days significantly affected both variables, with a predicted increase in testosterone and a decrease in motivation to train. These findings highlighted the complex, nonlinear, and lagged nature of the relationship between testosterone and training motivation in competitive sports, supporting theoretical models that link testosterone to competitive behaviors. The dataset used in this study was not explicitly named in the content provided.

Furthermore, \citet{scott2023classifying} identified performance indicators (PIs) that predicted match outcomes in Women's Rugby Union using isolated and relative datasets from 110 international matches (2017-2022), provided by OPTA. Random forest classification models were applied, revealing that the isolated model had an accuracy of 75\%, while the relative model had 78\%. Feature selection simplified these models without significant loss of accuracy. Key PIs included meters made, clean breaks, missed tackles, lineouts lost, carries, and kicks from hand. Both models showed high predictive performance on the World Cup 2021 data, with the reduced isolated model achieving 100\% accuracy and the relative model 96\%. The dataset utilized was from OPTA.

Building on this, \citet{howe2022modeling} used power law models to predict peak intensities in professional rugby union matches as a function of time. The model utilized Global Navigation Satellite Systems (GNSS) and accelerometers to collect movement data from 30 elite and 30 subelite rugby union athletes. The study analyzed mean speed, metabolic power, and PlayerLoad™ across durations from 5 seconds to 10 minutes. The results showed strong linear relationships (R² = .967–.993) between the logarithmic intensity of the highest exercise and the duration of the exercise, the backs having the highest predicted intensities for shorter durations than the forwards. However, the intensities for the backs decreased more dramatically as the duration increased. The random prediction errors of elite players ranged from moderate to large (5\% to 10\%), while the systematic prediction errors ranged from trivial to small (2\% to 4\%). Subelite players exhibited slightly greater errors. These findings suggested that power-law models could adequately predict peak intensities for prescribing training drills, although practitioners should consider prediction errors at the individual level. The dataset utilized was match movement data from professional rugby union athletes using wearable GNSS and accelerometers.

%{Lastly, \citet{stefani2009predicting} examined the predictability of score difference versus score total using least-squares and exponential smoothing methods. These methodologies involved calculating offensive and defensive ratings from past home and visiting scores, adjusted for home advantage. The results were first-order predictions of score difference and score total for future matches. The average absolute error was minimized using a 'shrinking factor' L, which indicated the reliability of past performances in predictions. The study, which analyzed over 3000 games from the Zurich/Guinness Premiership, Super 12/14 rugby, English Premier League, and Italian Serie A, found that past performance was a better predictor of score difference (with higher L values) than score total. This suggested strategic implications, where teams focus more on maintaining score differences to win or draw rather than on the total score. The study also explored gambling outcomes, showing that betting on score difference yielded higher profitability and reliability compared to betting on score total. This was supported by metrics such as average absolute error (AAE), and comparisons were made between different smoothing methods and their respective predictive accuracies. Dataset: Zurich/Guinness Premiership, Super 12/14 rugby, English Premier League, Italian Serie A.}

\begin{table*}[h!]
    \centering
    \caption{Summary of approaches in Golf prediction models}\label{lbl_golf}
    \centering
    \fontsize{6.5}{6.5}\selectfont
    \renewcommand{\arraystretch}{1.6}
    \begin{tabular}{|C{2cm}|C{2cm}|C{2cm}|C{1.5cm}|C{2.5cm}|C{2.5cm}|} \hline
\textbf{Approaches} & \textbf{Work} & \textbf{Performance} & \textbf{Metrics} & \textbf{Features} & \textbf{Datasets} \\ \hline
\textbf{Bayesian Linear Regression} & \citet{wiseman2016using} & Bayesian Linear Regression: 67\% within 3 shots & R², MAE, RMSE & Detailed shot-level data & ShotLink dataset (PGA Tour) \\ \hline
\textbf{SVM} & \citet{guo2014eeg} & SVM (Spectral Coherence): 62.11\% & Accuracy & Spectral coherence & Optiherence, LLC \\ \hline
\textbf{ANN} & \citet{chae2021victory}, \citet{bavcic2016predicting} & ANN: 80.2\%, ANN: 87\% & Classification accuracy & GIR, PA, birdies, swing plane segments & LPGA data (1993-2017), Device-embedded data \\ \hline
\textbf{OWGR Correlation} & \citet{laaksonen2023have} & OWGR Correlation: Pearson's & Pearson’s correlation & Strokes Gained, OWGR & PGA Tour, OWGR, major championship organizations \\ \hline
\textbf{Random Forest Classifier} & \citet{korpimies2020predicting} & RFC: Mean Absolute Difference 4.486\% & Mean Absolute Difference, standard deviation, variance & GIR, Putting Average, Driving Distance & PGA Tour data (2010-2019) \\ \hline
\textbf{Markovian Model} & \citet{maher2013predicting} & Markovian Model: Predicted 91.1\% chance for Europe to win & Winning, halving, or losing a hole probabilities & Historical probabilities of holes & Ryder Cup, major golf championships \\ \hline
\textbf{Stepwise Regression} & \citet{chae2018ranking} & Stepwise Regression: r = 0.689 & Regression analysis & Driving accuracy, GIR, putts per round & LPGA data (2013–2016 US Open) \\ \hline
\textbf{PCA and Regression Techniques} & \citet{leahy2014predicting} & PCA and Regression: Notable accuracy & Principal Component Analysis, regression techniques & Distance to pin, putts per round, driving accuracy & Shotlink dataset (PGA Tour) \\ \hline
\textbf{Multilevel Regression} & \citet{o2015can} & Multilevel Regression: Significant ICC values & Intraclass Correlation Coefficient & Par totals for par threes, fours, fives & 2014 US PGA Tour \\ \hline
\end{tabular}
\label{tab:golf_approaches}
\end{table*}

\begin{figure*}
    \centering
    \includegraphics[width=0.89\textwidth]{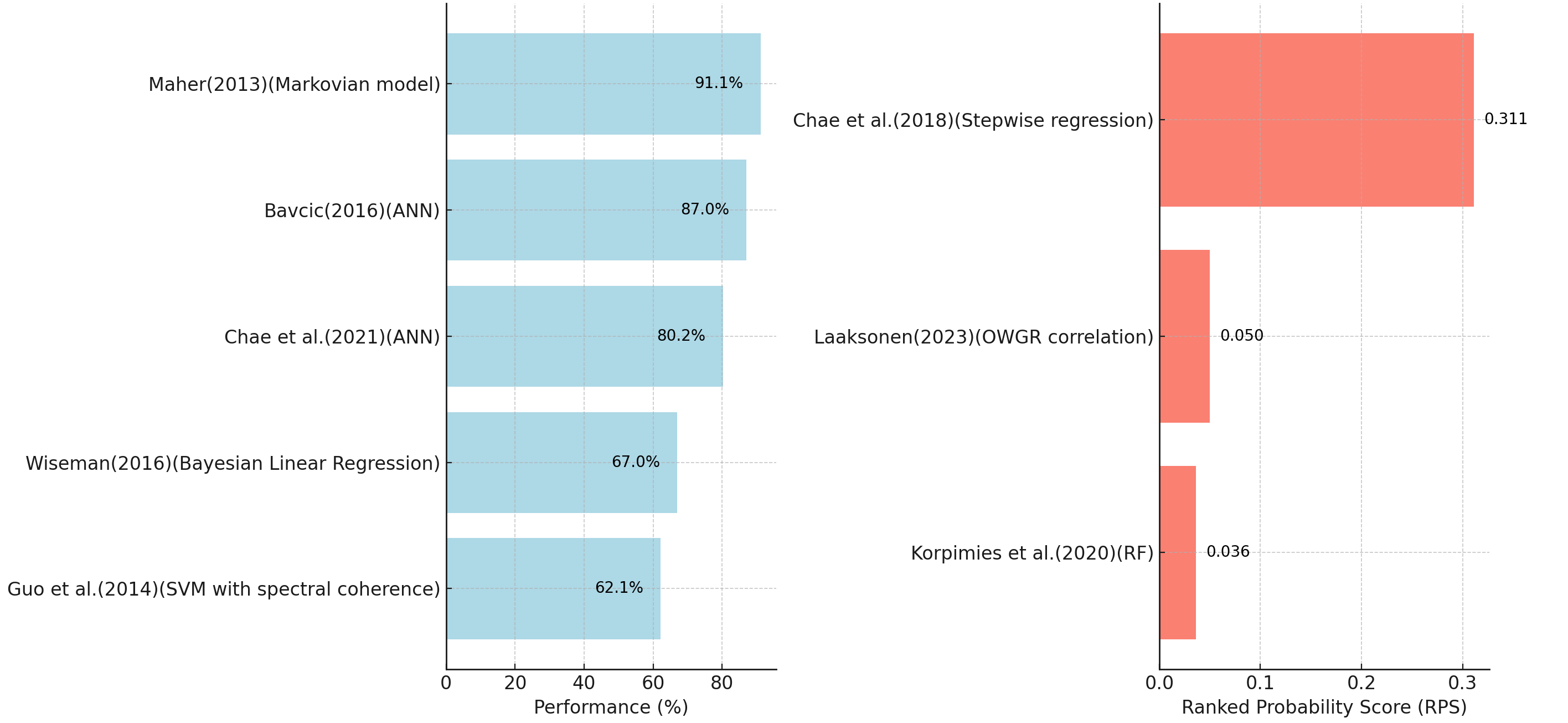}
    \caption{The best performances in Golf analytics based on accuracy, F1 and RPS}
    \label{fig:enter-label-soccer-performance-golf}
\end{figure*}

\subsection{Golf \includegraphics[height=0.3cm]{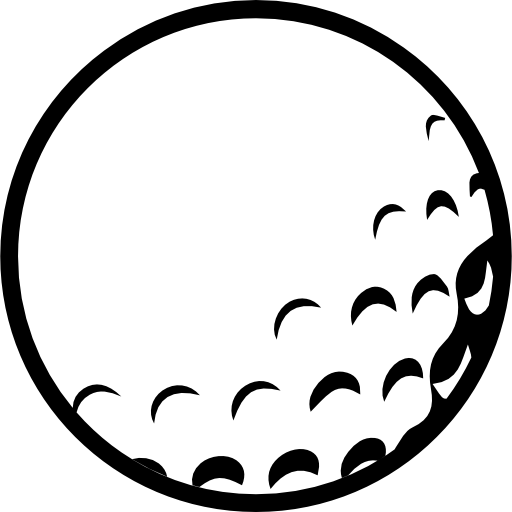}}\label{lbl_golf_sec}

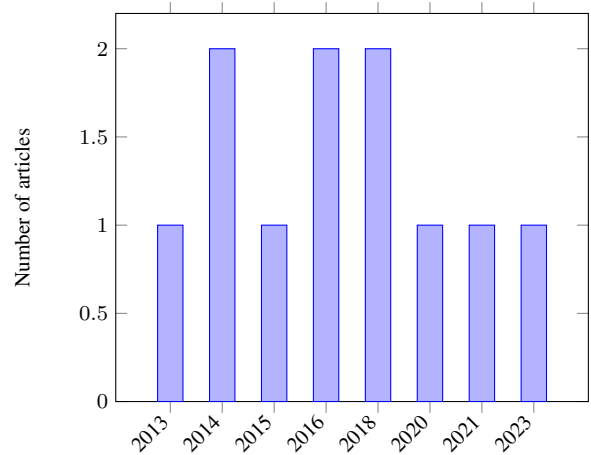
\begin{figure}[h!]
\fontsize{8}{8}\selectfont
\begin{tikzpicture}
    \begin{axis}[
        ybar,
        symbolic x coords={2013, 2014, 2015, 2016, 2018, 2020, 2021, 2023},
        xtick=data,
        x tick label style={rotate=45, anchor=east},
        ymin=0,
        xlabel={},
        ylabel={Number of articles},
        bar width=9.5pt,
        width=7.8cm,
        enlarge x limits=0.15,
    ]
        \addplot coordinates {(2013,1) (2014,2) (2015,1) (2016,2) (2018,2) (2020,1) (2021,1) (2023,1)};
    \end{axis}
\end{tikzpicture}
\caption{Histogram showing the number of articles per year in Golf betting.}
    \label{fig:articles_per_year_golf}
\end{figure}

The prediction of the outcome of golf tournaments has been studied using machine learning, leveraging historical results and player statistics. Various models consider factors such as course conditions, player form, and advanced metrics to better predict tournament results (Table \ref{lbl_golf} and Figures \ref{fig:enter-label-soccer-performance-golf} and \ref{fig:articles_per_year_golf}).

\citet{wiseman2016using} focused on predicting the winning score of golf tournaments using machine learning models. The study utilized data from the PGA Tour's ShotLink system and explored various algorithms, including Boosted Decision Tree, Bayesian Linear Regression, Decision Forest, Neural Network, and Linear Regression. The data spanned from 2004 to 2015 and included detailed shot-level data for each event. The key findings indicated that the Bayesian Linear Regression model performed the best, predicting the exact winning score in 22\% of the events and within 3 shots in 67\% of the events. The models significantly outperformed the existing methods by 50\% for predictions within one shot of the actual score. Important metrics utilized in the study included R², Mean Absolute Error (MAE), and Root Mean Squared Error (RMSE). The primary dataset utilized in this study was the ShotLink dataset provided by the PGA Tour.

Similarly, \citet{guo2014eeg} presented a method for predicting golf putt outcomes using EEG signals and SVM classification. The study collected multi-channel EEG data from golfers during a specific one-second interval prior to the putt, focusing on spectral coherence between different electrode pairs as the feature for SVM input. The dataset included 573 EEG trials from 26 golfers. The SVM model with spectral coherence features achieved a higher prediction accuracy compared to models using power spectral density (PSD), average PSD, and other commonly used features. Specifically, the SVM model with spectral coherence showed an overall prediction accuracy of 62.11\%, outperforming models with PSD and average spectral coherence features, which had accuracies of 43.16\% and 60.00\%, respectively. The study highlighted the efficacy of spectral coherence as a feature to classify EEG patterns related to golf putting performance and suggested that this method could be used to improve golfers' skills through brain-computer interface systems. The dataset utilized in the study was provided by Optiherence, LLC.

In a related vein, \citet{chae2021victory} identified the most accurate prediction model to determine the likelihood of victory in players of the Ladies Professional Golf Association (LPGA) using 25 seasons of annual average data (1993-2017). The models evaluated were decision tree, discriminant analysis, logistic regression, and ANN. Data were analyzed using SPSS 22.0 software and one-way ANOVA. Key predictive variables included greens in regulation (GIR), putting average (PA), birdies, top 10 finish percentage (T10), and official money (OM). The results indicated that the ANN model exhibited the highest accuracy in predicting victories, particularly with larger data sizes, showing classification accuracy rates of 75.3\% for skill variables, 75.7\% for skill results and 80.2\% for season outcomes. GIR and PA were crucial in all models for skill variables, while birdies were critical for skill results, and T10 and OM were significant for season outcomes. The study concluded that players who aim to win should focus on improving GIR, reducing PA, enhancing driving distance and accuracy, and increasing birdies to lower average strokes and increase the chances of being in the top 10.

Furthermore, \citet{bavcic2016predicting} applied ANNs to predict golf ball trajectories based on swing plane heuristics. The dataset comprised 531 samples from 14 golfers, collected using a device embedded in the handle of a driver club. Two ANN models, Radial Basis Function and SVM, were utilized to link variations in the swing plane to ball trajectories, achieving an overall classification accuracy of 87\%. Key metrics included the trajectory of the head of the club at impact and the segments of the swing plane, demonstrating that ANN can effectively validate and support empirical coaching rules in golf.

In addition, \citet{laaksonen2023have} aimed to determine whether the betting lines for the major golf championships were more influenced by the Strokes Gained model or the Official World Golf Ranking (OWGR). The study analyzed "To Win" odds for 19 major championships from 2018 to 2022. Laaksonen used Pearson’s correlation to assess the relationship between given odds and the players' OWGR positions and Strokes Gained statistics, including Total and Approach metrics. Furthermore, a Granger Causality test was conducted for 23 players ranked in the top 75 of the OWGR during the study period. The results indicated that the OWGR was the strongest correlating factor with the odds, especially for highly ranked players. However, for lower-ranked players, their odds were more influenced by their Strokes Gained metrics. The dataset utilized for this analysis was collected from the PGA Tour, OWGR, and relevant major championship organizations.

Likewise, \citet{korpimies2020predicting} developed and compared two classification models—Logistic Regression (LR) and Random Forest Classifier (RFC)—to predict the probability that golfers finish in the Top 10 in PGA Tour tournaments using historical data from 2010-2019. The dataset comprised various statistics tracked by the PGA Tour, such as Greens in Regulation (GIR), Putting Average, Driving Distance, and Driving Accuracy, with additional variables combined using Principal Component Analysis (PCA) to reduce multicollinearity. The RFC model outperformed the LR model, achieving a mean Absolute Difference (AD) of 4.486\% compared to 5.722\% for the LR model. The RFC model also had a lower standard deviation and variance, indicating more consistent predictions. Key predictive factors included previous Top 10 finishes and PGA Tour points accumulated over a player's career. The study highlighted that machine learning models can effectively predict golf outcomes, though the RFC model's reliance on historical success may limit its ability to identify emerging players. The PGA Tour data utilized in this study were publicly available and were sourced from official PGA Tour statistics.

Similarly, \citet{maher2013predicting} studied the outcome of the Ryder Cup using a Markovian model to provide a more informed prediction of the results of the matches based on the current score in each match, taking into account historical probabilities that the holes are won or halved from the Ryder Cup and other data from major professional golf tournaments. The model estimated expected points for each team and offered a probability distribution for potential outcomes, useful for 'in-play' betting. For example, during the 2010 Ryder Cup, the model predicted that Europe would win 16.11 points with a 91.1\% chance of securing at least 14½ points to win, a more nuanced prediction than conventional methods, which naively forecast 17 points. The metrics used included the probabilities of winning, halving, or losing a hole (pS, pW, pL), with historical data indicating higher halving probabilities for fourballs (0.595) than for foursomes or singles (around 0.478-0.483). The dataset comprised historical match data from Ryder Cup events and major golf championships, with detailed score analyzes that inform the probabilities of the model.

\citet{chae2018ranking} developed a ranking prediction model for Ladies Professional Golf Association (LPGA) players using data from the top 100 players on the tour money list from the 2013–2016 US Open. The study employed stepwise regression analysis to identify the impact of ten performance variables (driving accuracy, green in regulation, putts per round, driving distance, percentage of sand saves, par-3 average, par-4 average, par-5 average, birdies average, and eagle average) on five dependent season outcome variables (scoring average, official money, top-10 finishes, winning percentage, and 60-strokes average). Five different prediction models were generated, each targeting one of these outcome variables. The models were then tested on 38 players who participated in the 2016 Olympic Women's Golf Tournament, showing a significant correlation between predicted and actual rankings (r = 0.689, p < 0.001) and between predicted and actual average scores (r = 0.653, p < 0.001). This prediction model can help coaches and players identify potential participants for major competitions based on performance metrics. The dataset utilized in this study was the LPGA data from the 2013–2016 US Open.

In addition, \citet{leahy2014predicting} developed a model to predict the likelihood that professional golfers miss the cut in tournaments. The model utilized Shotlink data from the PGA Tour, focusing on various metrics such as distance to the pin, putts per round, and driving accuracy. Principal Component Analysis (PCA) and regression techniques were employed to reduce dimensionality and create predictive models. Key results included the identification of significant predictors for missing the cut, and the model achieved a notable accuracy rate in predictions. Metrics like "Strokes Gained - Putting" and average driving distance were pivotal in the analysis. The Shotlink dataset, which contains detailed shot-by-shot data for every PGA Tour event, was crucial for the research.

Furthermore, \citet{tanaka2018estimating} investigated the differences in sense-of-distance skill between expert and novice golfers by analyzing their putting-swing consistency and accuracy of outcome estimation. The experiment involved nine expert and nine novice golfers putting a ball to targets at distances of 1.2, 2.4, and 3.6 meters. The researchers measured putter head-swing movements using a motion-capture system and assessed outcome estimation accuracy by calculating the absolute error between the actual stopped position of the ball and the estimated position of the participants. The results showed that experienced golfers had significantly lower variability in putter-head kinematics and higher accuracy in outcome estimation compared to novices. Specifically, expert golfers exhibited less variability in maximum acceleration, velocity, amplitude, and movement time of their swings, which contributed to their higher consistency and accuracy. The study concluded that the consistency of the swing and the sense of distance are independent skills that contribute to the skill of putting, with experts demonstrating refined mental representations and sensorimotor feedback mechanisms that help to obtain more accurate predictions of the results of the putting. This research used a simulated golf green and motion-capture technology to gather its data. The specific dataset used in the study was derived from the recorded performances of the participating golfers in the experimental setup.

Finally, \citet{o2015can} examined the effect of player ability on scoring performance for holes of different par scores in PGA golf. Using multilevel regression modeling, they analyzed variance in par totals (PT) for par threes, fours, and fives based on data from the 2014 US PGA Tour. The model decomposed the variance into player and par total components, with the Intraclass Correlation Coefficient (ICC) indicating the proportion of variance attributable to the player. Significant ICC values were found for all pars, albeit small (.012 for threes, .046 for fours, .055 for fives). Controlling player strength revealed that as player skill decreased, ICC values and significance increased, indicating a greater influence of the player on score variance at lower skill levels. The results showed that player-specific modeling should be included in predicting golf scores, particularly for players at the bottom of the leaderboard, as the variance explained by player ability is more pronounced for less skilled players.

\begin{table*}[h!]
    \centering
    \caption{Summary of approaches in Hockey analytics}\label{lbl_hockey}
    \fontsize{6.5}{6.5}\selectfont
    \renewcommand{\arraystretch}{1.6}
    \begin{tabular}{|C{2cm}|C{2cm}|C{2cm}|C{1.5cm}|C{2.5cm}|C{2.5cm}|} \hline
\textbf{Approaches} & \textbf{Work} & \textbf{Performance} & \textbf{Metrics} & \textbf{Features} & \textbf{Datasets} \\ \hline
Markov Game Model & \citet{routley2015markov} & Improved accuracy with context and lookahead & Q-values, Accuracy & Player actions, goal scoring, penalties, game outcomes, play-by-play data & NHL play-by-play data from 2007 to part of the 2014-2015 season \\
\hline
Regularized Logistic Regression & \citet{gramacy2013estimating} & Identification of standout players, salary assessment & Accuracy, Precision, Recall, F1 score & Player contributions to team scoring, team and opponent effects & NHL data from 2007-2011, 1467 players, and 18154 goals \\
\hline
Neural Networks & \citet{weissbock2013use} & Highest accuracy of 59.38\% & Accuracy, Precision, Recall & Goals for and against, goal differential, Fenwick Close \%, PDO & 517 NHL games during the 2012-2013 season \\
\hline
Prediction Models & \citet{weissbock2014forecasting} & Accuracy of nearly 75\% for playoff series & Accuracy, Precision, Recall & Goals Against, Goal Differential, textual data from pre-game reports & In-game statistics and textual data from \url{www.NHL.com} \\
\hline
Meta-Classifiers & \citet{weissbock2014combining} & Accuracy of 60.25\% & Accuracy, Precision, Recall, F1 score & Goals Against, Goal Differential, location, word features from pre-game reports & 708 games from the 2012-2013 NHL season, sourced from \url{www.NHL.com} \\
\hline
Expert System & \citet{gu2019game} & Accuracy of 91.85\% with RobustBoost & Accuracy, Precision, Recall, F1 score & Various performance metrics including Corsi and Fenwick percentages & NHL regular season and playoff games from 2007-08 to 2016-17, over 1230 regular season games and 89 post-season games \\
\hline
SVM and Expert Judgments & \citet{gu2016expert} & Accuracy of 77.5\% & Accuracy, Precision, Recall, F1 score & Save percentage, Corsi, shooting percentage & 1,230 regular season games from the 2014-2015 NHL season, resulting in 2,460 records \\
\hline
Bayesian Predictive Density Estimator & \citet{sadeghkhani2021predicting} & Significant reduction in prediction error & Mean Squared Error (MSE), Prediction Error & Past performance data, points, expert opinions & Data from the 2016-17 to 2018-19 NHL seasons \\
\hline
Goal-Based vs. Shot-Based Metrics & \citet{found2016goal} & Goal-based metrics outperformed shot-based metrics & Accuracy, Precision, Recall, F1 score & Goal differential, shot differential, relative Corsi per minute & 10 years of NHL data (2005-2015) sourced from \url{www.NHL.com} and Behind The Net \\
\hline
Logistic Regression, Gradient Boosting, GAM & \citet{paerels2020play} & Highest AUC for cubic distance and quadratic angular terms & Area Under Curve (AUC), Accuracy & Shot attempts, distance, angle, type, player differences, rebounds, rushes & NHL Real-Time Scoring System (RTSS) data \\
\hline
Glicko Rating System & \citet{davis2021match} & Accuracy of 0.583, return on investment of 6.75\% & Brier Score, Accuracy & Box score events, team strength, variability & Detailed match statistics from Sportlogiq over four NHL seasons (2017-2021) \\
\hline
Decision Trees, ANN, ClusteR & \citet{pischedda2014predicting} & Highest accuracy of 61.54\% with VH-A dataset & Accuracy, Precision, Recall, F1 score & Categorical and continuous variables, team, location & Ottawa data from the 2012-2013 NHL season, encompassing 517 games \\
\hline
Logistic Regression, SVM, k-NN & \citet{chin2023predicting} & Highest accuracy of 77.8\% with Logistic Regression & Accuracy, Precision, Recall, Confusion Matrix & Team and match-specific features such as goals for, goals against, goal differential & NHL seasons 2015-2021, data from Kaggle \\
\hline
Human-Inspired Approach & \citet{ghazvini2011fast} & Neural network approach had the lowest errors & Mean Squared Error (MSE), Accuracy & Paddle motion, puck speed, and pose & Video captures from an air hockey table, simulations using an off-the-shelf camera and OpenCV for image processing \\
\hline
Penalty Corner Routines & \citet{vinson2013penalty} & Injecting ball from goalkeeper’s right increased scoring odds by 2.27 times & Accuracy, Precision & Tactical decisions, ball injection side, goalkeeper actions & 36 matches from the 2010-2011 England Hockey League Women’s Premier Division ‘Super Sixes’ competition \\
\hline
Entropy and Spatial Distribution & \citet{lord2023predicting} & Key strategies identified for goal shots & Accuracy, Precision, Recall, F1 score & Game possession, start and end locations of ball movements & 131 matches from the 2019 Pro League tournament \\
\hline
ML for Injury Risk Prediction & \citet{luu2020machine} & XGBoost AUC of 0.948 for position players, 0.956 for goalies & Area Under Curve (AUC), Precision, Recall, F1 score & Age, 85 performance metrics, injury history & Data for 2322 players from 2007 to 2017, including 2109 position players and 213 goalies \\
\hline
Decision Tree Model & \citet{morgan2013applying} & Accuracy of 64.3\%, AUC of 0.712 & Area Under Curve (AUC), Accuracy & Speed, direction, distance between players, angle of attack & 75 contests tracked using radio frequency devices \\
\hline
\end{tabular}
\end{table*}
\begin{figure*}
    \centering
    \includegraphics[width=0.79\textwidth]{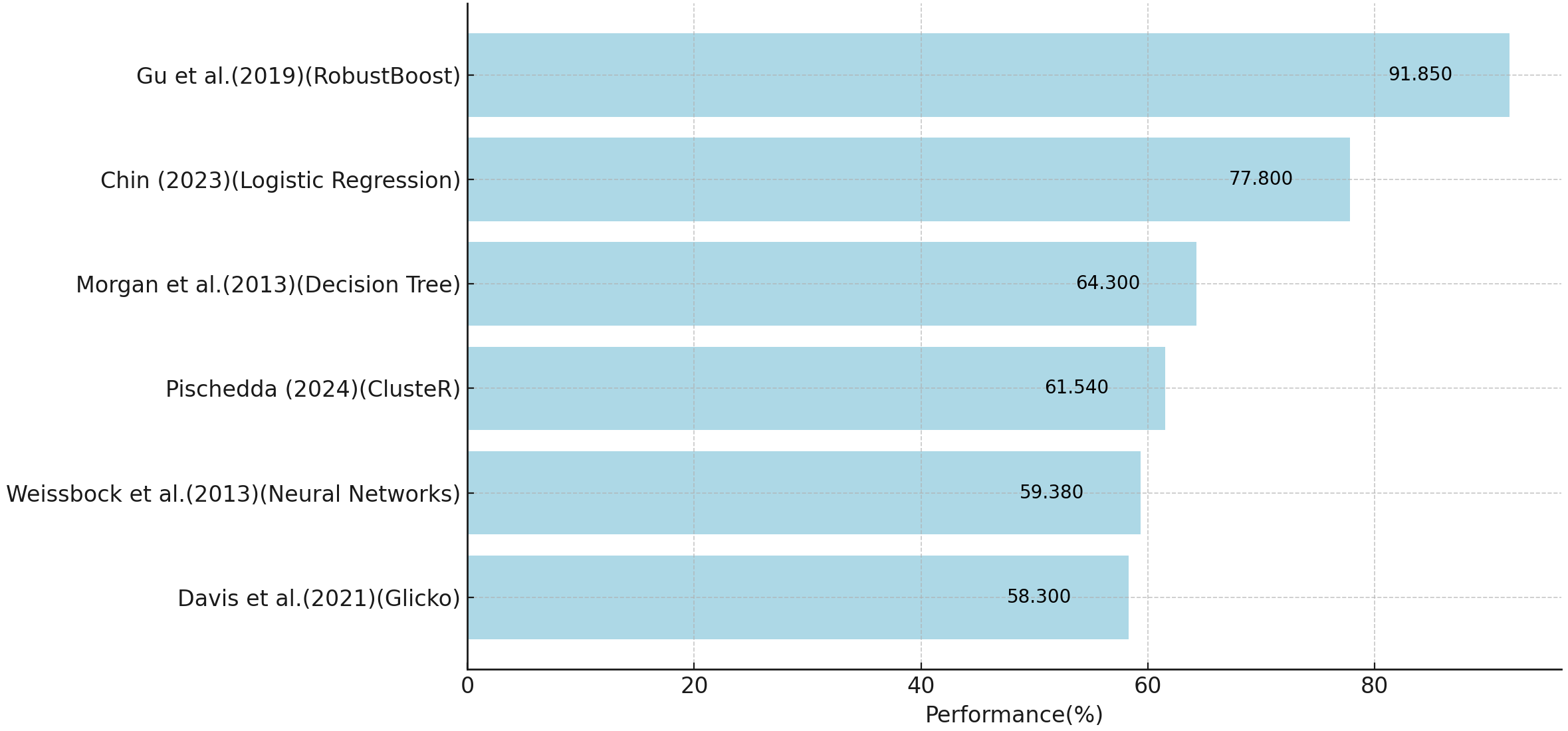}
    \caption{The best performances in Hokey analytics based on accuracy}
    \label{fig:enter-label-soccer-performance-hcky}
\end{figure*}

\subsection{Hockey \includegraphics[height=0.3cm]{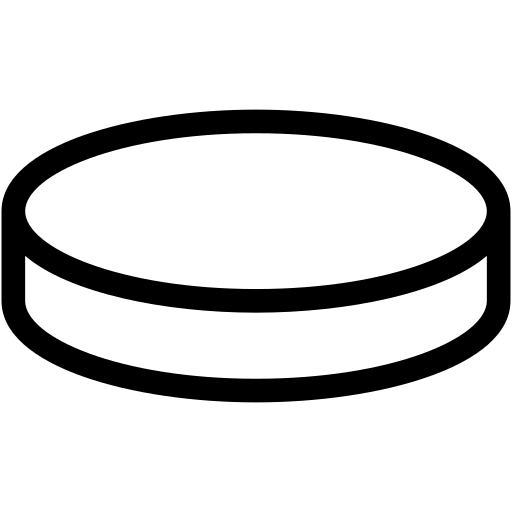}}\label{lbl_hockey_sec}

\begin{figure}[h!]
\fontsize{8}{8}\selectfont
\begin{tikzpicture}
    \begin{axis}[
        ybar,
        symbolic x coords={2011, 2013, 2014, 2015, 2016, 2019, 2020, 2021, 2023},
        xtick=data,
        x tick label style={rotate=45, anchor=east},
        ymin=0,
        xlabel={},
        ylabel={Number of articles},
        bar width=9.5pt,
        width=7.8cm,
        enlarge x limits=0.15,
    ]
        \addplot coordinates {(2011, 1)(2013, 4)(2014, 3)(2015, 1)(2016, 2)(2019, 1)(2020, 2)(2021, 2)(2023, 2)
        };
    \end{axis}
\end{tikzpicture}
\caption{Histogram showing the number of articles per year in Hokey betting.}
    \label{fig:articles_per_year_Hokey}
\end{figure}
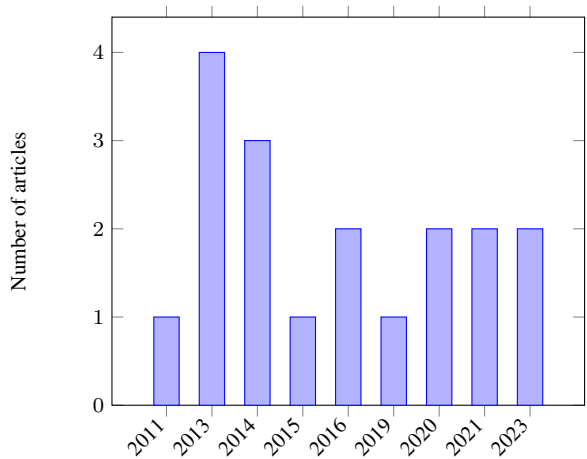

The prediction of the outcome of hockey games is based on statistical and machine learning models using historical data and stats of players. These models analyze team performance, individual skills, game conditions, and advanced metrics to forecast results with more accuracy (Table \ref{lbl_hockey} and Figures \ref{fig:enter-label-soccer-performance-hcky} and \ref{fig:articles_per_year_Hokey}).

\citet{routley2015markov} developed a Markov Game model to evaluate player actions in ice hockey using play-by-play data from the NHL, covering over 2.8 million events. The model incorporates context and lookahead to quantify the impact of player actions on goal scoring, penalties, and game outcomes. Using dynamic programming for value iteration, Q-values were learned for different game states, highlighting the context-sensitive nature of action impacts. The model's results indicated that including context features and lookahead substantially improved the accuracy of action impact scores. Players were ranked based on their actions' aggregate impact, showing significant correlations with traditional metrics like plus-minus and advanced statistics such as Corsi and Fenwick ratings. The dataset utilized for this research included NHL play-by-play data from 2007 to part of the 2014-2015 season.

%{Similarly, \citet{li2004description} utilized a Finite State Machine (FSM) model to analyze and predict player actions in ice hockey. The model took as input augmented player motion trajectory data, which included space-time points registered to rink coordinates, information on puck possession, and player attributes like shooting preference. The study focused on three specific game situations: 2-on-1 offensive attacks, defensive zone breakouts, and power play shots from the point. The FSM model represented knowledge of these situations and evaluated primitive hockey behaviors such as passes and shots. Results were presented as textual descriptions and simple 2D graphic animations, providing a detailed account of the players' actions, assessing outcomes, and suggesting better alternatives. The dataset utilized included trajectory data manually extracted and registered from game videos.}

\citet{gramacy2013estimating} developed a regularized logistic regression model to estimate the contributions of players to hockey. The traditional plus-minus statistic was found inadequate due to its noisy estimates and failure to account for sample size and team interactions. The proposed model assessed each player's contribution to the odds of their team scoring, controlling for team and opponent effects. The model employed L1 (Laplace) and L2 (Normal) penalties for regularization to handle the high-dimensional and imbalanced nature of hockey data. Data from the NHL covering four seasons (2007-2011) were used, which included 1467 players and 18154 goals. The results indicated that most players did not significantly deviate from the team average, allowing standout players to be more distinctly recognized. The model revealed that some highly paid players did not justify their salaries in terms of on-ice contributions, while others, like Pavel Datsyuk, showed substantial positive impacts. Traditional plus-minus metrics and the new model estimates showed discrepancies, highlighting the robustness of the new approach. The study also explored the implications of these findings for team strategies and player valuation, emphasizing the potential of the model in decision making for coaches and general managers.

Furthermore, \citet{weissbock2013use} applied machine learning techniques to predict the outcomes of NHL games, using a combination of traditional statistics (e.g., goals for and against, goal differential) and advanced performance metrics (e.g. possession metrics such as Fenwick Close \% and luck metrics such as PDO). The study found that neural networks outperformed other classifiers, achieving the highest accuracy of 59.38\% with a mixed dataset. Traditional statistics such as Goals Against and Goal Differential were identified as the most influential features. The dataset was collected from 517 NHL games during the 2012-2013 season, and data from each game were used to train and test various models using WEKA. The study highlighted the limited predictive power of advanced metrics for individual games compared to traditional statistics and suggested future research directions to improve model accuracy, such as incorporating additional features and analyzing larger datasets.

Similarly, \citet{weissbock2014forecasting} employed several models to predict the outcomes of NHL games and playoff series. The study primarily used traditional statistics and performance metrics in the game for prediction. For single-game predictions, traditional statistics provided the most value, achieving an accuracy of 59.8\%. Despite various combinations of features, the accuracy did not significantly exceed 60\%, suggesting a theoretical upper bound of about 62\% for predictions in a single-game. The predictions for the best-of-seven series of games, using more than 30 features, reached an accuracy of almost 75\%. The data set used included in-game statistics published on the NHL’s website and textual data from pre-game reports on \url{www.NHL.com}. Combining numerical data with textual analysis using Bag-of-Words and sentiment analysis in a multi-layer meta-classifier improved single-game prediction accuracy closer to the theoretical upper bound. The dataset utilized in this work was the in-game statistics published on the NHL’s website and textual data from pre-game reports on \url{www.NHL.com}.

Furthermore, \citet{weissbock2014combining} developed meta-classifiers to predict outcomes in the National Hockey League (NHL) by combining statistical data and pre-game textual reports. They utilized three classifiers: one based on numerical data from previous games (e.g., Goals Against, Goal Differential, Location), a second using word features from pre-game reports (unigrams, bigrams, trigrams), and a third analyzing sentiment (positive and negative word counts). The dataset comprised 708 games from the 2012-2013 NHL season, sourced from \url{www.NHL.com}. These classifiers were integrated into a meta-classifier using a cascade method, where outputs from individual classifiers were fed into a second-level classifier. They tested multiple machine learning algorithms (e.g., Neural Networks, Naïve Bayes, Support Vector Machines) and found that the best performing method was majority voting among classifiers, achieving an accuracy of 60.25\%, surpassing the accuracy of individual classifiers and traditional single classifier methods. The results indicated that combining different types of data sources, such as statistical and textual, improved the accuracy of the prediction, although the predictions were still constrained by the inherent randomness of sports outcomes. The baseline for comparison was a home-field advantage accuracy of 56\% and an upper bound accuracy of 62\%, highlighting the difficulty of making precise predictions in hockey due to its high level of unpredictability.

In a related vein, \citet{gu2019game} developed an expert system to predict the outcomes of National Hockey League (NHL) games. The system integrated principal component analysis (PCA), nonparametric statistical analysis, SVM, and ensemble machine learning algorithms. Data from NHL regular season and playoff games spanning the 2007-08 to 2016-17 seasons were used, comprising more than 1230 regular season games and 89 post-season games. The ensemble methods significantly improved predictive accuracy, exceeding 90\%, with discriminant analysis using RobustBoost achieving 91.85\% accuracy in the test set. The authors employed various performance metrics such as goals for (GF), goals against (GA), save percentage (SV\%), Corsi For percentage (CF\%), Fenwick For percentage (FF\%), and more to evaluate player and team performance. They found that goalie performance, especially save percentage, significantly impacts game outcomes, and that there is a notable home-ice advantage. The dataset utilized was compiled from multiple Web sources, including \url{www.NHL.com}, \url{www.ESPN.com}, and \url{www.hockey-reference.com}.

Moreover, \citet{gu2016expert} presented an expert system to predict the outcome of NHL games by integrating data mining with human judgment. The authors utilized an SVM model to assess the significance of various performance metrics and expert judgments in predicting the results of 89 post-season games. The dataset comprised of 1,230 regular season games from the 2014-2015 NHL season, resulting in 2,460 records. They reduced the initial 30 metrics to 17 key performance indicators using the Wilcoxon rank-sum test and validated these metrics with SVM, achieving an 88.3\% accuracy in classification. Their combined model, which incorporates both quantitative data and expert insights, yielded a prediction accuracy of 77.5\%, surpassing previous models in hockey game prediction. The metrics used included save percentage, Corsi, shooting percentage, and others, with correlations analyzed to fine-tune the model.

Furthermore, \citet{sadeghkhani2021predicting} proposed a Bayesian predictive density estimator to predict the time until the r-th goal is scored in a hockey game, utilizing past performance data, points, and expert opinions. The model employed a gamma distribution for the scoring time and introduced an advanced version of the weighted beta prime distribution for the estimator, which was shown to outperform traditional estimators through frequentist risk assessment and prediction error analysis. The dataset used included data from the 2016-17 to 2018-19 NHL seasons. The proposed method improved accuracy by incorporating ancillary information, such as the relative performance of teams, and demonstrated superior prediction accuracy. The Bayesian estimator's efficiency was highlighted by comparing prediction errors from the older dataset, showing a significant reduction in prediction error ($pe$) from 0.45 to 0.04 when ancillary information was used. The study specifically analyzed the waiting time for the Toronto Maple Leafs to score the third goal against the Montreal Canadiens, using points and performance data from previous seasons. The results indicated that the ancillary information led to more accurate predictions, as demonstrated by various statistical metrics and graphical comparisons of the predictive densities.

In conjunction with this, \citet{found2016goal} examined the predictive power of goal-based versus shot-based metrics on hockey success using statistical modeling of 10 years of NHL data (2005-2015) sourced from \url{www.NHL.com} and Behind the Net. The analysis revealed that goal-based metrics, such as team goal differential and individual relative plus-minus per minute of ice time, consistently outperformed shot-based metrics, such as the shot differential and relative Corsi per minute of ice time, in predicting team winning percentages and individual contributions to team success. The best model to predict team success was the goal differential model [winning \% = 0.5517 + (0.1811*goal differential)], which significantly predicted the winning percentages of the 2015-16 season (F1,28 = 12.29, p < 0.01), while the shot differential model did not. Furthermore, the study found that for the forwards, the relative goals per minute were significantly correlated with the success of the team, but the relative shots per minute were not, while for the defensemen, neither metric was a significant predictor of success.

Expanding on this concept, \citet{paerels2020play} used three types of models—logistic regression, gradient boosting, and generalized additive models (GAM)—to calculate expected goals (xG) in the NHL. The study focused on shot attempts in the final five minutes of regulation and overtime with the score tied or within one goal. Logistic regression identified the best model using a cubic distance term and a quadratic angular term, achieving the highest Area Under Curve (AUC) for predictive accuracy. The gradient boosting model had a lower AUC, but significant influence from distance and angle variables. The GAM showed similar significance for the distance, angle, and overtime variables, with the cubic distance and cubic angle model achieving the highest AUC. The dataset used was NHL Real-Time Scoring System (RTSS) data, which included variables like shot distance, angle, type, numerical player differences, rebounds, and rushes. The analysis showed that the percentages of xG in overtime were roughly twice as high as in the final five minutes of regulation, indicating a higher likelihood that goals were scored during overtime. Further research was suggested to explore the efficacy of gradient boosting and generalized additive models for xG prediction.

Complementing these findings, \citet{davis2021match} proposed a prediction model that utilizes the Glicko rating system to analyze box score data from NHL matches, in order to improve the prediction accuracy of game outcomes. The study leveraged detailed match statistics, such as shots on target and face-offs won, collected from Sportlogiq over four NHL seasons (2017-2021). By assigning Glicko ratings to various box score events for each team, the authors created dynamic covariates that accounted for team strength and variability, which were then incorporated into a logistic regression model. The effectiveness of the model was validated by comparing its performance against a baseline Glicko model that only used win-loss outcomes. The box score model demonstrated higher predictive accuracy and better Brier scores in both training and test sets, showing significant improvement in prediction and a notable ability to generate profit in simulated betting scenarios. The results indicated an accuracy of 0.583 and a Brier score of 0.244 for the 2020-2021 test season, outperforming the baseline model's 0.560 accuracy and 0.247 Brier score. Furthermore, the box score model achieved a return on investment of 6.75\% when betting on matches with significant probability deviations from market odds, highlighting its potential application in sports gambling. The dataset utilized was the Sportlogiq play-by-play data (\url{www.sportlogiq.com/}).

Moreover, \citet{pischedda2014predicting} employed multiple machine learning techniques to predict the outcomes of NHL games using data provided by a University of Ottawa team. The study utilized Decision Trees (DT), ANN, and proprietary software called ClusteR, which combines k-nearest neighbor techniques with rule-finding ones. The dataset, known as Ottawa data, included various attributes split into categorical and continuous variables. The models were built using both the original dataset and a new dataset created by subtracting the values of the home and away teams. The results indicated that the ClusteR model achieved the highest accuracy of 61.54\% when using the VH-A dataset, exceeding the perceived 60\% accuracy ceiling in predicting NHL games. The study highlighted that categorical variables, such as team and location, contributed significantly to prediction accuracy, more so than complex performance metrics. Furthermore, a practical application framework for betting was proposed, showing that ClusteR performed best with an average accuracy of 68.74\% in different model evaluations. The Ottawa dataset used for this research was originally collected from the 2012-2013 NHL season, which included 517 games.

In addition, \citet{chin2023predicting} explored the prediction of ice hockey results using various machine learning models, focusing on Logistic Regression (LR), SVM, and k-NN. They utilized data from the 2015-2021 National Hockey League (NHL) seasons, available from the Kaggle dataset (\url{https://www.kaggle.com/datasets/martinellis/nhl-game-data}), which included team and match-specific features such as goals for, goals against and goal differential. The study employed confusion matrices to present the results, summarizing the correct and incorrect predictions of the outcomes of the match. LR achieved the highest accuracy at 77.8\%, predicting 4449 games out of a total, outperforming the SVM with 77.7\% accuracy and k-NN with 69.7\%.

Lastly, \citet{ghazvini2011fast} focused on predicting the state of a high-speed puck in an air hockey game based on the state of the lower-speed paddle that hits it, using a human-inspired approach. This method eliminated the need for high-speed sensors by utilizing an off-the-shelf camera to capture the paddle's motion and predict the puck's speed and pose after being hit. Implemented and tested on both a simulator and real images, the model used linear and circular motion predictions, as well as a neural network approach. The results showed varying prediction accuracies: linear prediction had high errors, especially with wall banks, while circular prediction had lower errors except for linear paddle motions. The neural network approach had the lowest errors, indicating its potential superiority. The dataset included video captures from an air hockey table and simulations using an off-the-shelf camera and OpenCV for image processing.

\citet{vinson2013penalty} investigated the effectiveness of penalty corner routines in elite women’s indoor field hockey by analyzing 36 matches from the 2010-2011 England Hockey League Women’s Premier Division ‘Super Sixes’ competition. Using binary logistic regression, the researchers identified key predictors of successful penalty corners, focusing on tactical decisions such as the side from which the ball is injected and the goalkeeper's actions. The dataset comprised 319 penalty corners, with 72 resulting in goals (22.6\%). The strongest predictor of scoring was the injection of the ball from the right goalkeeper, which increased the chances of scoring by 2.27 times compared to injecting from the left. If the goalkeeper rushed to the edge of the circle, the odds that the attacking team did not score were 2.19 times higher than when the goalkeeper stayed close to the goal line. The study highlighted the importance of strategic decisions in the success of penalty corners and suggested further research on the technical and tactical aspects of these plays. The primary dataset used was from the round-robin phase of the 'Super Sixes' competition.

Similarly, \citet{lord2023predicting} analyzed the entropy and spatial distribution of ball movement patterns in international field hockey using video footage from the 2019 Pro League tournament, covering 131 matches. The notational analysis system in SportsCode was used to capture the start and end locations of ball movements and their outcomes. The metrics included game possession, entropy, possession per zone, and progression rates. Decision trees indicated higher circle possession, direct movements to the goal from deep attack, and lower entropy in build attack and defense as key strategies for goal shots. Unpredictability in ball movement was crucial for maintaining possession and destabilizing defense. The context of the match had minimal impact on the strategies, highlighting the diverse successful approaches. Key findings showed significant possession and entropy influencing play outcomes, with higher-ranked teams exhibiting more controlled and unpredictable ball movements, enhancing attacking opportunities. The dataset used was the video footage from the 2019 Pro League tournament.

In a related vein, \citet{luu2020machine} utilized machine learning (ML) to predict the risk of injury for the next season for players from the National Hockey League (NHL), comparing its performance with logistic regression (LR). The study compiled data for 2322 players from 2007 to 2017, including 2109 position players and 213 goalies, using attributes such as age, 85 performance metrics, and injury history from publicly reported databases Pro Sports Transactions and Hockey Reference. Five ML algorithms (random forest, K Nearest Neighbors, Naïve Bayes, XGBoost, and Top 3 Ensemble) were developed and validated using the area under the receiver operating characteristic curve (AUC). For position players, XGBoost achieved the highest AUC of 0.948, while LR had 0.937 (p< .0001). For goalies, XGBoost also performed best with an AUC of 0.956 compared to LR's 0.947 (p< .0001). SHAP scores identified the prior injury count as the greatest predictor of future injury.

Finally, \citet{morgan2013applying} used a decision tree model to analyze one-on-one interactions in elite female hockey players. The study involved 75 contests, tracked using radio frequency devices, to predict game outcomes (win or loss) based on the movements of the players. The model was trained with data from the first session and tested with data from the second session. Key predictive attributes included speed difference, lateral speed of the defender, and angle of attack. The results showed that if the attacker moved at least 0.5 m/s faster than the defender in the early epoch, the probability of winning was 100\%, while if it was slower, the probability of losing was 78\%. The model achieved an accuracy of 64.3\% and an area under the ROC curve of 0.712 when applied to the test dataset. The metrics used included speed, direction, distance between players, and angle of attack, providing a framework for understanding the dynamics of the attacker and defender in hockey.

\section{Discussion}\label{discussion}

This systematic review explores the application of machine learning in sports betting, focusing on various sports ranging from soccer to hockey. The studies included between Sections \ref{lbl_soccer} and \ref{lbl_hockey_sec} demonstrate the diverse methodologies and datasets used to predict outcomes and player performances, each highlighting the strengths and limitations of machine learning approaches in the context of sports betting.

In soccer, the extensive analysis of \citet{constantinou2013profiting} and the predictive models developed by \citet{tax2015predicting} underscore the significance of using historical match data and advanced statistical techniques to identify betting inefficiencies and predict match outcomes. The integration of public data and betting odds to achieve higher prediction accuracy, as shown in these studies, emphasizes the potential of hybrid models to outperform traditional betting approaches.

The review of basketball game predictions, particularly the work of \citet{chen2021hybrid} and \citet{cao2012sports}, reveals the effectiveness of combining various machine learning algorithms and feature engineering techniques. The high prediction accuracy achieved in NBA games through methods like XGBoost and logistic regression illustrates the adaptability of machine learning models to different types of sports data and their ability to provide valuable insights for betting strategies.

In tennis, the development of models by \citet{knottenbelt2012common} highlights the importance of considering player-specific statistics and match conditions. The hierarchical Markov model and Bayesian approaches employed in these studies demonstrate the nuanced understanding required to predict the outcomes of tennis matches accurately. The emphasis on integrating comprehensive datasets and the high return on investment achieved by these models underscore the economic viability of machine learning in tennis betting.

Cricket predictions, as explored by \citet{kumar2018outcome} and \citet{shenoy2022prediction}, focus on the application of decision trees, multilayer perceptrons, and ensemble methods to predict the outcomes of matches. These studies highlight the challenges of predicting outcomes in a sport with many variables and emphasize the need for robust feature selection and data pre-processing techniques to improve model performance.

American football studies, particularly those of \citet{otting2021predicting} and \citet{patel2023predicting}, demonstrate the potential of hidden Markov models and ensemble methods such as XGBoost to predict play calls and game outcomes. The high prediction accuracy achieved in these studies, along with the innovative use of play-by-play data and power rankings, illustrates the complexity of American football and the need for sophisticated models to capture the dynamic nature of the sport.

In baseball, the research by \citet{chang2021construction} and \citet{hamilton2014applying} highlights the use of Markov processes and machine learning algorithms to predict pitch types and game outcomes. These studies underscore the importance of feature selection and the use of detailed pitch data to achieve high prediction accuracy, showcasing the potential of machine learning to improve betting strategies in baseball.

Horse racing predictions, as investigated by \citet{terawong2024xgboost}, employ XGBoost and agent-based models to develop profitable betting strategies. The significant profitability achieved by these models emphasizes the potential of machine learning to revolutionize betting in dynamic and fast-paced environments such as horse racing.

Rugby predictions, analyzed by \citet{crewther2020longitudinal} and \citet{scott2023classifying}, explore the use of continuous-time models and random forests to predict game outcomes and player performances. The high prediction accuracy and identification of key performance indicators in these studies highlight the importance of integrating physiological and match data to improve betting strategies in rugby.

Golf predictions, as studied by \citet{laaksonen2023have} and \citet{leahy2014predicting}, focus on the use of advanced analytics and proprietary data to predict player performances. These studies illustrate the challenges of predicting results in individual sports and emphasize the need for detailed player statistics and environmental factors to improve the accuracy of the model.

Hockey predictions, as explored by \citet{wilkens2021sports} and \citet{davis2021match}, utilize various machine learning models to predict the outcomes of matches and the actions of the players. The high prediction accuracy achieved in these studies underscores the potential of machine learning to provide valuable information for betting strategies in complex dynamic team sports.

\section{Datasets, features, and evaluation metrics in sports analytics}\label{dataset_fts_evaluation}

This section provides a comprehensive overview of the datasets, features, and metrics used in various sports prediction models as summarized in Tables \ref{lbl_soccer_tbl} to \ref{lbl_hockey}. These tables detail approaches, works, performance metrics, features, and datasets for sports such as soccer, basketball, tennis, cricket, American football, baseball, horse racing, rugby, golf, and hockey. These elements help to understand the methodologies and tools used to predict outcomes in different sports disciplines ({\bf RQ1}).

\subsection{Datasets}

In Soccer, the datasets used for prediction models include extensive databases from various sources. For instance, \citet{tax2015predicting} used data from \url{www.football-data.co.uk}, \url{www.elfvoetbal.nl}, \url{www.transfermarkt.co.uk}, and \url{www.fcupdate.nl}. The studies by \citet{hervert2018prediction}, \citet{hervert2018bayesian}, and \citet{fialho2019predicting} used more than 200,000 match results from 52 leagues, sourced from the Open International Soccer Database. In the realm of geometric deep learning, \citet{wang2024tacticai} and \citet{goka2023prediction} used Liverpool FC datasets, corner kicks from the 2020-2021 Premier League seasons, and video clips from the 2019 and 2020 Japan J1 League seasons.

In Basketball, datasets such as the NBA 2018-2019 season data with 2460 game data points were used by \citet{chen2021hybrid}. Studies by \citet{cao2012sports}, \citet{lin2014predicting}, \citet{horvat2020impact}, \citet{houde2021predicting}, and \citet{sukumaran2022application} relied on NBA data that covered multiple seasons, including games from 1991-1998 and enhanced box scores from 2012-2018. In addition, data from Kaggle and the NBA official website were used.

Tennis prediction models have utilized datasets such as Wimbledon, OnCourt System, Jeff Sackmann's dataset, Tennis-Data.co.uk, and the Match Charting Project. Specific studies like \citet{sipko2015machine} and \citet{cornman2017machine} incorporated these datasets to predict match outcomes.

Cricket prediction models leverage data sets from sources such as ESPNcricinfo, the official IPL website, Cricinfo, and \url{www.Howstat.com}. Studies by \citet{kumar2018outcome}, \citet{vistro2019cricket}, and \citet{bharadwaj2024player} used these datasets to analyze past performance, ground conditions, and player statistics.

For the prediction of American football play, data from the NFL Big Data Bowl (2018), the Kaggle play-by-play dataset (2009-2017), NFL Next Gen Stats (2018-2019) and Pro Football Focus were instrumental. Studies by \citet{otting2021predicting} and \citet{cheong2021prediction} utilized these datasets for model training and validation.

In Baseball, datasets include the KBO, MLB data from 1961-2019, and specific datasets like those of Kaggle. \citet{lee2022prediction}, \citet{park2018deep}, and \citet{kim2023baseball} relied on these datasets to predict the results and the performances of the players.

Horse racing models have used data from the Bristol Betting Exchange (BBE) ABM, Horse Racing Company datasets (India) and Equibase website. \citet{terawong2024xgboost} and \citet{gupta2024horse} utilized these datasets to develop and test their predictive models.

Rugby datasets include those from the Italian Rugby Federation (FIR) youth draft camps, GPS tracking data from the National Rugby League, and OPTA. Studies by \citet{fontana2017player}, \citet{welch2018training}, and \citet{xu2022machine} used these datasets to evaluate player performance and game outcomes.

In golf, the ShotLink dataset (PGA Tour), LPGA data from 1993-2017, and Ryder Cup data have been used. Studies by \citet{wiseman2016using}, \citet{chae2021victory}, and \citet{maher2013predicting} leveraged these datasets for their predictive models.

Hockey analytics utilized NHL play-by-play data from 2007 to part of the 2014-2015 season, trajectory data manually extracted from game videos, and data from \url{www.NHL.com} and Behind The Net. Studies by \citet{routley2015markov}, and \citet{gramacy2013estimating} used these datasets for their analyses.

\subsection{Features}

In Soccer, features used in predictive models include dimensionality reduction, classifier combinations, historical patterns, team rankings, player attributes, spatio-temporal trajectory frames, event stream data, and player profiles. Studies by \citet{tax2015predicting}, \citet{hervert2018prediction}, and \citet{wang2024tacticai} utilized these features to improve the accuracy and predictive performance of the model.

Basketball prediction models incorporate features such as defensive rebounds, 2P FG\%, FT\%, offensive rebounds, assists, 3P FG attempts, comprehensive NBA statistics, points scored, FG attempts, team assists, steals, personal fouls, FG\%, 3P\%, FT\%, and offensive and defensive ratings. Studies by \citet{chen2021hybrid}, \citet{cao2012sports}, and \citet{lin2014predicting} relied on these features.

In tennis, features include historical match and player data, point-level data, match-specific data, serve statistics, player performance metrics, and environmental data. Studies by \citet{sipko2015machine} and \citet{cornman2017machine} employed these features for predictive modeling.

Cricket models use features such as past performance, ground conditions, innings, venue, player performance, team strength, runs, wickets, overs, player statistics, and weather conditions. Studies by \citet{kumar2018outcome}, \citet{vistro2019cricket}, and \citet{bharadwaj2024player} leveraged these features for their analyses.

American football play incorporate features such as game location, yards to go, down number, formation, score difference, field position, RFID tag sensor data, player movements, game time, distance to the goal line, score differential and percentage passing. Studies by \citet{otting2021predicting} and \citet{cheong2021prediction} utilized these features.

In baseball, features used include pitch types, locations, game attributes, spectators, sentiment analysis, starting lineups, OPS, runs, pitching data, and batting data. Studies by \citet{lee2022prediction}, \citet{park2018deep}, and \citet{kim2023baseball} incorporated these features into their models.

Horse racing models use features such as information gain, Chi-square filtering, Kelly betting strategy, previous prizes won, jockey and trainer characteristics, graph-based features, and basic race features. Studies by \citet{terawong2024xgboost} and \citet{gupta2024horse} utilized these features.

Rugby prediction models leverage features including body mass, height, body fat, fat-free mass, SJ, CMJ, sprint times, VO2max, total distance, high-speed distance, acceleration / deceleration load, high-metabolic power distance, impulse, mechanical work, kicks from hand, meters made, clean breaks, turnovers conceded, and scrum penalties. Studies by \citet{fontana2017player}, \citet{welch2018training}, and \citet{xu2022machine} employed these features.

In golf, features include detailed shot-level data, spectral coherence, GIR, PA, birdies, swing plane segments, strokes gained, OWGR, driving distance, driving accuracy, and putts per round. Studies by \citet{wiseman2016using}, \citet{chae2021victory}, and \citet{maher2013predicting} utilized these features for model development.

Hockey analytics use features such as player actions, goal scoring, penalties, game outcomes, play-by-play data, player motion trajectory data, puck possession, player attributes, player contributions to team scoring, team and opponent effects, goals for and against, goal differential, Fenwick Close \%, PDO, textual data from pre-game reports, and Corsi and Fenwick percentages. Studies by \citet{routley2015markov}, and \citet{gramacy2013estimating} employed these features.

\subsection{Metrics}

In Soccer, performance metrics include prediction accuracy, McNemar's test, RPS, top-3 accuracy, F1 score, human expert assessments, AP, precision, recall, AUC, BIC, RMSE, and cross-entropy. Studies by \citet{tax2015predicting}, \citet{hervert2018prediction}, and \citet{wang2024tacticai} utilized these metrics to evaluate the performance of the model.

Basketball prediction models measure performance using metrics such as accuracy, precision, recall, RMSE, kappa statistic, plus-minus, offensive rating, defensive rating, true shooting percentage, and logistic loss. Studies by \citet{chen2021hybrid}, \citet{cao2012sports}, and \citet{lin2014predicting} employed these metrics.

In tennis, metrics include accuracy, F1 score, ROI, logistic loss, Brier score, precision, recall, RMSE, kappa statistic, and percentage of first serves. Studies by \citet{sipko2015machine}, and \citet{cornman2017machine} used these metrics for performance evaluation.

Cricket prediction models use metrics such as precision, recall, F1 score, accuracy, AUROC, RMSE, error rates, and mean squared error. Studies by \citet{kumar2018outcome}, \citet{vistro2019cricket}, and \citet{bharadwaj2024player} relied on these metrics for their analyzes.

American football play prediction models measure performance using accuracy, precision, recall, RMSE, trajectory metrics, expected points (EP), win probability (WP), AUC, Brier scores, loss functions, error terms, and probabilistic forecasts. Studies by \citet{otting2021predicting} and \citet{cheong2021prediction} employed these metrics.

In baseball, performance metrics include accuracy, RMSE, MAPE, precision, recall, F1 score, AUC, Brier score, MAE, prediction error, return on investment, and probabilistic assignments. Studies by \citet{lee2022prediction}, \citet{park2018deep}, and \citet{kim2023baseball} utilized these metrics.

Horse racing models use metrics such as accuracy, ROC, profitability, NDCG, rate of return, R-squared, t-values, net profit function, ROI, correct bet ratio, prediction accuracy, and mean squared error. Studies by \citet{terawong2024xgboost} and \citet{gupta2024horse} employed these metrics.

Rugby prediction models measure performance using metrics like MANOVA, AUC, classification accuracy, mean decrease accuracy, temporal persistence, standardized effect, prediction errors, adjusted R², and average absolute error. Studies by \citet{fontana2017player}, \citet{welch2018training}, and \citet{xu2022machine} used these metrics.

In golf, metrics include R², MAE, RMSE, classification accuracy, Pearson’s correlation, mean absolute difference, winning probabilities, logistic regression, OLS methods, Spearman’s rank correlation, and principal component analysis. Studies by \citet{wiseman2016using}, \citet{chae2021victory}, and \citet{maher2013predicting} employed these metrics for evaluation.

Hockey analytics use metrics such as Q-values, accuracy, precision, recall, F1 score, prediction accuracy, mean squared error, AUC, net profit function, ROI, correct bet ratio, classification accuracy, and mean absolute difference. Studies by \citet{routley2015markov} and \citet{gramacy2013estimating} utilized these metrics to assess model performance.

\section{Machine learning platforms for betting tips}\label{ml_platforms_tips}

Machine learning platforms specializing in the commercialization of predictive analytics and insights offer bettors valuable tools to build their bet slips and parlays. \textbf{Sports AI} presents itself as a comprehensive AI-driven solution for sports bettors, leveraging machine learning to identify value bets and potentially profitable betting opportunities in various sports and bookmakers (\url{https://www.sports-ai.dev/}). Similarly, \textbf{DeepBetting}, a French startup, sells betting tips based on machine learning and deep learning algorithms trained on historical sports data, covering major football leagues (\url{https://deepbetting.io/}). \textbf{BetIdeas} analyzes statistics from more than 500 leagues around the world to provide free AI betting tips, exact score predictions, both teams to score tips, and a "bet of the day" feature (\url{https://betideas.com/}). Another notable platform, \textbf{1x2AI}, quantifies the confidence level of the predictions, allowing users to filter the tips based on how certain the algorithms are of the outcome (\url{https://1x2.ai/tips/}).

\textbf{Leans.ai} covers multiple sports such as soccer, tennis, basketball, and esports, offering a free trial and a 60-day money-back guarantee (\url{https://leans.ai/}). \textbf{FindYourBettingTips} focuses on AI sports betting predictions for football matches and publishes articles explaining the rationale behind each tip, supported by a Telegram community of over 5,000 members (\url{https://findyourbettingtips.com/}). \textbf{PredictBet} provides AI betting tips for matches up to two weeks in advance, together with a blog on sports news and betting insights (\url{https://predictbet.ai/}). \textbf{BetQL} offers detailed match predictions, expected value calculations, smart money tracking, and more using AI models (\url{https://betql.co/}). \textbf{WinnerOdds} employs AI algorithms to estimate real probabilities of match outcomes and find value bets, including tools like odds comparison and variable staking plans (\url{https://winnerodds.com/}). \textbf{BettorView} offers an AI betting tips free trial and boasts AI models that have achieved a 60\% ROI for NFL picks since 2016 (\url{https://www.bettorview.com/}). Lastly, \textbf{Scaleo} is an affiliate marketing platform that provides machine learning solutions for user segmentation, predictions of betting behavior, risk management and compliance checks in the gambling industry (\url{https://www.scaleo.io/}).

\section{Challenges and limitations}\label{challenge_limits}

The application of machine learning in sports betting presents several challenges and limitations that researchers and practitioners must navigate to enhance predictive accuracy and operational effectiveness. This section discusses key issues, including data availability, the dynamic nature of sports, model overfitting, feature selection, ethical concerns, computational resources, and regulatory challenges ({\bf RQ2}).

\subsection{Data availability and quality}
Data availability and quality are significant hurdles in the application of machine learning models for sports betting. Many sports may have limited historical data or incomplete records, which can hinder the development of robust predictive models. For instance, the effectiveness of machine learning techniques in basketball and soccer is highly dependent on comprehensive datasets that include player statistics, match conditions, and historical performance metrics \cite{miljkovic2010use,constantinou2013profiting}. Inadequate data can lead to biased predictions and limit the generalizability of models.

Moreover, the quality of the data is equally important. Poor-quality data, characterized by inaccuracies, inconsistencies, or missing values, can significantly impact the performance of machine learning algorithms. For example, in horse racing, the research by \citet{terawong2024xgboost} emphasizes the need for high quality datasets to develop profitable betting strategies using machine learning. Without reliable data, the predictive power of the models diminishes, leading to potentially costly betting decisions.

\subsection{Dynamic nature of sports}
The dynamic nature of sports introduces uncertainties that predictive models may struggle to account for. Factors such as player injuries, team dynamics, and changes in coaching strategies can significantly affect match outcomes \cite{taber2024holistic}. For example, rugby research by \citet{crewther2020longitudinal} and \citet{scott2023classifying} emphasizes the need for models that can adapt to these changing conditions to maintain the precision of prediction. Inability to incorporate real-time data and adapt to sudden changes can lead to outdated predictions and ineffective betting strategies.

Furthermore, the unpredictable nature of sports events, including unexpected player performances or weather conditions, further complicates the modeling process. In sports like golf, where individual player performance can vary widely based on external factors, studies by \citet{laaksonen2023have} highlight the challenges of accurately predicting results. As a result, models must be designed to incorporate a wide range of variables and remain flexible to adapt to new information, which can be a complex task.

\subsection{Model overfitting and generalization}

Model overfitting is a common challenge in machine learning, where a model learns the training data too well, capturing noise rather than the underlying pattern. This can lead to poor performance on unseen data. Studies in various sports, including basketball and soccer, highlight the importance of developing models that generalize well across different datasets \cite{horvat2020use,bunker2019machine}. Techniques such as cross-validation and regularization are essential to mitigate this issue, ensuring that models do not become too complex and lose predictive power.

Furthermore, the risk of overfitting is particularly pronounced in sports with a limited number of games or events, where the available data may not be sufficient to train robust models. %{For instance, in tennis, the hierarchical Markov models developed by \citet{klaassen2003forecasting} demonstrate the need for careful model selection to avoid overfitting while still capturing the nuances of player performance. Striking a balance between model complexity and generalization is crucial for achieving reliable predictions in sports betting.}

\subsection{Feature selection and engineering}
Feature selection and engineering are critical to improving the performance of machine learning models. The effectiveness of predictive models in sports betting often hinges on the inclusion of relevant features that capture the complexities of the sport. For instance, the work of \citet{kollar2021betting} emphasizes the need for advanced feature extraction techniques to handle the vast amounts of data generated in sports. Without careful selection and engineering of features, models may fail to capture important relationships, leading to suboptimal predictions.

In addition, the process of feature engineering can be resource intensive and requires domain expertise to identify the most relevant variables. In sports like cricket, where numerous factors influence the outcome of matches, the studies by \citet{kumar2018outcome} and \citet{shenoy2022prediction}  highlight the importance of integrating diverse datasets to improve model performance. As the landscape of sports analytics evolves, the ability to effectively engineer features will be paramount in developing models that can accurately predict outcomes and inform betting strategies.

\subsection{Ethical and integrity concerns}
The use of machine learning in sports betting raises ethical and integrity concerns, particularly regarding match-fixing and the potential for exploitation of insider information. Anomaly detection models are being developed to identify suspicious betting patterns that can indicate match-fixing \cite{kim2024ai,ramirez2023betting}. However, the implementation of such models must be approached with caution to avoid infringing on privacy rights and to ensure fair play in sports. The balance between leveraging data for predictive insights and maintaining the integrity of the sport is a critical consideration.

Additionally, the potential for machine learning to create an uneven playing field in betting markets raises ethical questions. As advanced algorithms become more accessible to certain bettors, there is a risk that those without access to sophisticated tools may be at a disadvantage. This disparity can lead to concerns about fairness and equity in sports betting, prompting discussions about the need for regulations that govern the use of machine learning technologies in this context \cite{matheson2021overview,gainsbury2018behavioral}. Addressing these ethical concerns is essential to foster trust and integrity in the sports betting industry.

\subsection{Computational resources}

The computational resources required to train complex machine learning models can be substantial. Advanced algorithms, such as neural networks and ensemble methods, require significant processing power and memory, which may not be accessible to all bettors or researchers \cite{walsh2024machine}. This disparity can create a competitive imbalance in the betting market, where only those with access to advanced computational resources can leverage sophisticated models effectively. 

Moreover, the need for real-time data processing further exacerbates the demand for computational resources. In fast-paced sports such as basketball and soccer, where decisions must be made quickly, the ability to analyze data in real-time is crucial for effective betting strategies. The studies on basketball predictions by \citet{chen2021hybrid} and \citet{cao2012sports} illustrate the importance of computational efficiency in the development of models that can provide timely information. As machine learning continues to evolve, ensuring equitable access to computational resources will be vital for fostering innovation and competition in the landscape of sports betting.

\subsection{Regulatory and legal challenges}

Finally, regulatory and legal challenges pose significant barriers to the widespread adoption of machine learning in sports betting. Different jurisdictions have different laws regarding gambling, data usage, and the application of predictive analytics. Researchers and practitioners must navigate these legal landscapes to ensure compliance while developing and deploying machine learning models \cite{matheson2021overview,gainsbury2018behavioral}. The evolving nature of regulations in the betting industry requires ongoing dialogue between stakeholders to address these challenges effectively.

In addition, the rapid advancement of machine learning technologies often outpaces the development of regulatory frameworks. This delay can create uncertainty for bettors and operators alike, as they may be unsure of the legal implications of using advanced predictive models. The need for clear guidelines and regulations that address the unique challenges posed by machine learning in sports betting is paramount to fostering a safe and responsible betting environment. As the industry continues to grow, collaboration between regulators, researchers, and practitioners will be essential to navigate these complexities.

\section{Future directions}\label{future_directions}

In traditional financial markets, portfolio optimization is a well-established strategy in which investors allocate assets in a way that balances risk and return, with the aim of maximizing profitability. This concept, rooted in Modern Portfolio Theory \cite{du2009note}, involves the selection of a mix of stocks, bonds, or other financial instruments that together provide the best possible return for a given level of risk. Portfolio management relies heavily on data analysis, predictive modeling, and optimization techniques to dynamically adjust asset allocations based on market conditions and investor goals. ML has further revolutionized this field by improving predictive capabilities and enabling real-time adjustments to portfolios that significantly improve decision-making in finance \cite{bartram2021machine}.

When we make a parallel with sports betting, the concept of a 'betting portfolio' is similar to financial portfolio management, aiming to optimize bet combinations to maximize returns and minimize risk. In such a context, ML can play a key role by analyzing vast datasets, including game results, player statistics, odds, and external factors like weather and team morale. Hence, ML models can be exploited to design diversified betting portfolios that adapt dynamically to game conditions, much like financial portfolios adjust to market changes ({\bf RQ3}). Beyond win-loss predictions, ML could also be designed to enable sophisticated portfolio management, treating bets as assets that affect overall risk and return \cite{abinzano2021sports}.

As the use of ML in betting grows, there is also a critical need for transparency in the models being deployed. In the financial world, transparent models are increasingly valued for their ability to provide interpretable insights into how investment decisions are made, which is essential to maintain investor trust and regulatory compliance. Similarly, in sports betting, the adoption of Explainable AI techniques, such as SHAP (Shapley Additive Explanations) and LIME (Local Interpretable Model-agnostic Explanations), can demystify complex ML predictions and provide bettors and stakeholders with clear, understandable explanations of why certain bets are favored or why odds are set in a particular way \cite{ribeiro2016should}. This transparency is not only vital for ethical considerations, but also helps bettors make more informed decisions, ultimately contributing to a more fair and accountable betting environment \cite{rudin2019stop}.

\section{Conclusion}

We have explored the impact of machine learning on sports betting and have highlighted its potential to be leveraged as a key component of a financial portfolio. The integration of ML into sports betting marks a major shift, transforming the industry into a data-driven sector with strong parallels to traditional financial markets. With machine learning, diverse datasets can be analyzed, such as historical game data, real-time player statistics, and social media sentiment. This capability improves predictive accuracy and optimizes betting strategies. By treating bets as assets within a 'betting portfolio', similar to financial portfolio management, machine learning enables dynamic adjustment of strategies, improving overall risk and return for bettors and bookmakers as conditions evolve.

As machine learning models such as deep learning and reinforcement learning advance, they offer opportunities to elevate sports betting into a sophisticated investment strategy similar to stock trading. These models facilitate the creation of adaptive betting portfolios that optimize returns and manage risks, driving profitability in a competitive market. The emphasis on transparency and explainability will be essential for maintaining ethical standards and regulatory compliance. By fully embracing these technologies, sports betting can evolve from a game of chance into a strategic financial activity, unlocking new growth opportunities and positioning itself alongside traditional financial sectors.

%\section*{Acknowledgments}

\bibliography{custom.bib}
\end{document}